\documentclass[10pt,journal]{IEEEtran}
\usepackage{amsmath,amsfonts}
\usepackage{algpseudocode}
\usepackage{array}
\usepackage{amssymb}
\usepackage[caption=false,font=normalsize,labelfont=sf,textfont=sf]{subfig}
\usepackage[citecolor=blue, colorlinks]{hyperref}
\usepackage{textcomp}
\usepackage{stfloats}
\usepackage{url}
\usepackage{verbatim}
\usepackage{graphicx}
\usepackage{cite}
\let\OldAsterisk\Asterisk 
\usepackage{mathabx}
 
\let\Asterisk\OldAsterisk
\usepackage{bm}
\usepackage{xcolor}
\usepackage{mathtools}

\usepackage{booktabs}
\usepackage{mathtools}
\usepackage{soul}
\usepackage{multirow}
\usepackage{capt-of}
\usepackage{tabu}
\usepackage{cuted}
\usepackage{bbding}
\usepackage{dashrule}
\makeatletter
\newcommand\newtag[2]{#1\def\@currentlabel{#1}\label{#2}}
\makeatother
\newcommand{\revision}[1]{\textcolor{black}{#1}}
\newcommand{\secrevision}[1]{\textcolor{black}{#1}}
\usepackage[ruled, vlined, linesnumbered]{algorithm2e}

\SetCommentSty{mycommfont}

\usepackage{threeparttable}

\begin{document}

\title{Learning Efficient and Effective Trajectories for Differential Equation-based Image Restoration}

\author{Zhiyu~Zhu,~Jinhui~Hou,~Hui Liu,  Huanqiang Zeng, and ~Junhui~Hou, \textit{Senior Member, IEEE}
\thanks{This work was supported in part by the NSFC Excellent Young Scientists Fund 62422118, in part by the Hong Kong Research Grants Council under Grant 11218121, and in part by Hong Kong  Innovation and Technology Fund ITS/164/23. Zhiyu Zhu and Jinhui Hou contributed to this paper equally. \textit{(Corresponding author: Junhui Hou)}.}
\thanks{Zhiyu Zhu, Jinhui Hou, and Junhui Hou are with the Department of Computer Science, City University of Hong Kong, Hong Kong SAR (e-mail: zhiyuzhu2@my.cityu.edu.hk; jhhou3-c@my.cityu.edu.hk; jh.hou@cityu.edu.hk). }
\thanks{Hui Liu is with the School of Computing and Information Sciences, Saint
Francis University, Hong Kong SAR (e-mail: h2liu@sfu.edu.hk).}
\thanks{Huanqiang Zeng is with the School of Engineering, Huaqiao University, Quanzhou 362021, China, and also with the School of Information Science and Engineering, Huaqiao University, Xiamen 361021, China (e-mail: zeng0043@hqu.edu.cn).}
}
\markboth{Revised Manuscript submitted to IEEE TPAMI}
{}
\maketitle
 
\begin{strip}
\begin{minipage}{\textwidth}\centering
\vspace{-90pt}
    \includegraphics[width=0.9\linewidth]{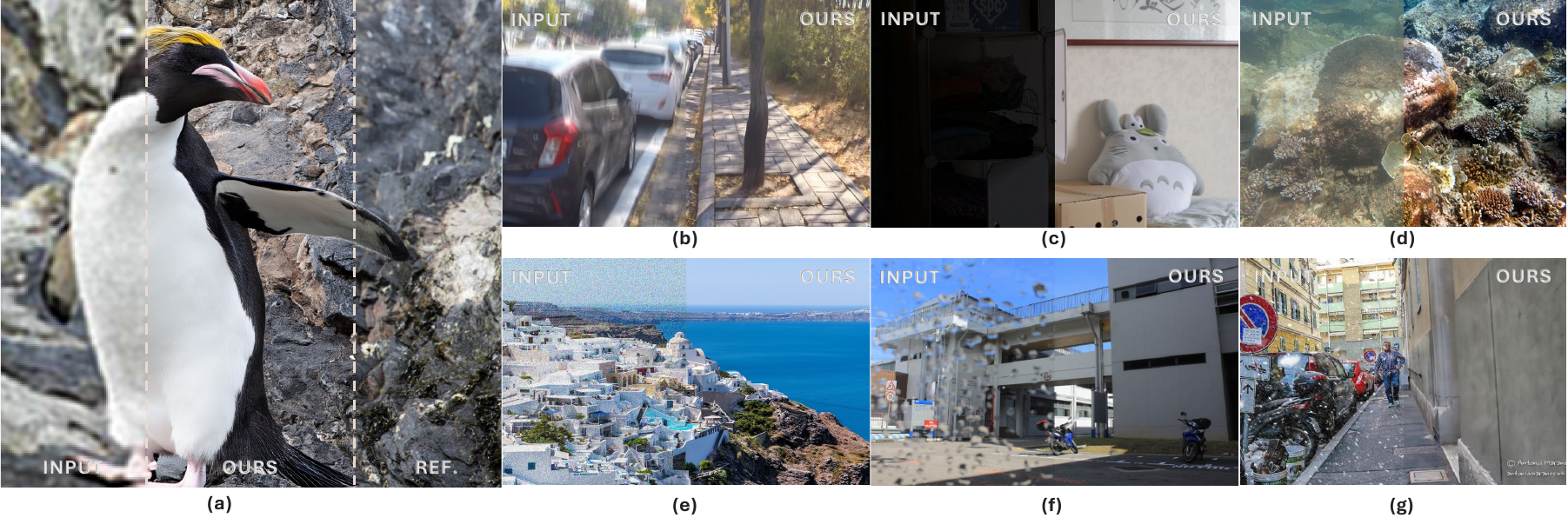}
\vspace{-0.5cm}
\captionof{figure}{Visual demonstration of the reconstruction performance of the proposed method across various image restoration tasks, where (a)-(e) witness its prowess in single image super-resolution, debluring, low-light enhancement, underwater enhancement, denoising, raindrop removal, and desnowing, respectively. Impressively, benefiting from the strong capacity of diffusion models, the proposed method excels at producing content clearer than the \textbf{Reference} images in (a). }
\vspace{-0.5cm}
\label{figurelabel}
\end{minipage}
\end{strip}

\begin{abstract}
The differential equation-based image restoration approach aims to establish learnable trajectories connecting high-quality images to a tractable distribution, e.g., low-quality images or a Gaussian distribution. In this paper, we reformulate the trajectory optimization of this kind of method, focusing on enhancing both reconstruction quality and efficiency. Initially, we navigate effective restoration paths through a reinforcement learning process, gradually steering potential trajectories toward the most precise options. Additionally, to mitigate the considerable computational burden associated with iterative sampling, we propose cost-aware trajectory distillation to streamline complex paths into several manageable steps with adaptable sizes.  Moreover, we fine-tune a foundational diffusion model (FLUX) with 12B parameters by using our algorithms, producing a unified framework for handling 7 kinds of image restoration tasks. Extensive experiments showcase the \textit{significant} superiority of the proposed method, achieving a maximum PSNR improvement of 2.1 dB over state-of-the-art methods, 
while also greatly enhancing visual perceptual quality. Project page: \url{https://zhu-zhiyu.github.io/FLUX-IR/}.
\end{abstract}


\section{Introduction}
\IEEEPARstart{I}{mage} restoration involves the enhancement of low-quality images affected by various degradations such as underwater and low-light conditions, raindrops, low-resolution, and noise to achieve high-quality outputs. It serves as a fundamental processing unit for visual recognition~\cite{noh2019better,dong2014learning}, communication~\cite{lu2019learned} and virtual reality~\cite{conde2022swin2sr}. Traditional image restoration methods typically rely on optimization procedures incorporating human priors such as sparsity, low rankness, and self-similarity~\cite{yang2010image,yang2008image}. The emergence of deep learning techniques~\cite{lecun2015deep} has significantly transformed this domain. Initially delving into neural network architectures~\cite{dong2015image}, image restoration has progressed beyond simple regression networks~\cite{dong2015image,zhang2018residual,zamir2022restormer,liang2021swinir}, exploring avenues like adversarial training~\cite{ledig2017photo,wang2018esrgan}, algorithm unrolling~\cite{monga2021algorithm,zhang2020deep,marivani2020multimodal} and flow-based methods~\cite{lugmayr2020srflow,kim2021noise,liang2021hierarchical}.

\begin{figure*}
    \centering
    \includegraphics[width=0.9\linewidth]{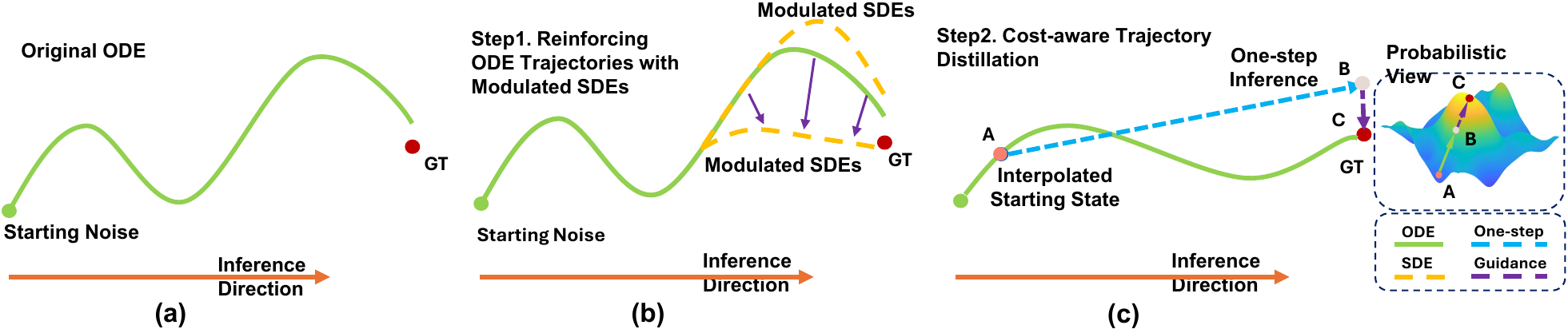}\vspace{-0.4cm}
    \caption{ Illustration of the workflow of the proposed method. Given a pre-trained diffusion model for image restoration, our trajectory optimization process contains the following two stages. (\textbf{1}) \textbf{Reinforcing ODE Trajectories with Modulated SDEs} in (b), which aligns the deterministic ODE trajectory shown in (a) to the most effective modulated SDE trajectory. (\textbf{2}) \textbf{Cost-aware trajectory distillation} in (\textbf{c}), which achieves high-quality one-step inference via delicate designs based on the task, with the knowledge of the original pre-trained model preserved.  Note that to preserve the original knowledge of the pre-trained diffusion model, we aim to find a trajectory with less modification of gradient $\frac{d\mathbf{X}}{dt}$. Through theoretical analyses and experimental validations, we find that for image restoration tasks, degraded measurements usually lie in the low probability region from the probabilistic space of high-quality samples. Thus, we also utilize the input measurements as negative guidance to rectify the gradient of log-density. As shown in the sub-figure of the probabilistic view, $\mathbf{A}$, $\mathbf{B}$, and $\mathbf{C}$ correspond to the low-quality measurement, reconstructed sample, and reconstruction by a low-quality image as negative guidance, respectively. We refer the readers to Fig.~\ref{fig:trajectory}, which illustrates the trajectories directly from the diffusion data points.}
    \label{fig:illustration}
\vspace{-0.6cm}
\end{figure*}

Recently, a new category of generative models, namely diffusion models, has shown their strong potential for image synthesis and restoration~\cite{xia2023diffir,saharia2022image}. Generally, diffusion models construct probabilistic flow (PF) between a tractable distribution and the target distribution. The forward process typically involves incrementally introducing noise until reaching a manageable distribution, often a Gaussian. On the other hand, the reverse process can be obtained by maximizing the posterior of the forward Markov chain~\cite{ho2020denoising} or by sampling the reverse stochastic differential equation (SDE) \cite{anderson1982reverse, song2020score} or ordinary differential equation (ODE) \cite{song2020score}.

Three categories of methods stand out for harnessing the potent generative capabilities of diffusion models for image restoration. The first category~\cite{kawar2022denoising,chung2022diffusion,wang2022zero} leverages the progressive integration nature of the differential equations to maximize the diffusion posterior on the scene of a low-quality image, then progressively reversing to the high-quality samples. Although sampling-based methods can directly take advantage of large, pre-trained models and get rid of network training, the posterior optimization process may take more time and computational resources~\cite{fei2023generative,kawar2022denoising}, and their performance is noncompetitive compared with supervised training diffusion model~\cite{luo2023image}. The second category incorporates the reconstruction outcomes from a fixed pre-trained diffusion model as a prior~\cite{wang2024exploiting}, subsequently refining these outcomes through trainable neural networks, akin to algorithm unrolling techniques. Lastly, the third set of methods~\cite{xia2023diffir,luo2023image} trains a diffusion trajectory conditioned on low-quality samples~\cite{luo2023image} or establishes a direct linkage between low-quality and high-quality image distributions~\cite{chung2024direct}. Since the diffusion trajectory is explicitly refined by training on the paired dataset, this kind of approach has the most potential and effectiveness.

Due to the probabilistic nature and separate training approach of diffusion models, the reverse generation trajectory might exhibit instability or chaos~\cite{liu2022flow}.
To tackle such an issue, some work tends to rectify the generation trajectories to be straight~\cite{liu2022flow} or directly train a consistency model~\cite{song2023consistency}. Although this manner allows us to achieve adversarial training during the diffusion process, excessive regularization could significantly impair diffusion performance. As shown in Fig.~\ref{fig:illustration}, we aim to re-align the diffusion trajectory towards the most effective path using a reinforcement learning approach. Furthermore, considering that the resulting trajectories may be complex and require extensive steps for sampling, we then propose a novel trajectory distillation process to alleviate this issue, which analyzes and lessens the cost of a diffusion model distillation process. Extensive experiments demonstrate the significant advantages of the proposed trajectory augmentation strategies on various image restoration tasks over state-of-the-art methods.

In summary, we make the following key contributions in this paper: 
\begin{itemize}
    \item we propose a novel trajectory optimization paradigm for boosting both the efficiency and effectiveness of differential equation-based image restoration methods;
    \item we theoretically examine the accumulated score estimation error in diffusion models for image restoration and introduce a reinforcement learning-based ODE trajectory augmentation algorithm that leverages the modulated SDE to generate potential high-quality trajectories;
    \item we improve the inference efficiency by introducing an acceleration distillation process, where we preserve the model's original capacity via investigating the distillation cost and utilizing the low-quality images for initial state interpolation and diffusion guidance;
    \item we establish new records on various image restoration tasks, including de-raining, low-light, under-water enhancement, image super-resolution, image de-blurring, and image de-noising;
    \item we calibrate a unified diffusion model for various image restoration tasks, based on the recent foundational diffusion model named \textit{FLUX-DEV} with 12B parameters.
\end{itemize}

The remainder of this paper is organized as follows. Sec.~\ref{Sec:RelatedWork} briefly reviews related work concerning diffusion models. Additionally, Sec.~\ref{Sec:Preliminary} offers essential mathematical formulations serving as the foundational backdrop for the proposed approach. Sec.~\ref{Sec:Method} details the proposed method, followed by comprehensive experiments in Sec.~\ref{Sec:Exp}.  Finally, Sec.~\ref{Sec:conclusion} concludes this paper.

\section{Related Work}
\label{Sec:RelatedWork}
The differential equation-based deep generative model~\cite{song2020score} represents a kind of learning strategy inspired by physical non-equilibrium diffusion processes~\cite{sohl2015deep}. It involves a forward diffusion process that progressively introduces noise into the data until a tractable distribution is reached, followed by a reversal to establish the data generation process. Ho~\textit{et al.}~\cite{ho2020denoising} firstly explored an effective diffusion formulation by parameterizing the reverse process as maximizing the posterior of reverse steps. Song ~\textit{et al.}~\cite{song2020score,song2019generative,song2020improved,song2021maximum} generalized such discrete diffusion steps into a continuous formulation by stochastic differential equation. Thus, the diffusion process can then be treated as an integral of the corresponding differential equation. Besides, the ordinary differential equation introduced in~\cite{song2020improved} removes the additional noise in the reverse process, which enables more inference acceleration designs~\cite{lu2022dpm,lu2022dpm2,zheng2024dpm}. Xu~\textit{et al.}~\cite{xu2022poisson} introduced a Poisson noise-based diffusion model. Robin~\textit{et al.}~\cite{rombach2022high} converted the diffusion process into the latent domain to achieve high fidelity high-resolution diffusion. Blattmann~\textit{et al.}~\cite{blattmann2023stable,blattmann2023align} extended the image diffusion model into a high-dimensional video diffusion process. Karras~\textit{et al.}~\cite{karras2022elucidating} examined different formulations of the diffusion model and proposed a concise formulation.
Dockhorn~\textit{et al.}~\cite{dockhorn2021score} augmented the diffusion space by introducing an auxiliary velocity variable and constructing a diffusion process running in the joint space. Chen~\textit{et al.}~\cite{chen2022sampling} conducted theoretical convergence analysis on the score-based diffusion model. 

In addition to the above, some diffusion-based methods bridge the distributions between different types of images. For example, Liu~\textit{et al.}~\cite{liu2022let} constructed a diffusion bridge by applying maximum likelihood estimation of latent trajectories with input from an auxiliary distribution. Li~\textit{et al.}~\cite{li2023bbdm} proposed an image-to-image translation diffusion model based on a Brownian Bridge diffusion process. Zhou~\textit{et al.}~\cite{zhou2023denoising} utilized the $h$-transform to make a constraint of the end-point of forward diffusion with formulations of reverse process derived by reformulating Kolmogorov equations. 

\subsection{Sampling Strategies for Accelerating Diffusion Process}

Given the progressive noise reduction process, hundreds of steps are usually required to sample high-quality images. One solution to expedite inference involves crafting a refined sampling strategy. For example, Song~\textit{et al.}~\cite{song2020denoising} pioneered the use of denoising diffusion implicit models (DDIM) to hasten diffusion sampling by disentangling each step in a non-Markov Chain fashion. Leveraging the semi-linear attributes of ODE and SDE formulations in diffusion models, Lu~\textit{et al.}~\cite{lu2022dpm,lu2022dpm2,zheng2024dpm} introduced a series of integral solvers with analytic solutions for the associated ODEs or SDEs.  Zhang~\textit{et al.}~\cite{zhang2022fast} delved into the significant variance in distribution shifts and isolated an exponential variance component from the score estimation model, thereby mitigating discretization errors.
Zhou~\textit{et al.}~\cite{zhou2024fast} introduced learnable solvers grounded in the mean value theorem for integrals.  Xue~\textit{et al.}~\cite{xue2024sa,xue2024accelerating} proposed accelerating diffusion sampling through an improved stochastic Adams method and precise ODE steps.  Dockhorn~\textit{et al.}~\cite{dockhorn2022genie} advocated for higher-order denoising diffusion solvers based on truncated Taylor methods.

\subsection{Trajectory Distillation-based Diffusion Acceleration}
An alternative solution for accelerating generation involves directly adjusting diffusion trajectories.
Liu~\textit{et al.}~\cite{liu2022flow}  introduced a method for straight flow regularization, which certifies the diffusion generation trajectory to be linear. Song~\textit{et al.}~\cite{song2023consistency,song2023improved} presented the consistency model, aligning each point on the trajectory directly with noise-free endpoints. Kim~\textit{et al.}~\cite{kim2023consistency} introduced trajectory consistency distillation to regularize gradients that can consistently map to corresponding points on the trajectory. Moreover, Zhou~\textit{et al.}~\cite{zhou2024score,zhou2024long} proposed to make distillation of a pre-trained diffusion model to a student one-step generator via measuring discrepancy by additional learned score function. 

Moving beyond training the diffusion model into a single-step generator, segmenting the process into multiple sections, with each being linear, presents a viable solution for rapidly generating high-quality outcomes in diffusion models~\cite{ren2024hyper,wang2024phased,xie2024mlcm}.

\subsection{Diffusion-based Image Restoration}
In the realm of diffusion-based image restoration, we outline works that fall into two main categories: those employing a training-free diffusion sampling strategy
~\cite{kawar2022denoising,chung2022diffusion,wang2022zero,zhu2023denoising,ozdenizci2023restoring} and those relying on model training~\cite{xia2023diffir,zhang2024unified,luo2023refusion,luo2023controlling,luo2023image}. 

Within the first category,  Kawar~\textit{et al.}~\cite{kawar2022denoising} introduced an unsupervised posterior sampling method for image restoration using a pre-trained diffusion model. Chung~\textit{et al.}~\cite{chung2022diffusion} proposed to regularize the intermediate derivative from the reconstruction process to improve image restoration. Wang~\textit{et al.}~\cite{wang2022zero} decoupled the image restoration into range-null spaces and focused on the reconstruction of null space, which contains the degraded information.  Zhu~\textit{et al.}~\cite{zhu2023denoising} combined the traditional plug-and-play image restoration method into the diffusion process. Regarding the training-based methods, Luo~\textit{et al.}~\cite{luo2023image} proposed a mean-reverting SDE with its reverse formulation to boost diffusion-based image restoration. Jiang~\textit{et al.}~\cite{jiang2023low} proposed a wavelet-based diffusion model for low-light enhancement. Yi~\textit{et al.}~\cite{yi2023diff} introduced a dual-branches diffusion framework combining reflectance and illumination reconstruction process. Wang~\textit{et al.}~\cite{wang2024sinsr} proposed a DDIM-inspired diffusion-based framework for the distillation of an image restoration model. Tang~\textit{et al.}~\cite{tang2023underwater} introduced a transformer-based model for underwater enhancement using diffusion techniques.
\revision{
Recently, several works \cite{potlapalli2306promptir, lin2023diffbir, jiang2023autodir, luo2023controlling, cui2024adair, yan2024textual, zheng2024selective, ai2024multimodal, conde2024instructir, yu2024multi, zhang2024diff} have focused on all-in-one image restoration using diffusion models. For example, TextPromptIR \cite{yan2024textual} and DA-CLIP \cite{luo2023controlling} integrate textual or visual prompts to guide diffusion processes. Ai~\textit{et al.}~\cite{ai2024multimodal} proposed a multimodal prompt learning method designed to leverage the generative priors of Stable Diffusion. Additionally, the Mixture-of-Experts (MoE) approach is adopted in MEASNet \cite{yu2024multi} and WM-MoE \cite{luo2024wm}. We also refer readers to \cite{jiang2024survey} for the comprehensive survey on this topic.
}

Based on the aforementioned analysis for related work, although there are many works introducing the diffusion model to the field of image restoration, there is a limited number of works considering the accumulated score-estimation error by PF characteristics of the diffusion model. Moreover, acceleration of the sampling speed is also an essential research topic for differential equation-based image restoration frameworks.  

\section{Preliminary}
\label{Sec:Preliminary}
In this section, we provide a succinct overview of the diffusion model techniques, acceleration strategies (including trajectory solvers and distillation), and boosting strategies through alignment, laying the foundation for the subsequent sections.

\subsection{Diffusion Model} 
Here, we consider image restoration as a case study to briefly elucidate the diffusion model.

Given a low-quality measurement denoted as $\mathbf{Y}\in \mathbb{R}^{H\times W \times 3}$, image restoration methods strive to reconstruct corresponding high-quality outputs, represented as $\mathbf{X} \in \mathbb{R}^{H' \times W' \times 3}$. The probabilistic nature of this restoration process involves maximizing a posterior $p_{\theta}(\mathbf{X}|\mathbf{Y})$, where $\bm{\theta}$ the parameter set of the learnable neural module. The differential equation-based methods~\cite{kawar2022denoising} generally learn to construct a probabilistic flow (PF), e.g., $P(\mathbf{X}_0|\mathbf{X}_1,\mathbf{Y}),~ P(\mathbf{X}_1|\mathbf{X}_2,\mathbf{Y}), \cdots, P(\mathbf{X}_{N-1}|\mathbf{X}_{N},\mathbf{Y})$, to bridge the marginal distributions $P(\mathbf{X}_0)$ (same as $P(\mathbf{X})$) and $P(\mathbf{X}_N)$, usually as $\mathcal{N}(\mathbf{0}, \sigma_N^2 \mathbf{I})$ with $\sigma$ being the standard deviation of noise. 

In particular, the formulation based on DDPM~\cite{ho2020denoising,saharia2022image} constructs the PF connecting $P(\mathbf{X})$ and a standard Gaussian distribution $\mathcal{N}(\mathbf{0}, \mathbf{I})$, where the forward process is formulated as a diffusion process gradually substituting the data component with Gaussian noise, expressed as $\mathbf{X}_{t} = \Bar{\alpha}_t \mathbf{X}_0 + \sqrt{(1-\Bar{\alpha}_i^2)}\bm{\epsilon}_{t}, \bm{\epsilon}_{t}  \sim \mathcal{N}(\mathbf{0}, \mathbf{I})$.  Conversely, the reverse process is deduced by computing the posterior $P(\mathbf{X}_{t}|\mathbf{X}_{t+1}, \mathbf{X}_0) = \frac{P(\mathbf{X}_{t+1} | \mathbf{X}_{t}) P(\mathbf{X}_t|\mathbf{X}_0)}{P(\mathbf{X}_{t+1}|\mathbf{X}_0)}$ with its resolution derived through the following ancestral sampling process:
\begin{equation}
\small
\begin{aligned}
\label{Eq:DDPM}
\mathbf{X}_{t} \sim \mathcal{N} \Bigg(\frac{1}{\sqrt{\alpha_{t+1}}} 
\Big(\mathbf{X}_{t+1} - &\frac{1-\alpha_{t+1}}{\sqrt{1 - \Bar{\alpha}_{t+1}}}\bm{\epsilon}_{t+1}\Big), \\
& \qquad \frac{1-\Bar{\alpha}_{t}}{1 - \Bar{\alpha}_{t+1}}\left(1-\alpha_{t+1}\right) \mathbf{I} \Bigg).
\end{aligned}
\end{equation}

As the noise component $\bm{\epsilon}_{t+1}$ is typically challenging to handle, it is represented by a neural network parameterization $\bm{\epsilon}_{\bm{\theta}}(\mathbf{X}_{t+1}, t+1)$, trained using a loss term $\mathcal{L}_{DDPM} = \| \bm{\epsilon} - \bm{\epsilon}_{\bm{\theta}}(\Bar{\alpha}_t \mathbf{X}_0 + \sqrt{(1-\Bar{\alpha}_i^2)}\bm{\epsilon}_{t}) \|_2^2$. By iteratively sampling Eq.~\eqref{Eq:DDPM}, we can deduce $\mathbf{X}_0$ from the sample $\mathbf{X}_N$ of a tractable distribution. 
Essentially, these sampling procedures serve to connect samples from two distributions along a high-dimensional trajectory, 
when minimizing the step size from the discrete to continuous spaces, i.e., $\Delta \Bar{\alpha} \rightarrow d \alpha$. 

Score-based models~\cite{song2020score} achieve diffusion model generalization by reformulating the forward process via the following SDE:
\begin{equation}
\label{Eq:ForwardDiff}
    d\mathbf{X} = f(\mathbf{X},t) dt + g(t)d \bm{\omega},
\end{equation}
where $t\in [\delta,~T]$ denotes the timestamp of the diffusion process, with $\delta$ serving as a small number for numerical stability. The reverse SDE can be expressed as 
\begin{equation}
\label{Eq:ReverseSDE}
    d^{(s)}\mathbf{X} = \left[ f(\mathbf{X},t) - g^2(t)\nabla_x \log P(\mathbf{X}) \right] dt + g(t) d \bm{\omega}.
\end{equation}
Furthermore, by redefining Kolmogorov’s forward equation~\cite{song2020score}, an equivalent reverse ODE formulation emerges:
\begin{equation}
\label{Eq:ReverseODE}
    d^{(o)}\mathbf{X} = \left[ f(\mathbf{X},t) - \frac{1}{2} g^2(t)\nabla_x \log P(\mathbf{X}) \right] dt.
\end{equation}

Based on the diffusion strategies for $f(\mathbf{X},t)$ and $g^2(t)$, the score-based diffusion models can be categorized into two variants, namely \textit{variance preserving} (VP) or \textit{variance exploding} (VE). Specifically, VP entails $f(\mathbf{X},t) = \frac{d \log \alpha_t}{d t} \mathbf{X}$ and $g^2(t) = \frac{d \sigma_t^2}{dt} - 2\frac{d \log \alpha_t}{dt} \sigma_t^2$, where $\sigma_t = \sqrt{1-\alpha_t^2}$, while VE involves $f(\mathbf{X},t) = \mathbf{0}$ and $g^2(t) = \frac{d \sigma_t^2}{dt}$.

\subsection{Integral Solver} To derive the reconstruction sample, we can calculate the integral of the reverse ODE trajectory in Eq.~\eqref{Eq:ReverseODE} as $\hat{\mathbf{X}}_0 = \mathbf{X}_T + \int_T^0\left[ f(\mathbf{X},t) - \frac{1}{2} g^2(t)\nabla_x \log \mathbf{P}(\mathbf{X}) \right] dt$. Given the deterministic nature of the entire process, an analytical formulation can be leveraged. For the first time, DPM-Solver~\cite{lu2022dpm,lu2022dpm2} introduces an exact solution based on the semi-linear property of the diffusion model, expressed as
\begin{equation}
    \label{Eq:DPMSolver}
    \mathbf{X}_{t-\Delta t} = \frac{\alpha_{t - \Delta t}}{\alpha_{t}}\mathbf{X}_{t} - \alpha_{t - \Delta t} \int_{\lambda_t}^{\lambda_{t - \Delta t}} e^{-\lambda}  \hat{\bm{\epsilon}}_{\bm{\theta}}(\hat{\mathbf{X}}_\tau,\tau) d \lambda,
\end{equation}
where $\lambda$ denotes the log-SNR (Signal-to-Noise Ratio), i.e., $\lambda := \log( \frac{\alpha_t}{\sigma_t})$. By computing this integral using various Order Taylor series for the non-linear term $\hat{\bm{\epsilon}}_{\bm{\theta}}(\hat{\mathbf{X}}_\lambda,\lambda)$, we can finally derive the result.

\subsection{Rectified Flow \& Trajectory Distillation } As a kind of naturally easily sampled model, the rectified flow-based method learns a straight flow between the random noise and target image domain, whose forward and reverse generation differential equations are identical, formulated as
\begin{equation}
\label{Eq:rectified}
    d^{(o)} \mathbf{X}_t = (\mathbf{X}_0 - \mathbf{X}_T)dt.
\end{equation} It parameterizes a neural network to directly learn the slope $\mathbf{X}_0 - \mathbf{X}_T$ from $\mathbf{X}_t$.
Similarly, distillation-based enhancement~\cite{meng2023distillation,song2023consistency} of diffusion models aims to regularize the inherent trajectory pattern to derive the diffusion models, which can be easily sampled. The nature of such kind of methods is to regularize 
\begin{equation}
\label{eq:Consistency}
   \mathcal{L}_{slope} =  \mathcal{D}(\mathcal{C}(\mathbf{X}_t, t, \bm{\theta}), \mathcal{C}(\mathbf{X}_{s},s, \bm{\phi})),
\end{equation}
where $\mathcal{C}(\cdot)$ denotes a consistency model or the ODE/SDE integrator, solving the integral with outputting $\hat{\mathbf{X}}_0$ from an arbitrary intermediate state $\mathbf{X}_s$ (resp. $\mathbf{X}_t$) with timestamp $s$ (resp. $t$); $\bm{\theta}$ (resp. $\bm{\phi}$ ) represents the model weights of the student (resp. teacher) network;  $\bm{\phi}$ can be exponential moving average between the student net and original pre-trained model; and $\mathcal{D}(\cdot, \cdot)$ indicates the divergence metric for two inputs, such as L1/L2 norm and LPIPS~\cite{zhang2018perceptual}. 

\subsection{Reinforcement Tuning of Diffusion Models}
Given that the diffusion model is trained using discrete trajectory points, the accrued errors during the inference process are typically overlooked. Promising solutions to address these issues are found in reinforcement learning-based methods~\cite{fan2024reinforcement,wallace2024diffusion}, which optimize entire trajectories in a decoupled fashion. Specifically, these methods begin by constructing diffusion trajectories using the reverse SDE formulation of the diffusion model. The quality of the generated samples is then assessed using a reward model. The approach further enhances the probability of high-quality trajectories by aligning the trajectory expectations toward the high-quality points.

\section{Proposed Method}
\label{Sec:Method}

\subsection{Overview}
Learning effective and efficient trajectories is critical for differential equation-based image restoration. \revision{Due to the inherent Markov Chain property of the diffusion model, which leads to an interdependent iterative generation process, optimizing the entire trajectory requires the simultaneous optimization of multiple timestamps, i.e., networks, making it both time- and resource-consuming, and potentially intractable.} 
In this work, we propose a reinforcement learning-inspired alignment process in Sec. \ref{Sec:ODEAlign} for improving restoration ability. Specifically, by projecting the accumulated error back to different steps, we theoretically reason the necessity of adaptively modulating the noise intensity of differential equations. Based on that, we align the ODE trajectory with the most effective alternative drawn from multiple candidate trajectories that are sampled by solving different modulated SDEs. 

Subsequently, in Sec. \ref{Sec:ODEDistillation}, we propose a  cost-aware trajectory distillation strategy to improve efficiency. \revision{This strategy capitalizes on the intrinsic properties of generative models and image restoration tasks to alleviate the burden of distillation. Given that low-quality samples typically reside in the low-probability regions of the target high-quality distribution, we guide the distillation trajectory in the opposite direction using the provided low-quality samples. Furthermore, since low-quality samples correspond to the same scene as their high-quality counterparts, we initiate the generation process from an interpolated state by the given low-quality sample, rather than the pure noise.}
Note that the proposed strategy can be adapted to both score-based~\cite{song2020score} and rectified flow-based~\cite{liu2022flow} diffusion models. Due to page limits, we mainly utilize the formulation of a score-based diffusion model in this paper, and we also
refer readers to the \textit{Supplementary Material} for extensive theoretical elaborations.

\subsection{Reinforcing ODE Trajectories with Modulated SDEs}
\label{Sec:ODEAlign}
Diffusion models are usually trained through individual steps originating from the decoupled PF. However, during the inference phase, they usually operate in a progressively noise-removing manner. Due to inherent score function errors, these models often accumulate inaccuracies. While optimizing the entire trajectory could mitigate this issue, traditional diffusion models, even with ODE-solvers, struggle to yield satisfactory outcomes within a limited number of steps.
Differently, we propose to align the learned trajectories with the most effective alternatives through reinforcement learning. 
Generally, our reinforcement learning-inspired approach~\cite{kaelbling1996reinforcement,wiering2012reinforcement} aims to maximize the expectation of a reward function as
\begin{equation}
  \nabla_{\bm{\theta}} \mathcal{L} = - \nabla_{\bm{\theta}} \mathbb{E}_{x\sim P_{\bm{\theta}}(\mathbf{X})}\mathcal{R}(\mathbf{X}),
\end{equation}
where $\mathcal{R}(\cdot)$ measures the quality of the sample $\mathbf{X}$; $P_{\bm{\theta}}(\mathbf{X})$ illustrates the distribution of $\mathbf{X}$ conditioned on parameter $\bm{\theta}$.
Given the deterministic nature of ODE sampling, optimizing the ODE towards the optimal trajectory could lead to maximizing the likelihood of obtaining the optimal sample \revision{($P_{\bm{\theta}}(\mathbf{X}_{optim}) \rightarrow 1$)}. Unfortunately, the deterministic property of ODE trajectory also makes it difficult to generate diverse trajectories given a randomly initialized starting noise point. It cannot meet the reinforcement learning needs, which requires diverse alternatives to measure and select a better optimization direction. Thus, we argue that a \textit{potential solution} for reinforcement training-based ODE trajectory augmentation should involve leveraging SDE to produce diverse restoration trajectories and aligning the deterministic ODE trajectory with the most effective SDE trajectory. \revision{However, the SDE is formulated with a fixed noise intensity level, which is too rigid and inflexible to adapt to different conditions. As theoretically proven in Sec.~\textcolor{red}{I} of the \textit{Supplementary Material}, for an image restoration trajectory ended with $\mathbf{X}_0$, we need to adjust the intensity of injected noise, conditioned on the reconstruction error $\|\mathbf{X}_0 - \mathbf{X}_0^* \|_2$ and corresponding timestamp $t$, where $\mathbf{X}_0^*$ represents the ground-truth high-quality samples.} Thus, a more flexible and controllable SDE formulation is necessary to reinforce the ODE trajectory.

\revision{Building upon above analyses, we utilize \textit{modulated SDE} (M-SDE), dynamically adjusted by a non-negative factor learned from $\|\mathbf{X}_0 - \mathbf{X}_0^* \|_2$ and timestamp $t$ through a small MLP parameterized with $\bm{\psi}$, denoted as $\gamma_{\bm{\psi}}\left(\|\mathbf{X}_0 - \mathbf{X}_0^* \|_2,~t\right) >0 $.
Serving as a versatile sampling trajectory, M-SDE encompasses ODE, SDE, and DDIM-like sampling, each tailored through distinct parameterizations~\cite{zhang2022fast,zhang2021diffusion}. In what follows, we give the detailed formulations of M-SDE with score-based and rectified flow-based diffusion models.}

\subsubsection{Score-based Diffusion} The inverse M-SDE of the diffusion forward process, as described in Eq. \eqref{Eq:ForwardDiff}, can be explicitly formulated as 
\begin{align}
\label{Eq:reverseMSDE}
\begin{split}
   d^{(\gamma)} \mathbf{X} =& \left[ f(x,t)  - \frac{1 +\gamma_{\bm{\psi}}^2}{2}  g^2(\mathbf{X},t) \nabla_x \log p(\mathbf{X}|\mathbf{Y}) \right] d t \\ &+ \gamma_{\bm{\psi}} g(\mathbf{X},t) d\hat{\bm{\omega}}.
\end{split}
\end{align}
Moreover, the integral solvers for the VP-M-SDE and VE-M-SDE can be respectively written as
\begin{align}
    \label{Eq:VPSolver}
    \mathbf{X}_{t-\Delta t} &= \frac{\alpha_{t-\Delta t}}{\alpha_t} \mathbf{X}_{t} - (1+\gamma^2_{\bm{\psi}})\bm{\epsilon}_{\bm{\theta}} \left(\frac{\alpha_{t-\Delta t}}{\alpha_t} \sigma_t - \sigma_{t-\Delta t}\right)
    \nonumber\\
    &- \sqrt{2}\gamma_{\bm{\psi}}\bm{\epsilon} \alpha_{t-\Delta t}\sqrt{\log \frac{\alpha_{t-\Delta t}}{\alpha_t}}, 
\end{align}
\begin{align}
    \label{Eq:VESolver}
    \mathbf{X}_{t-\Delta t} = & \mathbf{X}_{t} - (1+\gamma^2_{\bm{\psi}})\bm{\epsilon}_{\bm{\theta}} ( \sigma_t - \sigma_{t-\Delta t})
    \nonumber\\
    &- \sqrt{2}\gamma_{\bm{\psi}}\bm{\epsilon} \sqrt{ \sigma_{t}^2 - \sigma_{t-\Delta t}^2}.
\end{align}
We refer readers to Sec.~\textcolor{red}{II} of the \textit{Supplementary Material} for the proof. 

\subsubsection{Rectified Flow-based Diffusion}  The following M-SDE formulation has the same PF as the ODE formulation of the rectified flow-based model in Eq.~\eqref{Eq:rectified}:
\begin{align}
\label{Eq:SDE-RecFlow}
    &  d^{(\gamma)} \mathbf{X} = \mathbf{X}_0 dt + \frac{t-\gamma}{1-t} \mathbf{X}_Tdt +\sqrt{2(\gamma-1)} d \bm{\omega}.
\end{align}
Moreover, the integration formula of Eq.~\eqref{Eq:SDE-RecFlow} can be formulated as 
\begin{align}
    \label{Eq:Rectfied_stochastic_process}
    \mathbf{X}_{t-\Delta_t} =  \frac{\left[\mathbf{X}_{t} - \alpha_t \Delta_t \frac{d\mathbf{X}_t}{dt} - \beta_k\bm{\epsilon}\right]}{(1+\alpha_t\Delta_t-t)+\sqrt{(t-\alpha_t\Delta_t)^2+\beta_k^2}}, 
\end{align}
where $\alpha_t$ is a scalar ($\alpha_t > 1$) and 
\begin{equation}
\label{eq:beta_}
    \beta_k = \sqrt{\frac{(t-\Delta_t)^2\left[1 - (t -\alpha\Delta_t)\right]^2}{[1-(t-\Delta_t)]^2} - \left(t - \alpha \Delta_t\right)^2}.
\end{equation}
We refer readers to Sec.~\textcolor{red}{III} of the \textit{Supplementary Material} for the detailed proof.

With the M-SDE formulation derived, we adopt the following loss to align the ODE trajectory to that of the ``Best-of-$N$" M-SDE at each step:
\begin{equation}
\label{Eq:AlignLoss}
\mathcal{L}_{A}(t_i) = \mathcal{D} \left( \mathbf{X}_{t_i}^{(o)},  \mathbf{X}_{n^*,t_i}\right),
\end{equation}
\revision{where $\mathbf{X}_{t_i}^{(o)} =\mathbf{\Psi}_{t_i}^{\tau} \left(d^{(o)}\mathbf{X}, \mathbf{X}_{\tau}\right)$ denotes a noised state on the reverse ODE trajectory with timestamp $t_i$; $\mathbf{X}_{n^*,t_i} = \mathbf{\Psi}_{t_i}^{\tau}\left(d^{(\gamma)}\mathbf{X}, \mathbf{X}_{\tau}\right)$ is the corresponding noised state from the ``Best-of-$N$" reverse M-SDE trajectory; 
$\mathbf{\Psi}_{t_i}^{\tau} \left(\cdot, \cdot\right)$ denotes the integration from timestamp $\tau$ to $t_i$ through a typical integral solver (e.g., Eqs. \eqref{Eq:DPMSolver}, \eqref{Eq:VPSolver},  \eqref{Eq:VESolver}, or \eqref{Eq:Rectfied_stochastic_process}) with the reverse process formulation of the differential equation (i.e., Eqs. \eqref{Eq:ReverseODE}, \eqref{Eq:rectified}, \eqref{Eq:reverseMSDE}, or \eqref{Eq:SDE-RecFlow} ) and the initial state as the first and second inputs. 
}

\revision{Algorithm~\ref{alg1} summarizes the entire optimization process. Through such a reinforcement learning process, the proposed method can even be trained with some non-differentiable metrics, e.g., NIQE. We refer readers to Sec.~\ref{sec:exponetoone} (resp. \ref{sec:exponetoall}) for the detailed settings of the reward and divergence measurements in task-specific (resp. unified) experiments.}

\revision{Notably, by integrating the reinforcement learning strategy, our method dynamically approximates the ODE trajectories through adaptively sampled high-performance M-SDE trajectories, diverging from conventional reliance on static or predefined solutions. This shift can effectively mitigate overfitting and over-regularization constraints inherent to rigid optimization frameworks, thereby enhancing model generalization capabilities through stochastic trajectory sampling and iterative optimization refinement.}

\begin{algorithm}[t]
\SetKwInOut{KIN}{Input}
\SetKwInOut{KOUT}{Output}
\SetKwInput{KwReturn}{Return}
\caption{\revision{Reinforcing ODE Trajectories with M-SDEs}}
\label{alg1}
\footnotesize
\LinesNumbered
\KIN{diffusion model parameter  $\bm{\theta}$, dataset distribution $P(\mathbf{X}_0)$, the number of SDE candidates $N$, and diffusion integrator $\mathbf{\Psi}_{}^{}(\cdot,\cdot)$.}
\KOUT{ Updated parameter $\bm{\theta}$.}
\label{Line:while}\While{not converged}{
Sampling $\mathbf{X}_0^* \sim P(\mathbf{X}_0)$\\
$\tau\leftarrow\mathrm{Rand}(\{1,2,...,T\})$\\
$\mathbf{X}_T \sim\mathcal{N}(0, \mathbf{I})$;\\
\If{\underline{Score-based diffusion}}{
\tcp*[h]{Calculate noisy states on the reverse ODE trajectory via Eqs. \eqref{Eq:ReverseODE} and \eqref{Eq:DPMSolver}}\\
$\mathbf{X}_{\tau} \leftarrow \mathbf{\Psi}_T^\tau(d^{(o)}\mathbf{X}, \mathbf{X}_T)$;\\
$\left\{\mathbf{X}_{t_i}^{(o)}\right\} \leftarrow \left\{ \mathbf{\Psi}_{t_i}^\tau (d^{(o)} \mathbf{X}, \mathbf{X}_{\tau}), t_i \in [0,\tau]\right\}$;\\
\tcp*[h]{Generate states on the reverse M-SDE trajectories using Eqs.~\eqref{Eq:reverseMSDE} and ~\eqref{Eq:VPSolver} (or \eqref{Eq:VESolver})}\\
\For{n = 1: N}{$\left\{\mathbf{X}_{n,t_i}\right\} \leftarrow \left\{ \mathbf{\Psi}_{t_i,n}^\tau (d^{(\gamma)} \mathbf{X}, \mathbf{X}_{\tau}), t_i \in [0,\tau]\right\}$;\\
}
}
\ElseIf{\underline{Rectified Flow}}{
\tcp*[h]{Calculate the noisy states on the reverse ODE trajectory via Eq.~\eqref{Eq:rectified}}\\
$\mathbf{X}_{\tau} \leftarrow \mathbf{\Psi}_\tau^T (d^{(o)} \mathbf{X}, \mathbf{X}_T)$;\\
$\left\{\mathbf{X}_{t_i}^{(o)}\right\} \leftarrow \left\{ \mathbf{\Psi}_{t_i}^\tau (d^{(o)} \mathbf{X}, \mathbf{X}_{\tau}), t_i \in [0,\tau]\right\}$;\\
\tcp*[h]{Generate noisy state candidates on the reverse M-SDE trajectory using Eqs.~\eqref{Eq:SDE-RecFlow} and~\eqref{Eq:Rectfied_stochastic_process}}\\
\For{n = 1: N}{$\left\{\mathbf{X}_{n,t_i}\right\} \leftarrow \left\{ \mathbf{\Psi}_{t_i,n}^\tau (d^{(\gamma)} \mathbf{X}, \mathbf{X}_{\tau}), t_i \in [0,\tau]\right\}$;}}
 \tcp*[h]{Select the best trajectory via the reward process in reinforcement learning}\\
$n^* \leftarrow \mathop{\arg \max}\limits_{n \in [0,N]} 
 \mathcal{R}(\mathbf{X}_{n,0})$;\\
 \tcp*[h]{Model update based on the loss in Eq.~\eqref{Eq:AlignLoss}}\\
 $\bm{\theta}\leftarrow
  \bm{\theta} + \alpha \nabla_{\bm{\theta}}\left[ \mathcal{D}\left(\mathbf{X}_{t_i}^{(o)},\mathbf{X}_{n^*,t_i}\right)+ \|\mathbf{X}_{t_i}^{(o)} - \mathbf{X}_{0}^*\|_2 \right]$ 
 } 
\KwReturn{$\bm{\theta}$.}
\end{algorithm}
\vspace{-0.4cm}

\subsection{Distillation Cost-aware Diffusion Sampling Acceleration}
\label{Sec:ODEDistillation}
 
In this section, we first explicitly model the cost-value of the diffusion model distillation process. Then, based on both empirical and theoretical results of distillation cost, we propose a novel trajectory distillation pipeline to manage high-quality few-step inference, which consists of a multi-step distillation strategy and a negative guidance policy from low-quality images.

\vspace{0.5em}
\noindent \textbf{Distillation Cost Analysis.} To accelerate diffusion sampling, model distillation~\cite{song2023consistency,liu2023instaflow,liu2022flow} condenses the knowledge from precise and multi-step sampling outcomes into shorter procedures, like the direct regularization in Eq.~\eqref{eq:Consistency}. However, this condensed distillation process may require adjustments to the initial neural parameter distributions, potentially decreasing network performance. To derive efficient and effective reconstruction, we argue that a good distillation training method for diffusion models should not only enable precise integration with fewer inference steps but also leverage the original Neuron-ODEs while minimizing alterations to neural network parameters. To this end, we propose a distillation cost-aware diffusion acceleration strategy that leverages the special characteristics of image restoration tasks to lessen the learning burden of the diffusion network. 

Specifically, to quantify the extent of neural network adjustments, we introduce the trajectory distillation cost defined as
\begin{equation}
\label{Eq:Distillation}
\mathcal{C} = \sum_{i=1}^k \left\| \check{\bm{\bm{\epsilon}}}\left(\frac{\mathbf{X}\big|^{i-1}_{i}}{t \big|^{i-1}_{i}} \Bigg| \frac{d \mathbf{X}_{\bm{\epsilon}}}{d t}\right) -  \epsilon_{\bm{\theta}}( \mathbf{X}_{t_i} ,t_i) \right\|_2,
\end{equation}
where $k$ refers to the total number of steps in the student model, which is also known as the number of function evaluations (NFE), function $\check{\epsilon}(\mathbf{A}| \mathbf{B})$ denotes the inverse of $\bm{\epsilon}(\cdot)$, calculating the corresponding $\bm{\epsilon}$ value via making the $\mathbf{B}$ term identical to the $\mathbf{A}$ term, $\mathbf{X}\big|_i^{i-1} $ symbols $\mathbf{X}_{i-1} - \mathbf{X}_i$. 
We refer readers to Sec.~\textcolor{red}{IV-A} of the \textit{Supplementary Material} for the detailed formulation process of  $\check{\bm{\epsilon}}$ and $\mathcal{C}$.

\begin{figure}[!t]
    \centering
    \vspace{-0.2cm}
    \begin{minipage}{0.15\textwidth}
        \centering
        \includegraphics[width=\textwidth]{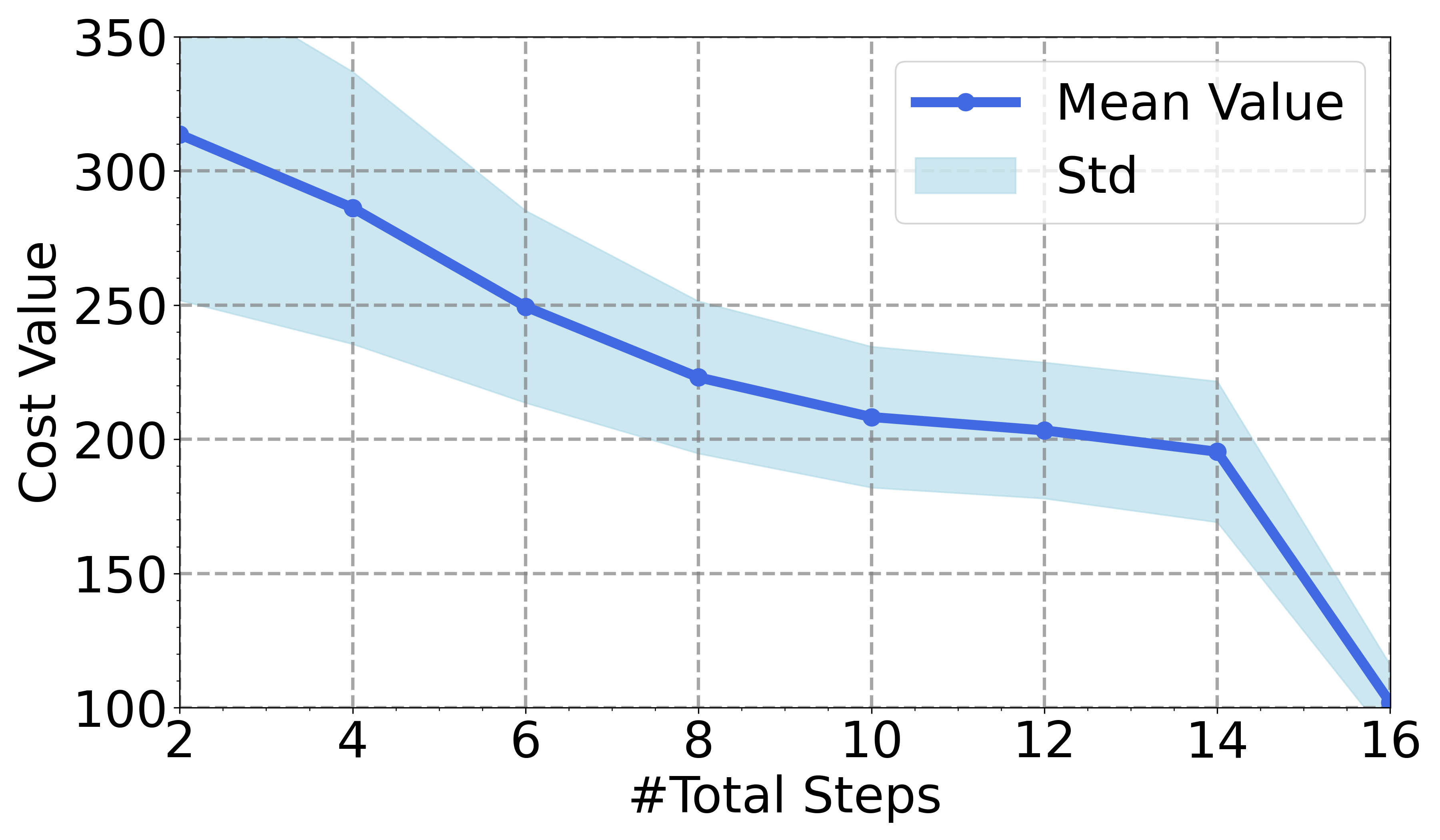} 
        (a)
    \end{minipage}
    \begin{minipage}{0.15\textwidth}
        \centering
        \includegraphics[width=\textwidth]{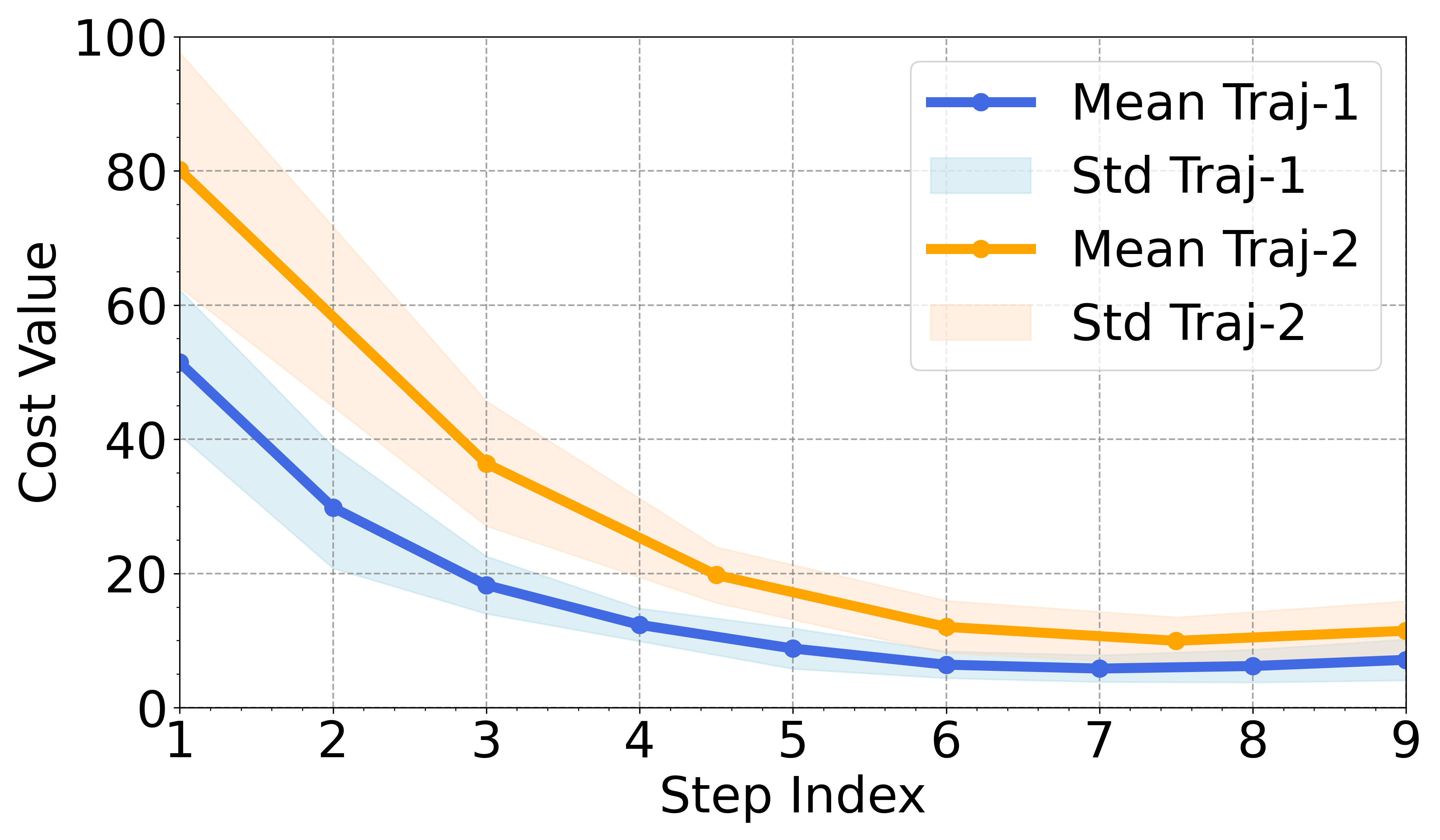} 
        (b)
        \label{fig:image2}
    \end{minipage}
    \begin{minipage}{0.16\textwidth}
        \centering
        \includegraphics[width=0.9\textwidth]{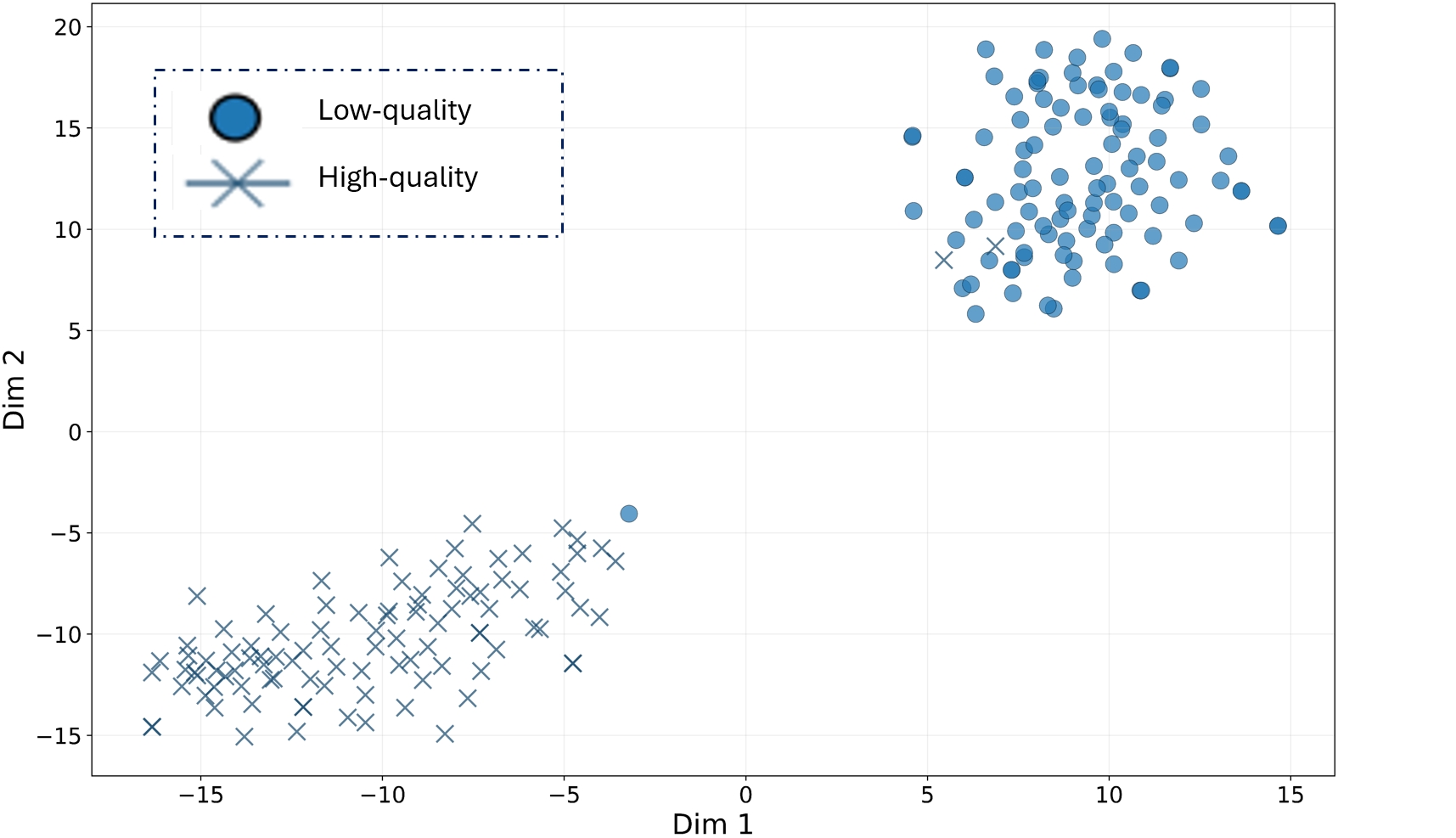} 
        (c)
        \label{fig:image2}
    \end{minipage}
    \vspace{-0.1cm}
    \caption{\label{fig:Cost}\revision{Visualization of distillation costs and data distribution. \textbf{(a)} illustrates the \textit{total distillation cost} (summation across steps) across trajectories with varying numbers of inference steps; \textbf{(b)} provides a detailed illustration of \textit{the distillation cost for each step}. We first derive a dense trajectory with 40 timestamps between timestamp $0$ to $T$. Subsequently, we implement two settings for few steps inference, utilizing step sizes of 5 and 8, resulting in the following timestamp-sparse trajectories, referred to as Traj-1 (8 steps) and Traj-2 (5 steps), respectively. Finally, we compute and visualize the distillation costs between the sparse and dense trajectories; and \textbf{(c)} illustrates the low-dimensional distribution of high-quality and low-quality images, revealing a clear distinction between the two data categories. This visualization highlights the divergent distribution patterns of high-quality and low-quality images. (Due to the page limits, we  zoom out those figures.)}}
\end{figure}

\revision{To investigate the characteristics of the distillation cost of the diffusion model, we calculate both the distillation cost of the whole trajectory and each step and visualize them in Fig.~\ref{fig:Cost} (\textcolor{red}{a}) and (\textcolor{red}{b}). Moreover, we further visualize the distributions of low-quality and high-quality image via t-SNE in Fig.~\ref{fig:Cost} (\textcolor{red}{c}).
With acknowledgment of the aforementioned distillation-based prior, we can draw the following observations:} 
\begin{itemize}
    \item the distillation cost exhibits a negative correlation with the total number of steps~(Fig.~\ref{fig:Cost} (\textcolor{red}{a})); 
    \item initial steps contribute significantly to the overall distillation cost~(Fig.~\ref{fig:Cost} (\textcolor{red}{b}));
    \item \revision{the low-quality images lie in the distinctive distribution compared with the high-quality images, as illustrated in Fig. \ref{fig:Cost} (\textcolor{red}{c})}. 
\end{itemize}
The first observation supports the superiority of recent multi-step distillation models~\cite{ren2024hyper,kim2023consistency,sauer2023adversarial} over their single-step counterparts. Nonetheless, for efficient inference of the diffusion model, the inclusion of few or even single-step models remains essential. Furthermore, in Sec.~\textcolor{red}{IV} of the \textit{Supplementary Material}, we delve into the theoretical exploration of the existence of cost-effective multi-step distillations. Moreover, drawing on the second and third observations, we can outline the subsequent steps to alleviate the substantial learning burden associated with the distillation process.

\begin{algorithm}[t]
\SetKwInOut{KIN}{Input}
\SetKwInOut{KOUT}{Output}
\SetKwInput{KwReturn}{Return}
\caption{\revision{Diffusion Acceleration Distillation}}
\label{alg2:Distill}
\footnotesize
\LinesNumbered
\KIN{Diffusion model parameter $\bm{\theta}$, ODE $d^{(o)}\mathbf{X}$, dataset distribution $P\left(\mathbf{X}_0,\mathbf{Y}\right)$.}
\KOUT{Updated parameter $\bm{\theta}$.}
    \While{not converged}{
    \tcp*[h]{Sampling data pairs and noise, where $\mathbf{X}_0^{\triangledown}$ is the mixture of multi-step results and GT high-quality images.}\\
    $\{\mathbf{X}_0^{\triangledown}, \mathbf{Y}\} \sim P\left(\mathbf{X}_0,\mathbf{Y}\right)$ and $\bm{\epsilon} \sim\mathcal{N}(0, \mathbf{I})$;\\
    \tcp*[h]{Initial state interpolation through Eq.~\eqref{Eq:start_state}}\\
    $\mathbf{X}_{T-\delta} \leftarrow \alpha_{T-\delta} \mathbf{Y} + \sigma_{T-\delta} \bm{\epsilon}$; \\
    \tcp*[h]{Sample generation with negative guidance of Eq.~\eqref{Eq:Guided_Noise}}\\
    $\widehat{\mathbf{X}}_0 \leftarrow \mathbf{\Psi}_{T-\delta}^{0} \left(d^{(o)}\mathbf{X}, \mathbf{X}_{T-\delta}\right)$;\\
    \tcp*[h]{Update via Eq. \eqref{Eq:DistillLoss}}\\
    $\bm{\theta} \leftarrow \bm{\theta} - \nabla_{\bm{\theta}} \mathcal{D} \left(\widehat{\mathbf{X}}_0, \mathbf{X}_0^{\triangledown}\right)$.\\}
\KwReturn{$\bm{\theta}$.}
\end{algorithm}
\begin{algorithm}[t]
\SetKwInOut{KIN}{Input}
\SetKwInOut{KOUT}{Output}
\SetKwInput{KwReturn}{Return}
  \caption{\revision{Inference Process of the Augmented Image Restoration Diffusion Models}}
  \label{alg3:Inference}
\footnotesize
\LinesNumbered
\KIN{diffusion model $\bm{\epsilon}_{\bm{\theta}}(\cdot)$, low quality sample $\mathbf{Y}$.}
\KOUT{ Reconstruction $\widehat{\mathbf{X}}_0$.}
    $\bm{\epsilon} \sim\mathcal{N}(0, \mathbf{I})$\\
    \tcp*[h]{Interpolate the starting state via Eq.~\eqref{Eq:start_state}}\\
    {$\mathbf{X}_{T-\delta} \leftarrow \alpha_{T-\delta} \mathbf{Y} + \sigma_{T-\delta} \bm{\epsilon}$;}\\  
    \For{$t = T-\delta : 0$}{
    \tcp*[h]{Derive the noise prediction with negative guidance of Eq.~\eqref{Eq:Guided_Noise}}\\
    $\hat{\bm{\epsilon}}_\theta \leftarrow (1 + w)\bm{\epsilon}_\theta(\mathbf{X}_t) - w \widetilde{\bm{\epsilon}}$\\
    \tcp*[h]{Calculate the reverse integration via substituting $\hat{\bm{\epsilon}}_\theta$ and $\mathbf{X}_{t}$ into Eq.~\eqref{Eq:DPMSolver}}\\
     $\mathbf{X}_{t-\Delta t} \leftarrow \mathbf{\Psi}_{t-\Delta t}^t(d^{(o)}\mathbf{X}, \mathbf{X}_t)$ ;}
\KwReturn{$\widehat{\mathbf{X}}_0$.}
\end{algorithm}

\subsubsection{Interpolation of the Initial State} Regarding the second observation, during the initial inference stages, diffusion models must synthesize data from pure noise—a challenging yet crucial aspect of the generation process. \revision{Considering the inherent characteristics of the image restoration task, the available low-quality input inherently retains a higher fidelity to the structural and semantic content of the target high-quality reconstruction compared to stochastically sampled Gaussian noise. Consequently, it can be leveraged as a robust preliminary estimation to guide the iterative refinement process toward optimal restoration results.} Thus, to alleviate this burden, we propose synthesizing the noised latent representation as
\begin{equation}
    \label{Eq:start_state}
    \mathbf{X}_{T-\delta} = \alpha_{T-\delta} \mathbf{Y} + \sigma_{T-\delta} \bm{\epsilon},
\end{equation}
where  $\delta \geq 0$ is chosen sufficiently small to  ensure SNR~($\frac{\alpha_{T-\delta}}{\sigma_{T-\delta}}$) to be sufficiently small.  Subsequently, we can train our acceleration distillation neural network via the following loss term:
\begin{align}
\label{Eq:DistillLoss}
\mathcal{L}_{D} = \mathcal{D} &\left( \widehat{\mathbf{X}}_0, \mathbf{X}^{\triangledown}_0 \right),
\end{align}
where $\widehat{\mathbf{X}}_0 = \mathbf{\Psi}_{0}^{T-\delta}\left(d^{(o)}\mathbf{X}, \mathbf{X}_{T-\delta} \right)$ represents reverse integration result from timestamp $T-\delta$ to $0$, and $\mathbf{X}^{\triangledown}_0$ indicates the reference high-quality images, which combines the multi-step generation and known GT images together. Furthermore, we theoretically analyze this noised latent interpolation method in Sec.~\textcolor{red}{V} of \textit{Supplementary Material}.

\subsubsection{Low-quality Images as Sampling Guidance} In the realm of image restoration, our objective is to reconstruct a high-quality image $\mathbf{X}$ from a low-quality measurement $\mathbf{Y}$. Moreover, based on the previous observation, we propose leveraging low-quality images as sampling guidance, which can amplify the positive restoration components from the diffusion model, to ease the learning burden of the diffusion model. From a probabilistic perspective, the training of diffusion-based image restoration models strives to improve the alignment of final reconstruction $\mathbf{X}_0$ with reference image $\mathbf{X}$ under the given condition $\mathbf{Y}$, i.e., improving $\mathbf{P}\left(\mathbf{X}_0 = \mathbf{X} | \mathbf{Y} \right)$. Let $\mathbf{Y} = \mathcal{H}(\mathbf{X})$ with $\mathcal{H}(\cdot)$ being the degradation function. Considering that the degradation Jacobian matrix $\frac{\partial \mathcal{H}(\mathbf{X})}{\partial \mathbf{X}}$  often deviates from the identity matrix, there exists a notable discrepancy between the $\mathbf{P}(\mathbf{X}_0)$ and $\mathbf{P}(\mathbf{Y}_0)$. To address this, we propose a parameterized score function guided by the following principles:
\begin{equation}
    \label{Eq:Guided_Noise}
    \hat{\bm{\epsilon}}_\theta = (1 + w)\bm{\epsilon}_\theta - w \widetilde{\bm{\epsilon}},
\end{equation}
where $\omega$ denotes a scalar for guidance strength, and $\widetilde{\bm{\epsilon}}$ indicates predicted noise by maximizing the posterior likelihood on low-quality images, serving as guidance for the diffusion process, which can be calculated by inverting the integral process:
\begin{equation}
    \widetilde{\bm{\epsilon}} = \frac{\frac{\alpha_0}{\alpha_t}\mathbf{X}_t - \mathbf{Y}}{\frac{\sqrt{1-\alpha_t^2}\alpha_0}{\alpha_t} - \sqrt{ 1 - \alpha_0^2}}.
\end{equation}

Algs.~\ref{alg2:Distill} and \ref{alg3:Inference} summarize the training and inference process in detail, respectively.

\vspace{0.5em}
\noindent \textbf{Remark.} During the distillation phase, we harness the intrinsic characteristics of the image restoration task to mitigate the substantial challenges associated with few-step inference. Specifically, we implement interpolation of the initial state and negative guided sampling to address two critical issues: the significant estimation errors encountered during the high-noise initial stage and the complexities associated with modeling the probabilistic data space. These features distinguish our approach from existing techniques. Besides, our framework aims to improve the overall efficiency and effectiveness of inference processes in complex environments, thereby paving the way for future research and applications.

\vspace{-0.4cm}
\begin{figure*}
    \centering
    \vspace{-0.4cm}
    \includegraphics[width=0.9\linewidth]{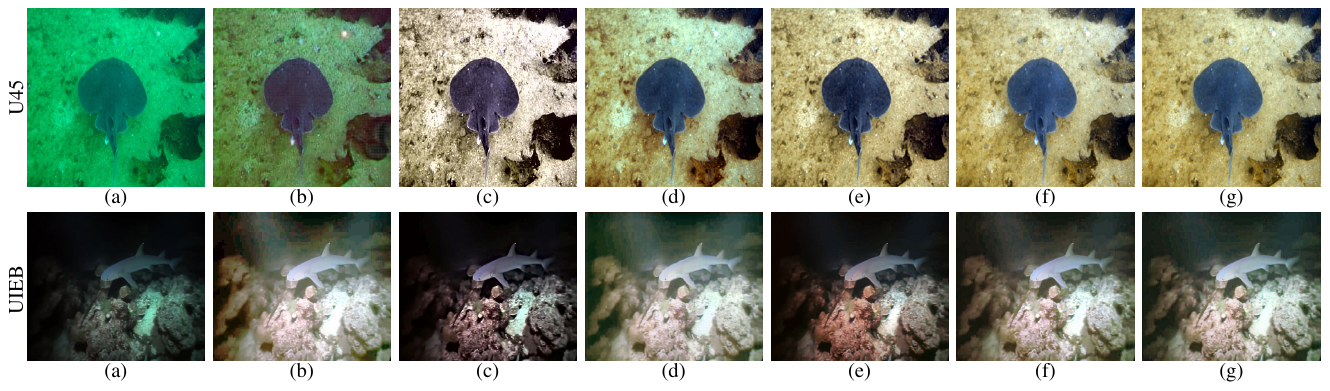}
    \vspace{-0.5cm}
    \caption{Visual comparison of underwater image enhancement on U45~\cite{ancuti2012enhancing} and UIEB~\cite{li2019underwater} datasets. U45 (\textbf{top}): (a) low-quality input, (b) CycleGAN~\cite{zhu2017unpaired}, (c) MLLE~\cite{zhang2022underwater}, (d) HCLR~\cite{zhou2024hclr}, (e) SemiUIR~\cite{huang2023contrastive}, (f) Ours($NFE=1$) and (g) Ours($NFE=10$). UIEB (\textbf{bottom}): except (b) reference image, the remaining columns are the same as those of U45. 
    \revision{More visual results can be found in the \textit{Supplementary Material}}.}
    \label{fig:underwater}
    \vspace{-0.4cm}
\end{figure*}
\begin{table}[!t]
\tiny
  \caption{\label{Table:underwater01}Quantitative comparisons of different methods on \textbf{underwater enhancement}. The best and second-best results are highlighted in \textbf{bold} and \underline{underlined}, respectively.
  ``$\uparrow$" (resp. ``$\downarrow$") means the larger (resp. smaller), the better. 
  ``NFE" denotes the number of function evaluations, which can be interpreted as the inference steps. $\ddagger$ indicates results from the first stage: reinforcing ODE trajectories with modulated SDEs.}
  \vspace{-0.2cm}
  \centering
  \setlength{\tabcolsep}{0.3mm}{
  \renewcommand{\arraystretch}{1.0}
  \begin{tabu}{l|c|cccc|cc|cc}
    \toprule[1.0pt]
    \multirow{2}{*}{Method} &\multirow{2}{*}{NFE}
    &\multicolumn{4}{c}{UIEBD} &\multicolumn{2}{|c|}{C60} &\multicolumn{2}{c}{U45} \\
    \cmidrule(lr){3-6} \cmidrule(lr){7-8} \cmidrule(lr){9-10} 
	&&~~PSNR$\uparrow$~  &~SSIM$\uparrow$~ &~UCIQE$\uparrow$~ &~UIQM$\uparrow$~
    &~UCIQE$\uparrow$~ &~UIQM$\uparrow$~  &~UCIQE$\uparrow$~ &~UIQM$\uparrow$~  \\
    \midrule
    Water-Net \cite{li2019underwater} \tiny{TIP'19}      &1 &16.31 &0.797 &0.606 &2.857 &\underline{0.597} &2.382 &0.599 &2.993 \\
    Ucolor \cite{li2021underwater} \tiny{TIP'21}         &1 &21.09 &0.872 &0.580 &3.048 &0.553 &2.482 &0.573 &3.159 \\
    MLLE \cite{zhang2022underwater}  \tiny{TIP'22}       &1 &19.56 &0.845 &0.588 &2.646 &0.569 &2.208 &0.595 &2.485 \\
    NAFNet \cite{chen2022simple}  \tiny{ECCV'22}         &1 &22.69 &0.870 &0.592 &3.044 &0.559 &2.751 &0.594 &3.087 \\
    Restormer \cite{zamir2022restormer}  \tiny{CVPR'22~}  &1 &23.70 &0.907 &0.599 &3.015 &0.570 &2.688 &0.600 &3.097 \\
    SemiUIR \cite{huang2023contrastive} \tiny{CVPR'23}   &1 &24.31 &0.901 &0.605 &3.032 &0.583 &2.663 &0.606 &3.185 \\
    HCLR-net \cite{zhou2024hclr} \tiny{IJCV'24}          &1 &25.00 &{0.925} &0.607 &3.033 &0.587 &2.695 &0.610 &3.103 \\
    \midrule
    \textbf{Ours}$^\ddagger$ &10 &25.08  &0.913 &0.615 &\textbf{3.142} &0.571 &3.663 &0.612 &4.282  \\ 
    [-1ex]\multicolumn{10}{c}{\hdashrule[0.3ex]{\linewidth}{0.4pt}{2pt 1pt}} \\ [-0.5ex] 
    \textbf{Ours} &2  &\revision{\textbf{26.30}}  &\revision{\textbf{0.939}} &\revision{\textbf{0.624}} &\revision{3.112} &\revision{\textbf{0.610}} &\revision{\textbf{4.004}} &\revision{\textbf{0.636}} &\revision{\textbf{4.527}} \\
    \textbf{Ours} &1  &\underline{26.25}  &\underline{0.938} &\underline{0.623} &\underline{3.135} &0.582 &\underline{3.814} &\underline{0.617} &\underline{4.413} \\
     \bottomrule[1.0pt]
  \end{tabu}}
  \vspace{-0.2cm}
\end{table}

\begin{table}[t]
\tiny
\caption{\revision{Quantitative comparison of recent state-of-the-art methods on \textbf{underwater enhancement} using \textbf{no-reference metrics} and \textbf{computational efficiency analysis}. $\ddagger$ indicates results from the first stage: reinforcing ODE trajectories with modulated SDEs.}}
  \vspace{-0.2cm}
  \label{tab-exp-uie02}
  \centering
  \setlength{\tabcolsep}{0.9mm}{
  \renewcommand{\arraystretch}{1.0}
  \begin{tabu}{l|cccc|ccc}
    \toprule[1.0pt]
    \multirow{2}{*}{Method} 
    &\multicolumn{4}{c}{C60} &\multicolumn{3}{|c}{Complexity} \\
    \cmidrule(lr){2-5}  \cmidrule(lr){6-8} 
    &MUSIQ$\uparrow$ &CLIPIQA$\uparrow$  &MANIQA$\uparrow$  &NIQE$\downarrow$         
    &Param (M) &TFLOPs &Runtime (s) \\
    \midrule
    SemiUIR \cite{huang2023contrastive} \tiny{CVPR'23} &41.768 &0.256 &0.191 &5.379 &1.65 &0.210 &0.13  \\
    HCLR-net \cite{zhou2024hclr} \tiny{IJCV'24}~ &\underline{45.524} &0.331 &0.221 &\underline{4.834} &19.55 &2.099 &0.18  \\
    \midrule
    \textbf{Ours}$^\ddagger$ (NFE=10) &45.514  &\underline{0.350}  &\underline{0.220} &5.055 &68.03 &1.305 &3.17  \\ [-1ex]\multicolumn{8}{c}{\hdashrule[0.3ex]{0.95\linewidth}{0.4pt}{2pt 1pt}} \\ [-0.5ex] 
    \textbf{Ours}~~ (NFE=2) &\textbf{47.019}  &\textbf{0.458} &\textbf{0.293} &\textbf{4.271} &68.03 &1.305 &0.65  \\
    \textbf{Ours}~~ (NFE=1) &44.583  &0.337 &0.205 &4.926 &68.03 &1.305 &0.31  \\
     \bottomrule[1.0pt]
  \end{tabu}}
  \vspace{-0.5cm}
\end{table}

\begin{figure*}
    \centering
    \includegraphics[width=0.9\linewidth]{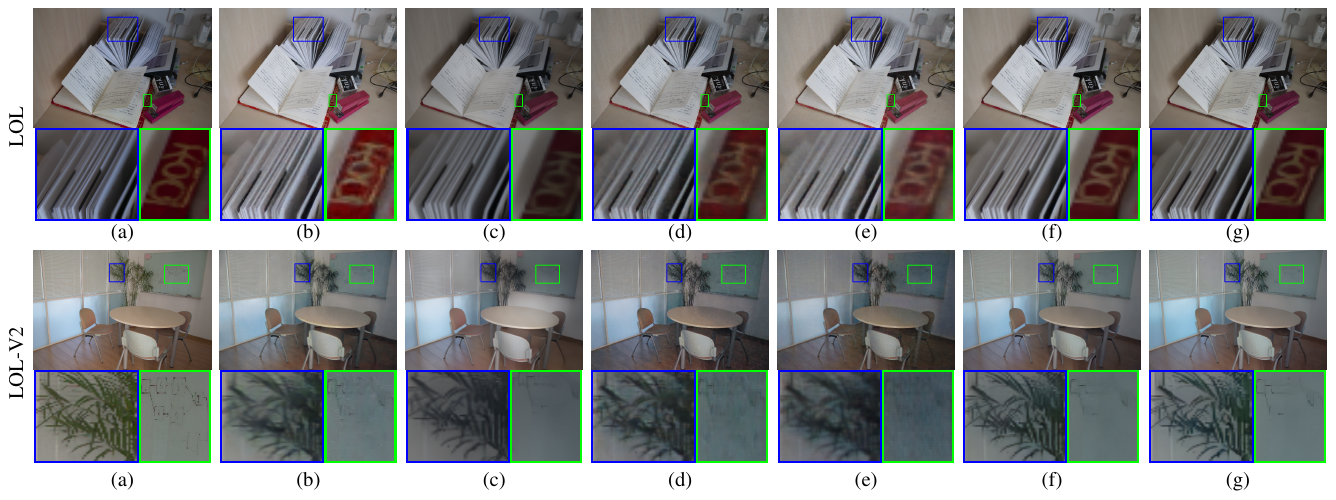}
    \vspace{-0.4cm}
    \caption{Visual comparison of low-light enhancement on LOL and LOLV2 datasets. LOL (\textbf{top}): (a) reference images, (b) CID~\cite{feng2024you}, (c) LLFlow~\cite{wang2022low}, (d) RetinexFormer~\cite{cai2023retinexformer}, (e) LLFormer~\cite{wang2023ultra}, (f) Ours ($NFE=1$), (g) Ours ($NFE=10$). LOLV2 (\textbf{bottom}): except (b) SNR-Aware~\cite{xu2022snr}, the remaining columns are the same as those of LOL. Below each figure, we also visualize zoom-in regions marked by the blue and green boxes. 
    \revision{More visual results can be found in the \textit{Supplementary Material}}.}
    \label{fig:low_light}
\vspace{-0.2cm}
\end{figure*}
\begin{table}[htpb]
\tiny
  \caption{\label{Table:lowlight01} Quantitative comparisons of different methods on \textbf{low-light enhancement}. The best and second-best results are highlighted in \textbf{bold} and \underline{underlined}, respectively.``$\uparrow$" (resp. ``$\downarrow$") means the larger (resp. smaller), the better.``NFE" denotes the number of function evaluations, which can be interpreted as the inference steps. $\ddagger$ indicates results from the first stage: reinforcing ODE trajectories with modulated SDEs.}
  \vspace{-0.2cm}
  \label{tab-exp-lolv1v2}
  \centering
  \setlength{\tabcolsep}{0.3mm}{
  \renewcommand{\arraystretch}{1.0}
  \begin{tabu}{l|c|ccc|c|ccc}
    \toprule[1.0pt]
    \multirow{2}{*}{Method}
    &\multicolumn{4}{c}{LOL}  &\multicolumn{4}{|c}{LOL-v2}  \\
    \cmidrule(lr){2-5} \cmidrule(lr){6-9} 
    &~NFE~ &~~PSNR$\uparrow$~  ~&SSIM$\uparrow$~  &~LPIPS$\downarrow$~
	&~NFE~ &~~PSNR$\uparrow$~  ~&SSIM$\uparrow$~  &~LPIPS$\downarrow$~ \\
    \midrule
    Zero-DCE \cite{guo2020zero} \tiny{CVPR'20} &1 &14.861 &0.562 &0.335   &1 &18.059 &0.580 &0.313    \\
    EnlightenGAN \cite{jiang2021enlightengan} \tiny{TIP'21} &1 &17.483 &0.652 &0.322  &1 &18.640 &0.677 &0.309  \\
    RetinexNet \cite{wei2018deep} \tiny{BMVC'18} &1 &16.770 &0.462 &0.474   &1 &18.371 &0.723 &0.365  \\
    DRBN \cite{yang2020fidelity} \tiny{CVPR'20}  &1 &19.860 &0.834 &0.155   &1 &20.130 &0.830 &0.147  \\
    KinD \cite{zhang2019kindling} \tiny{MM'19}   &1 &20.870 &0.799 &0.207   &1 &17.544 &0.669 &0.375  \\
    KinD++ \cite{zhang2021beyond} \tiny{IJCV'20} &1 &21.300 &0.823 &0.175   &1 &19.087 &0.817 &0.180  \\
    MIRNet \cite{zamir2022learning} \tiny{TPAMI'22} &1 &24.140 &0.842 &0.131 &1 &20.357 &0.782 &0.317  \\
    LLFlow \cite{wang2022low} \tiny{AAAI'22}   &1 &25.132 &0.872 &0.117     &1 &26.200 &0.888 &0.137  \\
    Retinexformer \cite{cai2023retinexformer} \tiny{ICCV'23} &1 &27.180 &0.850 &- &1 &27.710 &0.856 &-   \\
    PyDiff \cite{zhou2023pyramid} \tiny{IJCAI'23} &4 &27.090 &0.879  &0.100    &-  &-  &-  &-   \\
    LLFormer \cite{wang2023ultra} \tiny{AAAI'23} &1 &25.758 &0.823 &0.167   &1 &26.197 &0.819 &0.209  \\
    SNR-Aware \cite{xu2022snr} \tiny{CVPR'22}   &1 &26.716 &0.851 &0.152    &1 &27.209 &0.871 &0.157  \\
    LLFlow-L-SKF++ \cite{wu2024towards} \tiny{TPAMI'24~} &1 &26.894 &0.879 &0.095 &1 &28.453 &\textbf{0.909} &0.117  \\
    GSAD \cite{hou2024global} \tiny{NeurIPS'23}  &20 &27.839 &0.877 &0.091 &10 &28.818 &0.895 &0.095  \\
    \midrule
    \textbf{Ours}$^\ddagger$ &10 &\textbf{28.581} &0.883 &\textbf{0.084} &10 &29.535 &0.898 &\textbf{0.086}   \\ 
    [-1ex]\multicolumn{9}{c}{\hdashrule[0.3ex]{\linewidth}{0.4pt}{2pt 1pt}} \\ [-0.5ex]
    \textbf{Ours} &2  &\underline{28.360} &\textbf{0.886} &0.088  &2 &\revision{\textbf{29.956}} &\revision{\underline{0.905}} &\revision{\underline{0.089}}   \\
    \textbf{Ours} &1  &28.184 &\underline{0.885} &\underline{0.086}  &1 &\underline{29.911} &0.904 &0.101   \\
    \bottomrule[1.0pt]
  \end{tabu}}
\end{table}

\begin{table*}[htpb]
\scriptsize
  \caption{\label{Table:lowlight02} \revision{Quantitative comparison of recent state-of-the-art methods on \textbf{low-light enhancement} using \textbf{no-reference metrics} and \textbf{computational efficiency analysis}. The best and second-best results are highlighted in \textbf{bold} and \underline{underlined}, respectively. $\ddagger$ indicates results from the first stage: reinforcing ODE trajectories with modulated SDEs.}}
  \vspace{-0.2cm}
  \centering
  \setlength{\tabcolsep}{1.3mm}{
  \renewcommand{\arraystretch}{1.0}
  \begin{tabu}{l|cccc|cccc|ccc}
    \toprule[1.0pt]
    \multirow{2}{*}{Method}
    &\multicolumn{4}{c}{LOL}  &\multicolumn{4}{|c}{LOL-v2}  &\multicolumn{3}{|c}{Complexity} \\
    \cmidrule(lr){2-5} \cmidrule(lr){6-9} \cmidrule(lr){10-12}
    &MUSIQ$\uparrow$ &CLIPIQA$\uparrow$  &MANIQA$\uparrow$  &NIQE$\downarrow$
	&MUSIQ$\uparrow$ &CLIPIQA$\uparrow$  &MANIQA$\uparrow$  &NIQE$\downarrow$ 
    &Param (M) &FLOPs (T) &Runtime (s) \\
    \midrule
    LLFlow \cite{wang2022low} \tiny{AAAI'22}   &73.212 &0.380 &0.560 &5.672  &71.001 &0.428 &0.590 &5.566 &37.68 &1.05 &0.58 \\
    SNR-Aware \cite{xu2022snr} \tiny{CVPR'22}   &65.228 &0.254 &0.334 &5.177  &62.507 &0.306 &0.334 &\underline{4.622} &39.13 &0.10 &0.02 \\
    LLFormer \cite{wang2023ultra} \tiny{AAAI'23} &60.760 &0.331 &0.314 &\textbf{3.582}  &52.127 &0.291 &0.265 &\textbf{4.379} &24.55 &0.22 &0.74 \\
    GSAD \cite{hou2024global} \tiny{NeurIPS'23}  &73.555 &0.458 &0.559 &5.805 &\underline{71.721} &\underline{0.552} &\underline{0.595} &5.150 &17.36 &0.27 &0.86 \\
    LLFlow-L-SKF++ \cite{wu2024towards} \tiny{TPAMI'24} &73.874 &0.422 &0.575 &5.549 &70.851 &0.384 &0.568 &5.590 &39.77 &1.10 &0.75 \\
    \midrule
    \textbf{Ours}$^\ddagger$ (NFE=10) &\textbf{74.407} &\underline{0.530} &\textbf{0.604} &5.246 &\textbf{72.586} &0.528 &\textbf{0.637} &5.269 &17.36 &0.27 &0.81 \\
    [-1ex]\multicolumn{12}{c}{\hdashrule[0.3ex]{\linewidth}{0.4pt}{2pt 1pt}} \\ [-0.5ex]
    \textbf{Ours}~ (NFE=2)  &\underline{73.976}  &{0.503} &{0.563} &5.047  &70.578 &\textbf{0.557} &0.565 &4.962 &17.36 &0.27 &0.15 \\
    \textbf{Ours}~ (NFE=1)  &73.966  &\textbf{0.570} &0.569 &\underline{4.817}  &71.633 &0.538 &0.592 &4.961 &17.36 &0.27 &0.06 \\
    \bottomrule[1.0pt]
  \end{tabu}}
\end{table*}

\section{Experiments}
\label{Sec:Exp}
In this section, we thoroughly evaluate the proposed trajectory optimization strategies across various image restoration tasks. Initially, we confirm the task-specific enhancement capabilities by training individual smaller networks for tasks such as de-raining, low-light enhancement, and underwater enhancement in Sec.~\ref{sec:exponetoone}. Furthermore, in Sec.~\ref{sec:exponetoall}, we produce a unified perceptual image restoration network by fine-tuning the state-of-the-art T2I foundational diffusion framework \textit{FLUX-DEV}~\cite{Flux}, which has 12B parameters.

\subsection{Experimental Settings}

\subsubsection{Datasets}\revision{\label{subsubsec:dataset} We employ multiple commonly used benchmark datasets to conduct experiments under two scenarios.}

\revision{
\noindent\textbf{Task-specific restoration.} We conducted three task-specific restoration tasks, including underwater image enhancement, low-light image enhancement, and image deraining. The dataset details are provided as follows:}
\revision{
\begin{itemize}
    \item \textit{\textbf{Underwater Enhancement.}} The UIEB dataset~\cite{li2019underwater} consists of 950 real-world underwater images and includes two subsets: 890 raw underwater images with the corresponding high-quality reference images and 60 unpaired challenging underwater images (labeled as C60). We randomly sampled 800 image pairs for training, and 90 image pairs for testing. We further utilize the unpaired dataset, i.e., C60 and U45 \cite{li2019fusion}, for real-world underwater image enhancement testing.
    \item \textit{\textbf{Low-light Enhancement.}} LOLv1~\cite{wei2018deep} contains 485 low/normal-light image pairs for training and 15 pairs for testing, captured at various exposure times from the real scene. LOLv2~\cite{yang2021sparse} is split into two subsets: LOLv2-real and LOLv2-synthetic. In this work, we utilize the LOLv2-real, which comprises 689 pairs of low-/normal-light images for training and 100 pairs for testing, collected by adjusting the exposure time and ISO.
    \item \textit{\textbf{Image De-raining.}} We utilize several datasets for this task, including Raindrop~\cite{qian2018attentive}, Outdoor-Rain~\cite{jiang2020multi}, Rain100L/H \cite{yang2017deep}, Rain200L/H \cite{yang2017deep}, DID-Data \cite{zhang2018density}, and DDN-Data \cite{fu2017removing}. Specifically, Raindrop contains 861 images for training and 58 images for testing. Outdoor-Rain consists of 8,250 and 750 images for training and testing, respectively. Rain100L/H and Rain200L/H include 1,800 images for training, with 100 images and 200 images for testing, respectively. DID-Data has 12,000 images for training and 1,200 images for testing. DDN-Data comprises 12,600 and 1,400 image pairs for training and testing, respectively.
\end{itemize}
\vspace{0.5em}
\noindent\textbf{Unified restoration.} We collect images for 7 kinds of image restoration tasks. The details of these datasets are listed below:
\begin{itemize}
    \item \textit{\textbf{Image Super-Resolution.}} The training dataset consists of 3,450 high-quality 2K images collected from DIV2K~\cite{Agustsson_2017_CVPR_Workshops} and Flickr2K~\cite{lim2017enhanced}. We utilize the Real-ESRGAN \cite{wang2021real} degradation pipeline to generate LQ-HQ training pairs. The testing set contains 100 images from DIV2K-Val~\cite{Agustsson_2017_CVPR_Workshops}.
    \item \textit{\textbf{Image Desnowing.}} The Snow100K-L~\cite{liu2018desnownet} dataset includes 100k synthesized snowy images with corresponding snow-free reference images and snow masks. We randomly selected 1,872 (resp. 601) images forming the training (resp. testing) dataset for image de-snowing.
    \item \textit{\textbf{Image Deblurring.}} The GoPro~\cite{nah2017deep} dataset contains 3,214 blurred images with a size of $1280\times 720$. The images are divided into 2,103 training images and 1,111 test images. The dataset consists of pairs of a realistic blurry image and the corresponding ground truth sharp images that are obtained by a high-speed camera.
    \item \textit{\textbf{Image Denoising.}} The noisy images were derived by randomly corrupting the aforementioned high-quality SR datasets with Gaussian noise by a standard deviation of $50$. We further utilize BSD500 \cite{arbelaez2010contour} and SIDD \cite{SIDD_2018_CVPR} for generalization ability testing.
    \item \textit{\textbf{Low-light Enhancement.}} We employ LOLv1~\cite{wei2018deep} for fine-tuning  and test.
    \item \textit{\textbf{Image De-raining.}} Raindrop~\cite{qian2018attentive} and Outdoor-rain~\cite{jiang2020multi} datasets are utilized.
    \item \textit{\textbf{Underwater Enhancement.}} The UIEB dataset~\cite{li2019underwater} is employed.
\end{itemize}
}

\subsubsection{Methods under comparison \& Evaluation metrics} On the experiments of specific image restoration tasks, we compared the proposed method with the state-of-the-art methods in the fields of underwater enhancement, low-light enhancement, and deraining. For unified image restoration, we mainly compared the proposed method with the unified method for fairness. Moreover, according to the ill-posed nature of the image restoration inverse problems and to preserve the strong image synthesis capacity of the pre-trained FLUX model, we did not enforce the proposed method to fully approach the reference image and utilized more unpaired and perceptual scores, e.g., NIQE~\cite{mittal2012making}, MUSIQ~\cite{ke2021musiq}, and CLIPIQA~\cite{wang2023exploring}, to validate the performance of our FLUX-IR. 

\subsection{Task-Specific Image Restoration Diffusion Models}
\label{sec:exponetoone}
\noindent\textbf{Implementation Details.} \revision{For task-specific image restoration, e.g., underwater enhancement, low-light enhancement, and deraining, we trained a diffusion model for each particular degradation. Specifically, by using the proposed algorithm, we trained three separate diffusion networks based on GSAD~\cite{hou2024global} for low-light enhancement, de-raining, and underwater enhancement tasks. \revision{During this process, we utilized PSNR (resp.MSE)  as the reward (resp. divergence measurement) for the ODE alignment step, respectively.} The model was trained on an RTX 3090 for 30,000 iterations for both alignment and acceleration, employing the Adam optimizer with a learning rate of $5e^{-5}$, a training patch size of $256\times 256$, and a batch size of $2$.} \secrevision{In each of all task-specific IR experiments, the NFE value is identical during training and inference,
i.e., the models were trained with 10, 2, and 1 steps and also tested with corresponding steps.}

\vspace{0.5em}
\noindent\textbf{Underwater Image Enhancement.} The quantitative comparisons are presented in Table~\ref{Table:underwater01}, demonstrating that the proposed method significantly outperforms state-of-the-art techniques, such as HCLR-net~\cite{zhou2024hclr} and SemiUIR~\cite{huang2023contrastive}, on the UIEBD dataset by \textbf{1.3} dB. The single step model even beats the multi-step counterparts. We visualized the enhanced results in Fig.~\ref{fig:underwater}. For the U45 dataset, due to the fact that there is no ground truth available, we only provided the low-quality input with the corresponding reconstruction. Our method reconstructs more clear details with visually pleasing color, especially for the $1^{st}$ and $3^{rd}$ rows on the U45 dataset. Moreover, on the UIEB dataset, our method may even generate more visually pleasing results than the reference image, shown as the first example. The enhanced image has more soft light, making it easier to distinguish the foreground object, e.g., the shark, and the background scene, e.g., coral. \revision{Additionally, quantitative evaluation on the C60 is provided in Table~\ref{tab-exp-uie02}, where our proposed method achieves substantial improvements over recent state-of-the-art methods across multiple no-reference metrics. Specifically, our two-step model (NFE=2) achieves the highest MUSIQ, CLIPIQA, and MANIQA scores, alongside the lowest NIQE, clearly surpassing SemiUIR~\cite{huang2023contrastive} and HCLR-net~\cite{zhou2024hclr}. Notably, our method also demonstrates competitive computational efficiency, balancing runtime and performance effectively.}

\vspace{0.5em}
\noindent\textbf{Low-light Image Enhancement.}  Table~\ref{Table:lowlight01} presents the performance of various methods in low-light enhancement. It is important to note that \textit{LOL-V1} serves as a highly competitive benchmark. Nonetheless, the proposed method demonstrates a notable improvement of \textbf{0.7} dB. Furthermore, it achieves an enhancement exceeding \textbf{1.1} dB on \textit{LOL-v2}. 
\revision{Table~\ref{Table:lowlight02} reveals the proposed method achieves competitive results across multiple no-reference metrics, such as the highest MUSIQ and MANIQA scores on both LOL and LOL-v2, indicating high perceptual quality. In terms of complexity, the proposed method with NFE=1 demonstrates remarkable efficiency, requiring only \textbf{0.06} seconds, which is significantly faster than other methods, e.g., LLFormer~\cite{wang2023ultra} and LLFlow-L-SKFF++~\cite{wu2024towards}.} 
Fig.~\ref{fig:low_light} visually compares the results of different methods. The proposed method generates more clear details than other compared methods, e.g., the leaves in the $1^{st}$ and $3^{rd}$ examples from the LOL-v2 dataset, and the small word "ano" in $3_{rd}$ examples marked by the green rectangle. Moreover, even for the region with extremely low-light conditions, our method can also accurately reconstruct it, shown as the local zoom in the region with green rectangle in the first example of the LOL dataset.

\begin{figure*}
    \centering
    \includegraphics[width=0.9\linewidth]{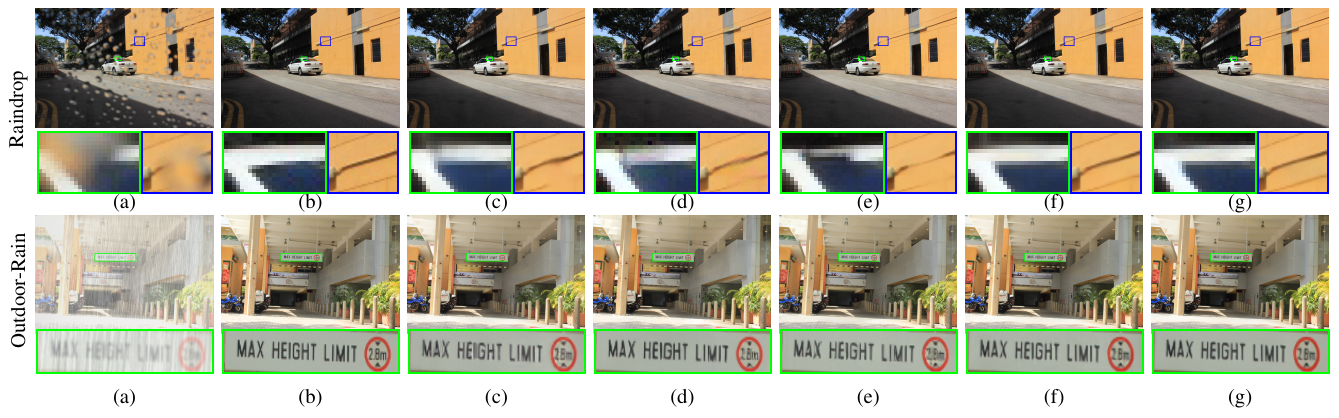} \vspace{-0.35cm}
    \caption{Visual comparison on the tasks of raindrop removal and image deraining. Raindrop removal (\textbf{top}): (a) low-quality input, (b) reference samples, (c) IDT~\cite{xiao2022image}, (d) GridFormer~\cite{wang2024gridformer}, (e) RainDropDiff~\cite{ozdenizci2023restoring}, (f) Ours ($NFE=1$), (g) Ours ($NFE=10$). Deraining (\textbf{bottom}): (a) low-quality input, (b) reference samples, (c) GridFormer~\cite{wang2024gridformer}, (d) WeatherDiff64~\cite{ozdenizci2023restoring}, (e) WeatherDiff128~\cite{ozdenizci2023restoring}, (f) Ours ($NFE=1$), (g) Ours ($NFE=10$). \revision{More visual results can be found in the \textit{Supplementary Material}}.}
    \label{fig:deraining}
\vspace{-0.2cm}
\end{figure*}
\vspace{-0.2cm}
\begin{figure*}[htpb]
  \centering
  \vspace{-0.2cm}
  \begin{minipage}[!t]{0.5\linewidth}
    \scriptsize
    \captionof{table}{Quantitative comparisons of different methods on \textbf{image deraining}. The best and second-best results are highlighted in \textbf{bold} and \underline{underlined}, respectively.``$\uparrow$" (resp. ``$\downarrow$") means the larger (resp. smaller), the better. ``NFE" denotes the number of function evaluations, which can be interpreted as the inference steps. $\ddagger$ indicates results from the first stage: reinforcing ODE trajectories with modulated SDEs.}
      \vspace{-0.2cm}
      \label{tab-exp-outdoorain}
      \centering
      \setlength{\tabcolsep}{1.5mm}{
      \renewcommand{\arraystretch}{1.0}
      \begin{tabu}{l|c|ccc}
        \toprule[1.0pt]
        \multirow{2}{*}{Method} &\multirow{2}{*}{NFE}
        &\multicolumn{3}{c}{Outdoor-Rain}  \\
        \cmidrule(lr){3-5} 
    	&&~~PSNR$\uparrow$~  &~SSIM$\uparrow$~  &~LPIPS$\downarrow$~  \\
        \midrule
        pix2pix \cite{isola2017image} \tiny{CVPR'17}       &1 &19.09 &0.7100 &- \\
        HRGAN \cite{li2019heavy}  \tiny{CVPR'19}           &1 &21.56 &0.8550 &0.154 \\
        PCNet \cite{jiang2021rain} \tiny{TIP'21}           &1 &26.19 &0.9015 &0.132 \\
        MPRNet \cite{zamir2021multi} \tiny{CVPR'21}        &1 &28.03 &0.9192 &0.089 \\
        Restormer \cite{zamir2022restormer} \tiny{CVPR'22} &1 &29.97 &0.9215 &0.074 \\
        \scriptsize{WeatherDiff$_{64}$ \cite{ozdenizci2023restoring}}  \tiny{TPAMI'23} &25 &29.41 &0.9312 &0.059 \\
        \scriptsize{WeatherDiff$_{128}$ \cite{ozdenizci2023restoring}} \tiny{TPAMI'23} &25 &29.28 &0.9216 &0.061 \\
        DTPM \cite{ye2024learning} \tiny{CVPR'24}   &50 &30.48 &0.9210 &\textbf{0.054} \\
        DTPM \cite{ye2024learning} \tiny{CVPR'24}   &10 &30.92 &0.9320 &0.062 \\
        DTPM \cite{ye2024learning} \tiny{CVPR'24}    &4 &30.99 &0.9340 &0.064 \\
        \midrule
        \textbf{Ours}$^\ddagger$ &10  &32.08  &\underline{0.9424} &0.065  \\ 
        [-1ex]\multicolumn{5}{c}{\hdashrule[0.3ex]{0.9\linewidth}{0.4pt}{2pt 1pt}} \\ [-0.5ex]
        \textbf{Ours} &2   &\textbf{33.10}  &\textbf{0.9439} &\underline{0.058}  \\
        \textbf{Ours} &1   &\underline{32.61}  &0.9419 &0.064  \\
         \bottomrule[1.0pt]
      \end{tabu}}
  \end{minipage}
  \hfill
  \begin{minipage}[!t]{0.49\linewidth}
  \scriptsize
    \captionof{table}{Quantitative comparisons of different methods on \textbf{raindrop removing}. The best and second-best results are highlighted in \textbf{bold} and \underline{underlined}, respectively.``$\uparrow$" (resp. ``$\downarrow$") means the larger (resp. smaller), the better. ``NFE" denotes the number of function evaluations, which can be interpreted as the inference steps. $\ddagger$ indicates results from the first stage: reinforcing ODE trajectories with modulated SDEs.}
      \vspace{-0.2cm}
      \label{tab-exp-raindrop}
      \centering
      \setlength{\tabcolsep}{1.5mm}{
      \renewcommand{\arraystretch}{1.0}
      \begin{tabu}{l|c|ccc}
        \toprule[1.0pt]
        \multirow{2}{*}{Method} &\multirow{2}{*}{NFE}
        &\multicolumn{3}{c}{Raindrop}  \\
        \cmidrule(lr){3-5} 
    	&&~~PSNR$\uparrow$~  &~SSIM$\uparrow$~  &~LPIPS$\downarrow$~  \\
        \midrule
        DuRN \cite{liu2019dual} \tiny{CVPR'19}               &1 &31.24 &0.9259 &- \\
        RaindropAttn \cite{quan2019deep} \tiny{ICCV'19}      &1 &31.44 &0.9263 &0.068  \\
        AttentiveGAN \cite{qian2018attentive} \tiny{CVPR'18} &1 &31.59 &0.9170 &0.055  \\
        IDT \cite{xiao2022image} \tiny{TPAMI'22}             &1 &31.87 &0.9313 &0.059 \\
        \scriptsize{RainDropDiff$_{64}$ \cite{ozdenizci2023restoring}} \tiny{TPAMI'23}  &25 &32.29 &0.9422 &0.058 \\
        \scriptsize{RainDropDiff$_{128}$ \cite{ozdenizci2023restoring}} \tiny{TPAMI'23} &25 &32.43 &0.9334 &0.058\\
        AST-B \cite{zhou2024adapt} \tiny{CVPR'24}               &1 &32.38 &0.9350 &0.066 \\
        DTPM \cite{ye2024learning} \tiny{CVPR'24}   &50 &31.44 &0.9320 &\textbf{0.044} \\
        DTPM \cite{ye2024learning} \tiny{CVPR'24}   &10 &31.87 &0.9370 &\underline{0.048} \\
        DTPM \cite{ye2024learning} \tiny{CVPR'24}   &4  &32.72 &0.9440 &0.058 \\
        \midrule
        \textbf{Ours}$^\ddagger$ &10  &33.32  &0.9388   &\textbf{0.044}  \\ 
        [-1ex]\multicolumn{5}{c}{\hdashrule[0.3ex]{0.9\linewidth}{0.4pt}{2pt 1pt}} \\ [-0.5ex]
        \textbf{Ours} &2   &\revision{\textbf{33.67}}    &\revision{\underline{0.9458}} &\revision{0.050}  \\
        \textbf{Ours} &1   &\underline{33.63}  &\textbf{0.9459} &0.052  \\
         \bottomrule[1.0pt]
      \end{tabu}}
  \end{minipage}
  \vspace{0.1cm}
\end{figure*}

\begin{figure*}
    \centering
    \includegraphics[width=0.9\linewidth]{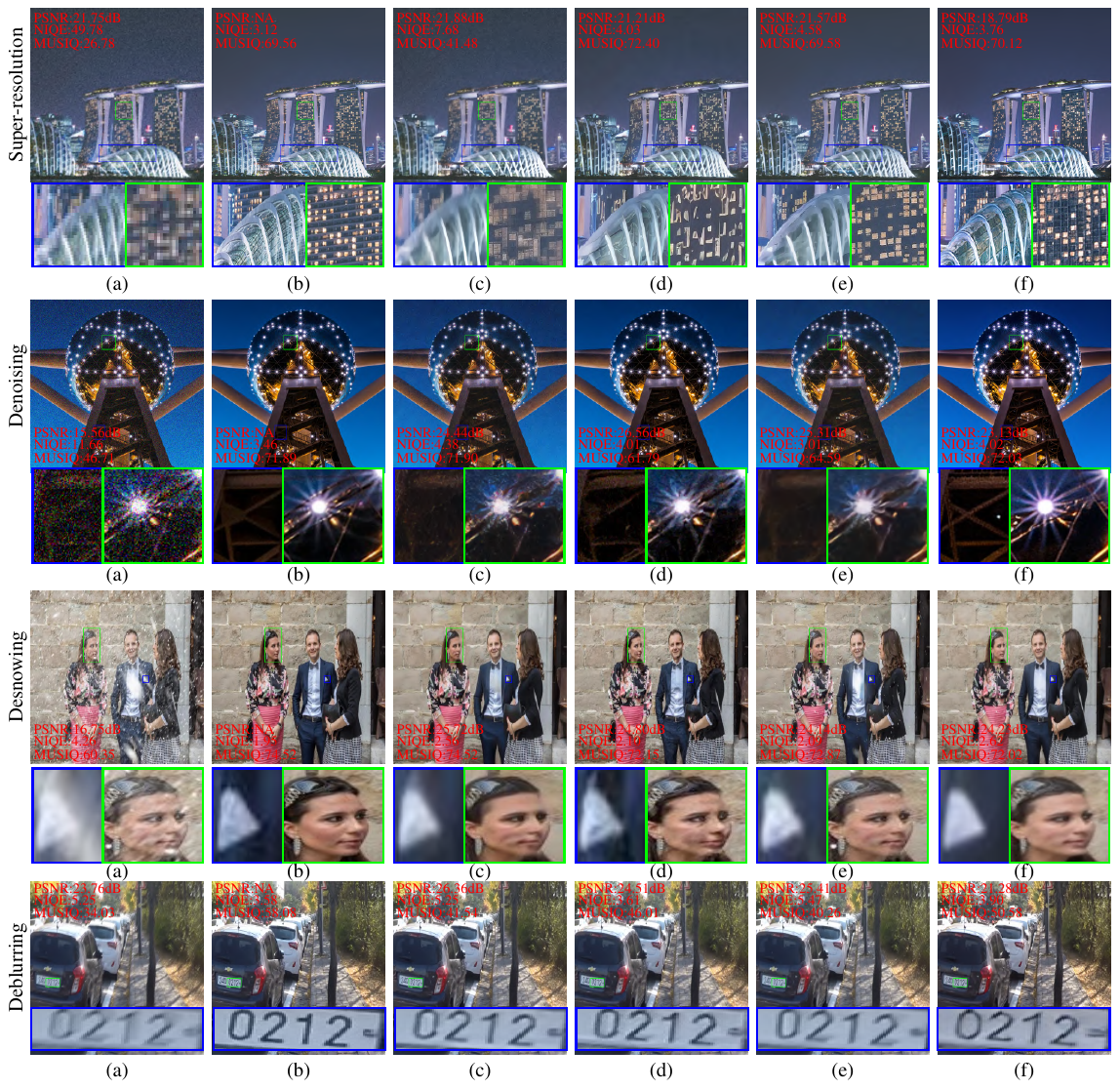}
    \vspace{-1.5em}
    \caption{Visual comparison of unified image restoration, where four different tasks are compared in the figure. In the super-resolution task, (a)-(f) indicates the input low-quality measurement, reference image, PASD~\cite{yang2023pasd}, SeeSR~\cite{wu2024seesr}, and FLUX-IR(Ours), respectively. For denoising, (c)-(e) represents AdaIR~\cite{cui2024adair}, DA-CLIP~\cite{luo2023controlling}, and PromptIR~\cite{potlapalli2306promptir}, respectively. Moreover, for desnowing, (c)-(e) shows WGWS-Net \cite{zhu2023learning}, DA-CLIP~\cite{luo2023controlling}, and DiffUIR~\cite{zheng2024selective}, respectively. Finally for deblurring, (c)-(e) shows AdaIR~\cite{cui2024adair}, DA-CLIP~\cite{luo2023controlling}, and DiffUIR~\cite{zheng2024selective}, respectively. We annotated evaluation metrics of corresponding images by PSNR~$\uparrow$, NIQE~$\downarrow$, and MUSIQ~$\uparrow$, respectively. \revision{More visual results can be found in the \textit{Supplementary Material}}.}
    \label{fig:unified}
  \vspace{-0.5cm}
\end{figure*}

\begin{table*}[h]
  \caption{\revision{Quantitative comparison with state-of-the-art \textbf{transformer-based} and \textbf{diffusion-based} methods on six \textbf{deraining} datasets. The best and second-best results are highlighted in \textbf{bold} and \underline{underlined}, respectively. $\ddagger$ indicates results from the first stage: reinforcing ODE trajectories with modulated SDEs.}}
\label{tab-exp-comparison-deraining-01}
  \vspace{-0.2cm}
  \scriptsize
  \centering
  \resizebox{0.99\textwidth}{!}{
  \setlength{\tabcolsep}{2mm}{
  \renewcommand{\arraystretch}{1.0}
  \begin{tabu}{l|c|cc|cc|cc|cc|cc|cc}
    \toprule[1.0pt]
    \multirow{2}{*}{Method} & \multirow{2}{*}{NFE} & \multicolumn{2}{c|}{Rain100L} & \multicolumn{2}{c|}{Rain100H} & \multicolumn{2}{c|}{Rain200L} & \multicolumn{2}{c|}{Rain200H} & \multicolumn{2}{c|}{DID-Data} & \multicolumn{2}{c}{DDN-Data} \\
     \cmidrule(lr){3-4} \cmidrule(lr){5-6} \cmidrule(lr){7-8} \cmidrule(lr){9-10} \cmidrule(lr){11-12} \cmidrule(lr){13-14}
    ~ & ~ & PSNR$\uparrow$ & SSIM$\uparrow$ & PSNR$\uparrow$ & SSIM$\uparrow$ & PSNR$\uparrow$ & SSIM$\uparrow$ & PSNR$\uparrow$ & SSIM$\uparrow$ & PSNR$\uparrow$ & SSIM$\uparrow$ & PSNR$\uparrow$ & SSIM$\uparrow$ \\

    \midrule
    \textit{Transformer-based:} & & & & & & & & & & & & & \\
    Restormer \cite{zamir2022restormer} \tiny{CVPR'22} &1 &38.99 &0.9782 &31.46 &0.9042 &40.99 &0.9890 &32.00 &0.9329 &35.29 &0.9641 &34.20 &0.9571 \\
    IDT \cite{xiao2022image} \tiny{TPAMI'22} &1 &41.59 &0.9882 &- &- &40.74 &0.9884 &32.10 &\underline{0.9344} &34.89 &0.9623 &33.84 &0.9549 \\
    DRSformer \cite{chen2023learning} \tiny{CVPR'23} &1 &42.50 &0.9902 &33.80 &0.9372 &\underline{41.23} &\underline{0.9894} &\underline{32.17} &0.9326 &35.35 &0.9646 &34.35 &0.9588 \\
    FADformer \cite{gao2024efficient} \tiny{ECCV'24} &1 &42.92 &0.9913 &33.99 &0.9394 &\textbf{41.80} &\textbf{0.9906} &\textbf{32.48} &\textbf{0.9359} &\underline{35.48} &\underline{0.9657} &34.42 &0.9602  \\
    \midrule
    \textit{Diffusion-based:} & & & & & & & & & & & & & \\
    DA-CLIP \cite{luo2023controlling} \tiny{ICLR'24} &100 &41.79 &0.9863 &33.91 &0.9260 &38.63 &0.9772 &28.53 &0.8594 &32.37 &0.9126 &32.16 &0.9241 \\
    GSAD \cite{hou2024global} \tiny{NeurIPS'23} &10 &42.57 &0.9903 &34.19 &0.9472 &39.45 &0.9851 &28.60 &0.9029 &33.38 &0.9431 &34.34 &0.9575 \\ \midrule
    \textbf{Ours}$^\ddagger$ &10 &43.77 &\underline{0.9926} &34.77 &\underline{0.9544} &40.44 &0.9871 &30.49 &0.9217 &35.37 &0.9656 &\textbf{35.19} &\textbf{0.9655} \\ 
    [-1ex]\multicolumn{14}{c}{\hdashrule[0.3ex]{0.95\linewidth}{0.4pt}{2pt 1pt}} \\ [-0.5ex]
    \textbf{Ours} &2 &\textbf{44.38} &\textbf{0.9931} &\textbf{35.80} &\textbf{0.9562} &40.51 &0.9871 &30.33 &0.9181 &\textbf{35.50} &\textbf{0.9662} &\underline{35.16} &\underline{0.9650} \\
    \textbf{Ours} &1 &\underline{43.79} &0.9922 &\underline{35.30} &0.9518 &40.29 &0.9863 &30.02 &0.9145 &35.22 &0.9647 &34.94 &0.9631 \\
    \bottomrule[1.0pt]
  \end{tabu}}}
  \vspace{-0.5cm}
\end{table*}

\begin{table}[!t]
\tiny
\captionof{table}{\revision{Quantitative comparison of recent state-of-the-art methods on \textbf{raindrop removing} using \textbf{no-reference metrics} and \textbf{computational efficiency analysis}. The best and second-best results are highlighted in \textbf{bold} and \underline{underlined}, respectively. $\ddagger$ indicates results from the first stage: reinforcing ODE trajectories with modulated SDEs}}
  \vspace{-0.2cm}
  \label{tab-exp-raindrop02}
  \centering
  \setlength{\tabcolsep}{0.5mm}{
  \renewcommand{\arraystretch}{1.0}
  \begin{tabu}{l|cccc|ccc}
    \toprule[1.0pt]
    \multirow{2}{*}{Method} 
    &\multicolumn{4}{c}{Raindrop} &\multicolumn{3}{|c}{Complexity} \\
    \cmidrule(lr){2-5}  \cmidrule(lr){6-8} 
    &MUSIQ$\uparrow$ &CLIPIQA$\uparrow$  &MANIQA$\uparrow$  &NIQE$\downarrow$         
    &Param (M) &FLOPs (T) &Runtime (s) \\
    \midrule
    \tiny{RainDropDiff$_{128}$ \cite{ozdenizci2023restoring}} \tiny{TPAMI'23} &\underline{71.631} &0.471 &\underline{0.501} &3.517  &82.96 &1.921 &103.70  \\
    AST-B \cite{zhou2024adapt} \tiny{CVPR'24} &69.751 &0.427 &0.463 &\textbf{3.272} &19.92 &0.152 &0.10  \\
    GridFormer \cite{wang2024gridformer} \tiny{IJCV'24} &69.829 &\underline{0.548} &0.486 &3.684 &30.20 &1.010 &0.87  \\
    \midrule
    \textbf{Ours}$^\ddagger$ (NFE=10) &70.464  &\textbf{0.555}  &0.472 &3.710 &68.03 &1.067 &1.52  \\ [-1ex]\multicolumn{8}{c}{\hdashrule[0.3ex]{\linewidth}{0.4pt}{2pt 1pt}} \\ [-0.5ex]
    \textbf{Ours}~~ (NFE=2) &\textbf{71.872}  &0.504 &\textbf{0.505} &3.535 &68.03 &1.067 &0.32  \\ 
    \textbf{Ours}~~ (NFE=1) &71.365  &0.475 &0.488 &3.545 &68.03 &1.067 &0.13  \\
     \bottomrule[1.0pt]
  \end{tabu}}
\end{table}

\begin{table*}[h]
\centering
\scriptsize
\caption{\revision{Quantitative comparisons of different methods on unified IR tasks. ``$\uparrow$" (resp. ``$\downarrow$") means the larger (resp. smaller), the better. $\dag$ \textbf{means the compared methods were designed as unified image restoration frameworks}.
}}\vspace{-0.2cm}
\label{Table:Flux-IR} 
\resizebox{\textwidth}{!}{
\setlength{\tabcolsep}{0.5mm}
\renewcommand{\arraystretch}{1.0}
\begin{tabular}{l|c|c|c|c|c|c|c|c|c}
\toprule[1.0pt]
\multirow{4}{*}{Method} & \multirow{4}{*}{NFE} 
& \multicolumn{1}{c|}{\begin{tabular}[c]{@{}c@{}}\textbf{Super-resolution}\\(DIV2K-Val \cite{Agustsson_2017_CVPR_Workshops})\end{tabular}} 
& \multicolumn{1}{c|}{\begin{tabular}[c]{@{}c@{}}\textbf{Desnowing}\\(Snow100K-L \cite{liu2018desnownet})\end{tabular}} 
& \multicolumn{1}{c|}{\begin{tabular}[c]{@{}c@{}}\textbf{De-blurring}\\(GoPro \cite{nah2017deep})\end{tabular}} 
& \multicolumn{1}{c|}{\begin{tabular}[c]{@{}c@{}}\textbf{Denoising}\\(DIV2K-Val \cite{Agustsson_2017_CVPR_Workshops})\end{tabular}} 
& \multicolumn{1}{c|}{\begin{tabular}[c]{@{}c@{}}\textbf{Low-Light Enhancement}\\(LOL \cite{wei2018deep})\end{tabular}} 
& \multicolumn{1}{c|}{\begin{tabular}[c]{@{}c@{}}\textbf{Raindrop removal}\\(Raindrop \cite{qian2018attentive})\end{tabular}} 
& \multicolumn{1}{c|}{\begin{tabular}[c]{@{}c@{}}\textbf{Deraining\&Dehazing}\\(Outdoor-rain \cite{jiang2020multi})\end{tabular}} 
& \multicolumn{1}{c}{\begin{tabular}[c]{@{}c@{}}\textbf{Underwater Enhancement}\\(UIEBD \cite{li2019underwater})\end{tabular}} \\
\cmidrule(lr){3-3} \cmidrule(lr){4-4} \cmidrule(lr){5-5}
\cmidrule(lr){6-6} \cmidrule(lr){7-7} \cmidrule(lr){8-8}
\cmidrule(lr){9-9} \cmidrule(lr){10-10}
& 
& \begin{tabular}[c]{@{}c@{}}NIQE$\downarrow$~/~MUSIQ$\uparrow$\\CLIPIQA$\uparrow$~/~PSNR$\uparrow$\end{tabular}
& \begin{tabular}[c]{@{}c@{}}NIQE$\downarrow$~/~MUSIQ$\uparrow$\\CLIPIQA$\uparrow$~/~PSNR$\uparrow$\end{tabular}
& \begin{tabular}[c]{@{}c@{}}NIQE$\downarrow$~/~MUSIQ$\uparrow$\\CLIPIQA$\uparrow$~/~PSNR$\uparrow$\end{tabular}
& \begin{tabular}[c]{@{}c@{}}NIQE$\downarrow$~/~MUSIQ$\uparrow$\\CLIPIQA$\uparrow$~/~PSNR$\uparrow$\end{tabular}
& \begin{tabular}[c]{@{}c@{}}NIQE$\downarrow$~/~MUSIQ$\uparrow$\\CLIPIQA$\uparrow$~/~PSNR$\uparrow$\end{tabular}
& \begin{tabular}[c]{@{}c@{}}NIQE$\downarrow$~/~MUSIQ$\uparrow$\\CLIPIQA$\uparrow$~/~PSNR$\uparrow$\end{tabular}
& \begin{tabular}[c]{@{}c@{}}NIQE$\downarrow$~/~MUSIQ$\uparrow$\\CLIPIQA$\uparrow$~/~PSNR$\uparrow$\end{tabular}
& \begin{tabular}[c]{@{}c@{}}NIQE$\downarrow$~/~MUSIQ$\uparrow$\\CLIPIQA$\uparrow$~/~PSNR$\uparrow$\end{tabular}
\\
\midrule
SeeSR \cite{wu2024seesr} \tiny{CVPR'24} &50
& \begin{tabular}[c]{@{}c@{}}4.073~/~72.437\\0.722~/~23.329\end{tabular}
& \XSolidBrush
& \XSolidBrush
& \XSolidBrush
& \XSolidBrush
& \XSolidBrush
& \XSolidBrush
& \XSolidBrush
\\
\midrule
PASD \cite{yang2023pasd} \tiny{ECCV'24} &20
& \begin{tabular}[c]{@{}c@{}}3.676~/~71.738\\0.695~/~23.143\end{tabular}
& \XSolidBrush
& \XSolidBrush
& \XSolidBrush
& \XSolidBrush
& \XSolidBrush
& \XSolidBrush
& \XSolidBrush
\\
\midrule
DiffBIR$^\dag$ \cite{lin2023diffbir} \tiny{ECCV'24} &50
& \begin{tabular}[c]{@{}c@{}}4.333~/~64.672\\0.651~/~22.548\end{tabular}
& \XSolidBrush
& \XSolidBrush
& \XSolidBrush
& \XSolidBrush
& \XSolidBrush
& \XSolidBrush
& \XSolidBrush
\\
\midrule
DA-CLIP$^\dag$ \cite{luo2023controlling} \tiny{ICLR'24} &100
& \XSolidBrush
& \begin{tabular}[c]{@{}c@{}}2.775~/~69.394\\0.375~/~28.641\end{tabular}
& \begin{tabular}[c]{@{}c@{}}3.937~/~39.925\\0.212~/~28.619\end{tabular}
& \begin{tabular}[c]{@{}c@{}}5.091~/~58.098\\0.584~/~26.979\end{tabular}
& \begin{tabular}[c]{@{}c@{}}5.208~/~70.500\\0.633~/~26.768\end{tabular}
& \begin{tabular}[c]{@{}c@{}}4.817~/~67.592\\0.488~/~31.207\end{tabular}
& \XSolidBrush
& \XSolidBrush
\\
\midrule
DiffUIR$^\dag$ \cite{zheng2024selective} \tiny{CVPR'24} &3
& \XSolidBrush
& \begin{tabular}[c]{@{}c@{}}3.163~/~70.265\\0.456~/~28.879\end{tabular}
& \begin{tabular}[c]{@{}c@{}}5.834~/~34.061\\0.203~/~27.815\end{tabular}
& \XSolidBrush
& \begin{tabular}[c]{@{}c@{}}4.904~/~71.389\\0.378~/~25.269\end{tabular}
& \XSolidBrush
& \XSolidBrush
& \XSolidBrush
\\
\midrule
AdaIR$^\dag$ \cite{cui2024adair} \tiny{ICLR'25} &1
& \XSolidBrush
& \XSolidBrush
& \begin{tabular}[c]{@{}c@{}}5.514~/~33.263\\0.198~/~28.464\end{tabular}
& \begin{tabular}[c]{@{}c@{}}4.458~/~59.714\\0.611~/~25.953\end{tabular}
& \begin{tabular}[c]{@{}c@{}}4.713~/~70.859\\0.404~/~22.409\end{tabular}
& \XSolidBrush
& \XSolidBrush
& \XSolidBrush
\\
\midrule
AutoDIR$^\dag$ \cite{jiang2023autodir} \tiny{ECCV'24} &50
& \begin{tabular}[c]{@{}c@{}}4.865~/~55.196\\0.458~/~23.939\end{tabular}
& \XSolidBrush
& \begin{tabular}[c]{@{}c@{}}6.164~/~33.354\\0.197~/~28.444\end{tabular}
& \begin{tabular}[c]{@{}c@{}}5.095~/~58.399\\0.515~/~28.081\end{tabular}
& \begin{tabular}[c]{@{}c@{}}4.200~/~71.095\\0.398~/~22.896\end{tabular}
& \begin{tabular}[c]{@{}c@{}}3.365~/~68.723\\0.403~/~32.332\end{tabular}
& \XSolidBrush
& \XSolidBrush
\\
\midrule
WeatherDiff \cite{ozdenizci2023restoring} \tiny{TPAMI'23} &25
& \XSolidBrush
& \XSolidBrush
& \XSolidBrush
& \XSolidBrush
& \XSolidBrush
& \begin{tabular}[c]{@{}c@{}}3.517~/~71.631\\0.471~/~30.713\end{tabular}
& \begin{tabular}[c]{@{}c@{}}3.544~/~71.397\\0.501~/~29.721\end{tabular}
& \XSolidBrush
\\
\midrule
AST-B \cite{zhou2024adapt} \tiny{CVPR'24} &1
& \XSolidBrush
& \XSolidBrush
& \XSolidBrush
& \XSolidBrush
& \XSolidBrush
& \begin{tabular}[c]{@{}c@{}}3.272~/~69.750\\0.427~/~32.380\end{tabular}
& \XSolidBrush
& \XSolidBrush
\\
\midrule
GridFormer \cite{wang2024gridformer} \tiny{IJCV'24} &1
& \XSolidBrush
& \begin{tabular}[c]{@{}c@{}}2.918~/~70.854\\0.488~/~30.792\end{tabular}
& \XSolidBrush
& \XSolidBrush
& \XSolidBrush
& \begin{tabular}[c]{@{}c@{}}3.684~/~69.829\\0.558~/~32.324\end{tabular}
& \begin{tabular}[c]{@{}c@{}}3.620~/~70.299\\0.511~/~31.874\end{tabular}
& \XSolidBrush
\\
\midrule
WGWS-Net \cite{zhu2023learning} \tiny{CVPR'23}~ &1
& \XSolidBrush
& \begin{tabular}[c]{@{}c@{}}3.345~/~70.167\\0.507~/~28.933\end{tabular}
& \XSolidBrush
& \XSolidBrush
& \XSolidBrush
& \begin{tabular}[c]{@{}c@{}}3.479~/~71.731\\0.435~/~33.430\end{tabular}
& \begin{tabular}[c]{@{}c@{}}3.968~/~70.835\\0.459~/~30.609\end{tabular}
& \XSolidBrush
\\
\midrule
NU$^2$Net \cite{guo2023underwater} \tiny{AAAI'23} &1
& \XSolidBrush
& \XSolidBrush
& \XSolidBrush
& \XSolidBrush
& \XSolidBrush
& \XSolidBrush
& \XSolidBrush
& \begin{tabular}[c]{@{}c@{}}4.717~/~47.810\\0.541~/~25.221\end{tabular}
\\
\midrule
HCLR-net \cite{zhou2024hclr} \tiny{IJCV'24} &1
& \XSolidBrush
& \XSolidBrush
& \XSolidBrush
& \XSolidBrush
& \XSolidBrush
& \XSolidBrush
& \XSolidBrush
& \begin{tabular}[c]{@{}c@{}}4.803~/~48.623\\0.593~/~24.998\end{tabular}
\\
\midrule
FLUX-IR (Ours) &20
& \begin{tabular}[c]{@{}c@{}}3.491~/~73.188\\0.721~/~20.248\end{tabular}
& \begin{tabular}[c]{@{}c@{}}2.769~/~65.985\\0.440~/~23.442\end{tabular}
& \begin{tabular}[c]{@{}c@{}}4.578~/~46.728\\0.233~/~23.884\end{tabular}
& \begin{tabular}[c]{@{}c@{}}3.631~/~72.657\\0.685~/~24.531\end{tabular}
& \begin{tabular}[c]{@{}c@{}}4.045~/~73.233\\0.484~/~25.029\end{tabular}
& \begin{tabular}[c]{@{}c@{}}3.201~/~70.105\\0.507~/~25.846\end{tabular}
& \begin{tabular}[c]{@{}c@{}}2.804~/~70.631\\0.514~/~24.734\end{tabular}
& \begin{tabular}[c]{@{}c@{}}4.515~/~49.867\\0.574~/~23.275\end{tabular}
\\ \midrule
FLUX-IR (Ours) &10
& \begin{tabular}[c]{@{}c@{}}4.269~/~72.069\\0.673~/~20.895\end{tabular}
& \begin{tabular}[c]{@{}c@{}}2.862~/~68.117\\0.482~/~24.840\end{tabular}
& \begin{tabular}[c]{@{}c@{}}4.521~/~42.045\\0.199~/~24.868\end{tabular}
& \begin{tabular}[c]{@{}c@{}}4.199~/~71.519\\0.631~/~25.373\end{tabular}
& \begin{tabular}[c]{@{}c@{}}4.163~/~72.728\\0.460~/~24.914\end{tabular}
& \begin{tabular}[c]{@{}c@{}}3.320~/~69.661\\0.515~/~27.918\end{tabular}
& \begin{tabular}[c]{@{}c@{}}3.023~/~69.862\\0.521~/~25.856\end{tabular}
& \begin{tabular}[c]{@{}c@{}}4.596~/~49.958\\0.574~/~23.409\end{tabular}
\\
\bottomrule[1.0pt]
\end{tabular}
}
\vspace{-0.5cm}
\end{table*}

\begin{table}[h]
\centering
\tiny
\caption{\revision{Quantitative comparison of the \textbf{generalization ability} of different unified image restoration methods on image denoising. Note that our training set does not include the BSD500 and SIDD datasets, where SIDD is a real-world denoising dataset. ``$\uparrow$" (resp. ``$\downarrow$") means the larger (resp. smaller), the better.
  }}\vspace{-0.2cm}
\label{Table:Flux-IR-ga} 
\resizebox{0.47\textwidth}{!}{
\setlength{\tabcolsep}{0.5mm}
\renewcommand{\arraystretch}{1.0}
\begin{tabular}{l|c|c|c|c}
\toprule[1.0pt]
\multirow{4}{*}{Method} & \multirow{4}{*}{NFE} 
&\multicolumn{2}{c|}{BSD500-Test \cite{arbelaez2010contour}} & \multirow{2}{*}{SIDD \cite{SIDD_2018_CVPR}} \\
& & \multicolumn{1}{c}{$\sigma=30$} & \multicolumn{1}{c|}{$\sigma=50$}  & \\
\cmidrule(lr){3-3} \cmidrule(lr){4-4} \cmidrule(lr){5-5}
&
& \begin{tabular}[c]{@{}c@{}}NIQE$\downarrow$/~MUSIQ$\uparrow$\\CLIPIQA$\uparrow$/~PSNR$\uparrow$\end{tabular}
& \begin{tabular}[c]{@{}c@{}}NIQE$\downarrow$/~MUSIQ$\uparrow$\\CLIPIQA$\uparrow$/~PSNR$\uparrow$\end{tabular}
& \begin{tabular}[c]{@{}c@{}}NIQE$\downarrow$/~MUSIQ$\uparrow$\\CLIPIQA$\uparrow$/~MANIQA$\uparrow$\end{tabular}

\\
\midrule
PromptIR \cite{potlapalli2306promptir} \tiny{NeurIPS'23}
&1
& \begin{tabular}[c]{@{}c@{}}10.724~/~29.481\\~\textbf{0.701}~/~26.214\end{tabular}
& \begin{tabular}[c]{@{}c@{}}12.797~/~37.948\\~\textbf{0.782}~/~22.211\end{tabular}
& \begin{tabular}[c]{@{}c@{}}\textbf{5.143}~/~44.790\\0.449~/~0.307\end{tabular}
\\
\midrule
AdaIR \cite{cui2024adair} \tiny{ICLR'25} 
&1 
& \begin{tabular}[c]{@{}c@{}}11.234~/~28.106\\~\underline{0.677}~/~25.820\end{tabular}
& \begin{tabular}[c]{@{}c@{}}12.513~/~37.451\\~\underline{0.766}~/~22.178\end{tabular}
& \begin{tabular}[c]{@{}c@{}}\underline{5.577}~/~44.719\\~\underline{0.481}~/~0.347\end{tabular}
\\
\midrule
DA-CLIP \cite{luo2023controlling} \tiny{ICLR'24}

&100
& \begin{tabular}[c]{@{}c@{}}10.972~/~29.167\\0.340~/~25.729\end{tabular}
& \begin{tabular}[c]{@{}c@{}}9.893~/~27.994\\0.485~/~23.819\end{tabular}
& \begin{tabular}[c]{@{}c@{}}8.197~/~43.735\\0.383~/~0.240\end{tabular}
\\
\midrule
AutoDIR \cite{jiang2023autodir} \tiny{ECCV'24}
&50 
& \begin{tabular}[c]{@{}c@{}}5.866~/~47.230\\0.452~/~\textbf{33.065}\end{tabular}
& \begin{tabular}[c]{@{}c@{}}\underline{4.959}~/~46.304\\0.513~/~\textbf{31.670}\end{tabular}
& \begin{tabular}[c]{@{}c@{}}9.414~/~42.997\\0.351~/~0.295\end{tabular}
\\
\midrule
FLUX-IR (Ours) &20 
& \begin{tabular}[c]{@{}c@{}}\textbf{4.942}~/~\textbf{65.967}\\0.527~/~27.027\end{tabular}
& \begin{tabular}[c]{@{}c@{}}\textbf{4.923}~/~\textbf{63.886}\\0.547~/~26.472\end{tabular}
& \begin{tabular}[c]{@{}c@{}}7.655~/~\textbf{62.950}\\~\textbf{0.505}~/~\textbf{0.427}\end{tabular}
\\
\midrule
FLUX-IR (Ours) & 10 
& \begin{tabular}[c]{@{}c@{}}\underline{5.826}~/~\underline{60.572}\\0.447~/~\underline{28.521}\end{tabular}
& \begin{tabular}[c]{@{}c@{}}5.894~/~\underline{58.529}\\0.482~/~\underline{27.021}\end{tabular}
& \begin{tabular}[c]{@{}c@{}}7.686~/~\underline{58.661}\\0.431~/~\underline{0.366}\end{tabular}
\\
\bottomrule[1.0pt]
\end{tabular}
}
\vspace{-0.4cm}
\end{table}

\vspace{0.5em}
\noindent\textbf{Deraining.} Substantial improvements of \textbf{2.1} dB and \textbf{0.9} dB are evidenced in Tables~\ref{tab-exp-outdoorain} and \ref{tab-exp-raindrop}, respectively. Notably, the proposed method, employing both dual-step and single-step approaches, demonstrates superior performance compared to specialized deraining diffusion models. \revision{This underscores the necessity and effectiveness of our strategies for ODE trajectory augmentation and simplification.}
\revision{Furthermore, Table~\ref{tab-exp-raindrop02} reveals that our proposed method achieves superior perceptual scores, such as MUSIQ and CLIPIQA, while exhibiting favorable computational efficiency.}
Fig.~\ref{fig:deraining} provides visual results of different methods. Specifically, for the task of raindrop removal, we visualized the regions seriously deteriorated by raindrops, e.g., the roof of a car and shelves on the playground, in zoom-in sub-figures, where the input measurement, i.e., Fig.~\ref{fig:deraining}-(\textcolor{red}{a}), indicates the object structure has been greatly damaged. However, even with this kind of degradation, the proposed method can accurately reconstruct the original structure, validating the superiority of the proposed method. The first sample of the Raindrop dataset also validates that the proposed method can correct the color of texture since the other methods show more red components compared with the proposed method. Meanwhile, we also visualize the deraining experimental results of \textit{Outdoor-Rain} dataset in Fig.~\ref{fig:deraining}. Note that our method is not specifically trained to reconstruct words. However, to our surprise, the strong restoration capacity of the proposed method enables accurate reconstruction of words and numbers, e.g., ``2.8" in the first sample and ``U" in the second sample of the Outdoor-Rain dataset.

\revision{Table~\ref{tab-exp-comparison-deraining-01} further presents comprehensive comparisons with state-of-the-art methods on more challenging deraining datasets, including Rain100L/H \cite{yang2017deep}, Rain200L/H \cite{yang2017deep}, DID-Data \cite{zhang2018density}, and DDN-Data \cite{fu2017removing}. Our method exhibits consistent superiority, achieving the highest PSNR and SSIM scores across most datasets. While the performance on Rain200L and Rain200H datasets appears moderate compared to some recent methods, our approach still demonstrates substantial improvements over the baseline diffusion model GSAD, with notable gains of \textbf{+1.06} dB on Rain200L and \textbf{+1.89} dB on Rain200H. \secrevision{Moreover, we want to note that the proposed strategy is orthogonal to the neural network structure, i.e., we can also utilize the transformer-based networks, e.g.,\cite{chen2023single} \cite{gao2024efficient}, to build a diffusion model and apply our algorithm to boost their performance. We refer readers to Sec.~\textcolor{red}{IX} of the \textit{Supplementary Material} for the results.} These improvements highlight the effectiveness of our method in handling complex rainy scenarios, even when faced with more challenging datasets.}

Based on the aforementioned analysis, we conclude that the proposed method can accurately reconstruct both natural texture and cultural markers, and greatly outperform the state-of-the-art method by a large extent, validating the effectiveness of the proposed image restoration diffusion augmentation strategy.

\begin{table}
\begin{threeparttable}[b]
\tiny
\centering
\caption{Results of the ablative study on the effect of the proposed two training augmentation techniques. ``RL" denotes the reinforcement learning-based alignment. ``DISTILL" represents the \revision{single}-step distillation for inference acceleration. ``INTERP" indicates the interpolation of the starting point. ``NGS" represents utilizing the low-quality image as negative guidance.}
\vspace{-0.2cm}
\label{table:Ablative_loss}
\setlength{\tabcolsep}{0.4mm}{
\renewcommand{\arraystretch}{1.0}
\begin{tabular}{ccccc|c|ccc|ccc|ccc}
\toprule[1.0pt]
\multicolumn{1}{c}{\multirow{2}{*}{IDX.}} &\multirow{2}{*}{RL} & \multirow{2}{*}{DISTILL} & \multirow{2}{*}{INTERP} & \multirow{2}{*}{NGS} & \multirow{2}{*}{NFE\tnote{(*)} \quad} & \multicolumn{3}{c|}{UIEBD} & \multicolumn{3}{c|}{Raindrop} & \multicolumn{3}{c}{LOL-v2} \\
\cmidrule(lr){7-9} \cmidrule(lr){10-12} \cmidrule(lr){13-15}
&& & & & & PSNR$\uparrow$ & SSIM$\uparrow$ & LPIPS$\downarrow$ & PSNR$\uparrow$ & SSIM$\uparrow$ & LPIPS$\downarrow$ & PSNR$\uparrow$ & SSIM$\uparrow$ & LPIPS$\downarrow$ \\
\midrule
\multirow{2}{*}{\newtag{\textbf{1}}{V:base}} &\multirow{2}{*}{\XSolidBrush} & \multirow{2}{*}{\XSolidBrush} & \multirow{2}{*}{\XSolidBrush} & \multirow{2}{*}{\XSolidBrush} & 10 & 24.31 & 0.916 & 0.151 & 32.86 & 0.942 & 0.059 & 28.78 & 0.895 &\underline{0.094} \\
&& & & & 1 & 19.99 & 0.802 & 0.284 & 20.92 & 0.357 & 0.674 & 20.35 & 0.717 & 0.255 \\
\midrule
\multirow{2}{*}{\newtag{\textbf{2}}{V:rein}} &\multirow{2}{*}{\Checkmark} & \multirow{2}{*}{\XSolidBrush} & \multirow{2}{*}{\XSolidBrush} & \multirow{2}{*}{\XSolidBrush} & 10 & 25.08 & 0.913 & 0.162 & 33.32 & 0.938 & \textbf{0.044} & 29.54 &\underline{0.898} &\textbf{0.086} \\
&& & & & 1 & 21.06 & 0.715 & 0.457 & 24.54 & 0.807 & 0.266 & 21.35 & 0.690 & 0.263 \\
\midrule
\multirow{2}{*}{\newtag{\textbf{3}}{V:distill}} &\multirow{2}{*}{\Checkmark} & \multirow{2}{*}{\Checkmark} & \multirow{2}{*}{\XSolidBrush} & \multirow{2}{*}{\XSolidBrush} & 10 & 24.04 & 0.856 & 0.225 & 30.26 & 0.920 & 0.057 & 28.19 & 0.863 & 0.137 \\
&& & & & 1 & 25.80 & 0.923 & 0.153 & 33.35 & 0.942 & \underline{0.049} &29.75 &0.904 & 0.096 \\
\midrule
\multirow{2}{*}{\newtag{\textbf{4}}{V:inter}} &\multirow{2}{*}{\Checkmark} & \multirow{2}{*}{\Checkmark} & \multirow{2}{*}{\Checkmark} & \multirow{2}{*}{\XSolidBrush} & 10 & 24.20 & 0.923 & 0.136 & 29.80 & 0.918 & 0.056  &28.30  &0.864  &0.135  \\
&& & & & 1 & \underline{25.94} & \underline{0.937} & \textbf{0.127} & \underline{33.53} & \textbf{0.946} & 0.052 & \underline{29.86}  &\textbf{0.904}  &0.102  \\
\midrule
\multirow{2}{*}{\newtag{\textbf{5}}{V:guidance}} &\multirow{2}{*}{\Checkmark} & \multirow{2}{*}{\Checkmark} & \multirow{2}{*}{\Checkmark} & \multirow{2}{*}{\Checkmark} & 10 & 21.87 & 0.845 & 0.261 & 29.25 & 0.877 & 0.129 &28.33  &0.863  &0.130  \\
&& & & & 1 & \textbf{26.25} & \textbf{0.938} & \underline{0.128} & \textbf{33.63} & \textbf{0.946} & 0.052 &\textbf{29.91}  &\textbf{0.904}  &0.101  \\
\bottomrule[1.0pt]
\end{tabular}}
 \begin{tablenotes}
 \scriptsize
   \item [(*)] \secrevision{This column indicates the NFE values during testing. In IDX. 1, the model was trained continuously. In IDX. 2, the model was trained with NFE equal to 10. Moreover, for IDXs. 3, 4 and 5, the models were trained with NFE equal to 1.}
 \end{tablenotes}
\end{threeparttable}
\vspace{-0.9cm}
\end{table}

\begin{table}[h]
\centering
\tiny
\label{table:Ablative_gamma}
\caption{\revision{Results of the ablation study on parameter $\gamma$ of the proposed Modulated SDE.}}
\vspace{-0.2cm}
\setlength{\tabcolsep}{0.4mm}{
\renewcommand{\arraystretch}{1.0}
\begin{tabular}{c|c|ccc|ccc|ccc}
\toprule[1.0pt]
\multirow{2}{*}{}  
 & \multirow{2}{*}{$\gamma$} & \multicolumn{3}{c|}{UIEBD} & \multicolumn{3}{c|}{Raindrop} & \multicolumn{3}{c}{LOL-v2} \\
\cmidrule(lr){3-5} \cmidrule(lr){6-8} \cmidrule(lr){9-11}
& & PSNR$\uparrow$ & SSIM$\uparrow$ & LPIPS$\downarrow$ & PSNR$\uparrow$ & SSIM$\uparrow$ & LPIPS$\downarrow$ & PSNR$\uparrow$ & SSIM$\uparrow$ & LPIPS$\downarrow$ \\
\midrule
\multirow{7}{*}{M-SDE}
 &Learned $\gamma$ & 25.08 & 0.913 & 0.162 & 33.32 & 0.938 & 0.044 & 29.54 &0.898 &0.086 \\ 
\cmidrule{2-11}
 &0.05  &24.84 &0.911 &0.165   &33.21 &0.944 &0.053   & 29.37 & 0.902 & 0.097 \\
 &0.10  &24.65 &0.915 &0.155   &32.45 &0.929 &0.053   & 29.52 & 0.905 & 0.093 \\
 &0.15  &24.48 &0.917 &0.159   &30.09 &0.831 &0.150   & 29.51 & 0.889 & 0.087 \\
 &0.20  &24.90 &0.932 &0.140   &29.69 &0.747 &0.238   & 27.96 & 0.772 & 0.142 \\
 &0.25  &24.32 &0.922 &0.145   &29.35 &0.744 &0.243   &27.18 &0.708 &0.152   \\
 &0.30  &24.18 &0.929 &0.142   &29.26 &0.742 &0.212   & 27.68 & 0.758 & 0.146 \\
\midrule
 \multicolumn{2}{c|}{SDE ($\gamma=1$)} & 24.52 & 0.928 & 0.147 & 32.18 & 0.939 & 0.065 & 29.23 & 0.895 & 0.097 \\
\bottomrule[1.0pt]
\end{tabular}}
\end{table}

\begin{table}[h]
\centering
\tiny
\caption{\revision{Results of the ablation study on different distillation schemes. The best results are highlighted in \textbf{bold}.}}
\vspace{-0.2cm}
\label{table:Ablative_distill}
\setlength{\tabcolsep}{0.7mm}{
\renewcommand{\arraystretch}{1.0}
\begin{tabular}{l|c|ccc|ccc|ccc}
\toprule[1.0pt]
\multirow{2}{*}{Method} & \multirow{2}{*}{NFE} & \multicolumn{3}{c|}{UIEBD} & \multicolumn{3}{c|}{Raindrop} & \multicolumn{3}{c}{LOL-v2} \\
\cmidrule(lr){3-5} \cmidrule(lr){6-8} \cmidrule(lr){9-11}
& & PSNR$\uparrow$ & SSIM$\uparrow$ & LPIPS$\downarrow$ & PSNR$\uparrow$ & SSIM$\uparrow$ & LPIPS$\downarrow$ & PSNR$\uparrow$ & SSIM$\uparrow$ & LPIPS$\downarrow$ \\
\midrule
Baseline &10 & 25.08 & 0.913 & 0.162 & 33.32 & 0.938 & \textbf{0.044} & 29.54 &{0.898} &\textbf{0.086} \\
CM \cite{song2023consistency} &1 & 25.65  & 0.904  & 0.183  & 33.09 & 0.933 & 0.049 & 29.50 & 0.882 & 0.105 \\
CTM \cite{kim2024consistency} &1 & 25.84  & 0.919  & 0.164  & 33.17 & 0.936 & {0.048} & 29.56 & 0.884 &{0.101} \\
Proposed Method & 1 & \textbf{26.25} & \textbf{0.938} & \textbf{0.128} & \textbf{33.63} & \textbf{0.946} & 0.052 &\textbf{29.91}  &\textbf{0.904}  &{0.101}  \\
\bottomrule[1.0pt]
\end{tabular}}
\vspace{-0.4cm}
\end{table}

\begin{table}[h]
\centering
\tiny
\caption{\revision{Results of the ablation study on text prompts of the unified FLUX-IR framework.}}
\vspace{-0.2cm}
\label{table:Ablative_text_input}
\setlength{\tabcolsep}{0.8mm}{
\renewcommand{\arraystretch}{1.0}
\begin{tabular}{l|cccc|cccc}
\toprule[1.05pt]
 \multirow{2}{*}{Text Prompt} 
 & \multicolumn{4}{c|}{Super-resolution (DIV2K-Val)} 
 & \multicolumn{4}{c}{Low-light Enhancement (LOL)} 
\\
\cmidrule(lr){2-5} 
\cmidrule(lr){6-9} 
&~NIQE$\downarrow$  & MUSIQ$\uparrow$  & CLIPIQA$\uparrow$ & PSNR$\uparrow$~
&~NIQE$\downarrow$  & MUSIQ$\uparrow$  & CLIPIQA$\uparrow$ & PSNR$\uparrow$
\\ \midrule
 No prompt            &4.093 &67.481 &0.602 &21.467   &4.172 &72.939 &0.478 &24.918    \\
 Generic prompt       &3.872 &68.942 &0.640 &21.536   &4.193 &73.605 &0.498 &24.616    \\
 Task-specific prompt &3.491 &73.188 &0.721 &20.248   &4.045 &73.233 &0.484 &25.029    \\
\bottomrule[1.05pt]
\end{tabular}}
\vspace{-0.7cm}
\end{table}

\vspace{-0.2cm}
\subsection{Unified Perceptual Image Restoration with FLUX-IR}
\label{sec:exponetoall}

\noindent\textbf{Implementation Details.} We trained a single diffusion model to handle various types of degradation. Here, we constructed the unified image restoration network based on FLUX-DEV~\cite{Flux}.     Specifically, we first trained a low-quality U-Net encoder to make its feature map consistent with those from high-quality images. Then, we further trained a Control-Net by \textit{XFLUX}~\cite{XFlux}. We began by training a low-quality U-Net encoder to ensure its feature map aligned with those from high-quality images. Subsequently, we trained a control network using \textit{XFLUX}~\cite{XFlux}. During this training, we integrated features from \textit{XFLUX} and the pre-trained encoder into the DiT structure of FLUX to enable rapid adaptation of our FLUX-IR framework. The model was trained on five NVIDIA H800 GPUs for 20,000 iterations, utilizing the Adam optimizer, a learning rate of $5e^{-5}$, a training patch size of $1024^2$, and a batch size of $128$ (with gradient accumulation). After this pre-training phase, we enhanced the FLUX-IR model using our proposed strategy. Given the significant size and training costs associated with FLUX, we combined reinforcement learning and distillation into a single training phase, setting the diffusion timestamp to 9 and interpolating the initial state using the low-quality latent. We then applied reinforcement learning with guidance to further improve performance over a few inference steps. During this phase, the reward was calculated as the average of the following metrics: LPIPS, NIQE, MUSIQ, and CLIPIQA, with normalization factors of $-1$, $-20$, $70$, and $1$, respectively. During training, we sampled the data from the task of super-resolution at a frequency of $5/10$ and others are the same as $5/60$.

\vspace{0.5em}
\noindent\textbf{Experimental Results.} We validated the effectiveness of the proposed trajectory augmentation techniques in the unified image restoration task, which contains \textit{7} distinct image restoration tasks.  Experimental results are shown in Table~\ref{Table:Flux-IR}. Benefiting from strong capacity and our delicately designed learning scheme, FLUX-IR achieves extraordinary performance on the perceptual metrics, which even beats the task-specific methods, e.g., Stable SR~\cite{wang2024exploiting} and PASD~\cite{yang2023pasd} from the task of super-resolution. We want to note that accurately quantifying the perceptual performance is a difficult issue. The outstanding performance of the proposed method, which may generate more reasonable and clear structures even than reference images, may make evaluation more difficult. 

We visually compare various methods in Fig.~\ref{fig:unified} across four tasks: super-resolution, denoising, deblurring, and desnowing, which were not previously illustrated. Corresponding metrics are annotated in the corners of the images. While reference-aware metrics such as PSNR are critical for traditional image restoration tasks, relying solely on these metrics is inadequate for assessing quality in real-world scenarios. For instance, in the super-resolution task, manual interpolation, as depicted in Fig. \ref{fig:unified}\textcolor{red}{-SR-(a)}, achieves superior PSNR scores compared to all other methods, yet its visual quality is inferior. Furthermore, evaluating quality without reference poses significant challenges, and a single perceptual metric may exhibit instability under certain conditions. Therefore, our approach of integrating multiple perceptual metrics for reinforcement learning is both practical and effective.

The following is a detailed analysis of the visual comparison results. Specifically, for the task of \textit{super-resolution}, FLUX-IR significantly outperforms the compared methods, particularly evident in the depiction of building details and the feathers of the parrot in the first and third rows, respectively. These results not only match but sometimes surpass the perceptual quality of the reference images. The strong capabilities of FLUX-IR are especially highlighted in human-related restoration, an area highly sensitive to our perception. As illustrated in the second row, FLUX-IR effectively generates realistic human faces and bodies, while the other methods struggle to produce plausible outcomes. Similar trends are observed in the tasks of denoising and desnowing. Lastly, for deblurring, this task differs from previous challenges involving noise, snowflakes, and downsampling, as the deblurring process merely mixes without compromising the original images. Consequently, all methods produce plausible images, as shown in the last row of Fig.~\ref{fig:unified}.

\revision{
Additionally, we conducted denoising evaluations on BSD500 \cite{arbelaez2010contour} and SIDD \cite{SIDD_2018_CVPR} datasets to demonstrate the generalization ability of our proposed Flux-IR, where we directly tested on these datasets without training on the relevant training sets. As shown in Table \ref{Table:Flux-IR-ga}, our Flux-IR demonstrates superior perceptual quality with best NIQE, MUSIQ, and MANIQA scores, highlighting the strong generalization ability of the proposed Flux-IR. \secrevision{We also validated the image restoration performance of the proposed method on the multiple degradations in Fig.~\textcolor{red}{6} of the \textit{Supplementary Material}.}
}

\vspace{-0.2cm}
\subsection{Ablation Studies}
\noindent \textbf{Reconstruction Performance.} We have applied the detailed ablation studies of each trajectory augmentation technique in Table~\ref{table:Ablative_loss}. Specifically, for reinforcement learning-based ODE alignment (\textit{RL}), it improves both single and multiple-step image restoration. Notably, models utilizing single-step alignment exhibit substantial improvements, such as an increase from $20.92$ dB to $24.54$ dB on the Raindrop dataset. This enhancement is attributed to the significant simplification of the trajectory post-alignment, which reduces integral estimation errors and benefits single-step inference. Conversely, few-step distillation (\textit{DISTILL}) improves single-step performance but diminishes network performance with multiple steps. Furthermore, interpolation (\textit{INTERP}) and guided sampling techniques (\textit{NGS}) generally enhance single-step diffusion models by alleviating the burden on neural networks to directly match Gaussian noise distributions with real-world data. However, for multi-step diffusion models, the use of interpolated noisy latent and negative guidance from low-quality images may present limitations. \secrevision{We refer readers to Sec.~\textcolor{red}{VII} of \textit{Supplementary Material} for further cross-validation of distillation and RL.}

\begin{figure}
    \centering
    \includegraphics[width=0.9\linewidth]{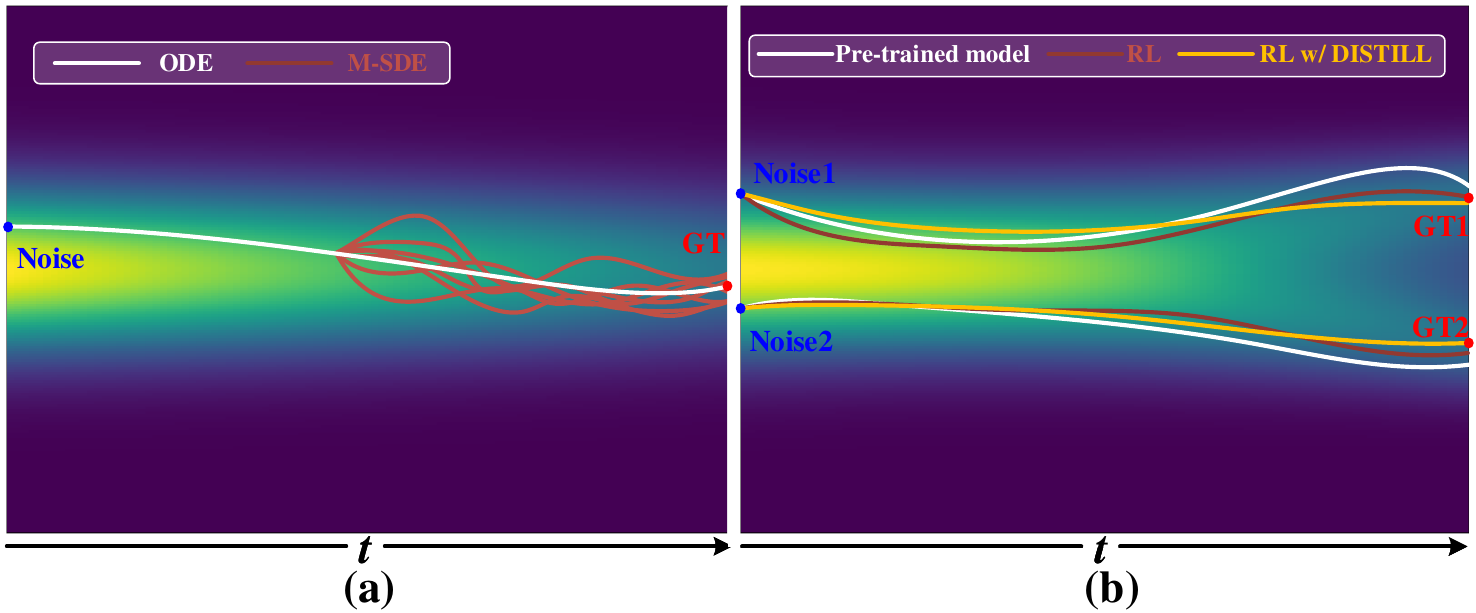}
\vspace{-0.4cm}
    \caption{Trajectories of different models. (a) Visualization of ODE and M-SDE trajectories for the pre-trained model. It can be seen that some M-SDE trajectories are more effective, being closer to the ground truth (GT) than the ODE counterpart. The optimal M-SDE trajectory is then selected as guidance for our reinforcement learning-based alignment process. (b) Visualization of ODE trajectories for the pre-trained model, our reinforced model, and our distilled model. ``$t$" indicates the diffusion timestamp. Both reinforced and distilled models generate more effective ODE trajectories compared to the pre-trained model.}
    \label{fig:trajectory}
\vspace{-0.6cm}
\end{figure}

\noindent \textbf{Visualization of Trajectories.} As our design is primarily centered on modeling trajectories of neural differential equations, we also analyze the variations in these trajectories to gain insights into the effects of our proposed method. As shown in Fig.~\ref{fig:trajectory}-(\textcolor{red}{a}), we illustrate the ODE and M-SDE trajectories of the pre-trained model. The M-SDE trajectories exhibit more variance and potential compared to the ODE counterpart, which enables our reinforced ODE trajectory learning. Fig.~\ref{fig:trajectory}-(\textcolor{red}{b}) depicts the ODE trajectories of the pre-trained, reinforced, and distilled models. Here, the reinforced ODE trajectory demonstrates a more direct path toward the target distribution, indicating that our reinforcement learning approach effectively optimizes the restoration path. Moreover, the distilled ODE trajectory closely approximates the reinforced trajectory, showcasing the efficacy of our trajectory distillation in preserving the optimized path while reducing computational complexity.

\noindent \revision{\textbf{Analysis of ODE Alignment with Modulated SDE.}
We investigated the effectiveness of aligning ODE with a modulated SDE (M-SDE) compared to direct ODE-SDE alignment. As shown in Table \ref{table:Ablative_gamma}, the M-SDE approach consistently performs better than direct SDE alignment across all evaluated datasets. This improvement holds whether using a learned $\gamma$ or an optimally chosen fixed value, validating the effectiveness of our proposed alignment method. Furthermore, to better understand the role of the modulation parameter, we conducted an ablation study with various fixed $\gamma$ values. Our experiments reveal that model performance is notably sensitive to the choice of $\gamma$, thus highlighting a key advantage of our learned $\gamma$ approach, i.e., the model can automatically optimize its behavior for different datasets and degradation types by dynamically adapting the alignment during training, eliminating the need for manual parameter tuning while ensuring robust performance.} \secrevision{We refer readers to the results in Sec.~\textcolor{red}{VII} of the \textit{Supplementary Material}, which demonstrates the advantage of the proposed RL method.}

\noindent \revision{\textbf{Different Distillation Schemes.} We compared our approach against two state-of-the-art diffusion distillation methods: Consistency Model (CM)~\cite{song2023consistency} and Consistency Trajectory Model (CTM)~\cite{kim2024consistency}. From Table~\ref{table:Ablative_distill}, it can be observed that the proposed distillation method yields consistent improvements across multiple datasets, highlighting its effectiveness in achieving efficient inference while ensuring high-quality image restoration.}

\vspace{0.5em}
\noindent \revision{\textbf{Text Conditioning Analysis.} Given FLUX's text-to-image nature, we examined text prompting effects on restoration performance. Table \ref{table:Ablative_text_input} compares three strategies, i.e., no prompt, generic prompt ("a high-quality image" for all tasks), and task-specific prompt (e.g., "high-resolution, ultra-sharp, detailed" for super-resolution). It can be seen that both generic and task-specific prompts demonstrate improvements in perceptual metrics relative to the no-prompt baseline, with task-specific prompts yielding superior performance, indicating that FLUX's text understanding capabilities can be effectively leveraged for image restoration, with appropriate text guidance helping steer the model toward better perceptual quality. Based on these findings, we adopt task-specific prompts in our unified FLUX-IR framework.}

\vspace{-0.2cm}
\section{Conclusion}
\label{Sec:conclusion}
We have presented an efficient yet effective trajectory optimization paradigm for image restoration-based diffusion models. Through reinforcement learning-based trajectory augmentation techniques, we boost the effectiveness of the image restoration diffusion network. By employing different reward functions, we can flexibly guide the learning of the diffusion model toward either more objective or perceptual restoration.  
Moreover, on the basis of distillation cost analysis, we introduced a diffusion acceleration distillation pipeline with several techniques to perverse the original knowledge of diffusion models and achieve single-step distillation. We have carried out extensive experiments on both task-specific image restoration diffusion and unified image restoration diffusion networks over more than 7 different image restoration tasks to validate the effectiveness of the proposed method. Moreover, we have also calibrated a 12B rectified flow-based model for the image restoration task. The experimental results demonstrate the effectiveness of the proposed method, which generates clear and meaningful results compared to the state-of-the-art methods. We believe that our insights and findings would push the frontier of image restoration.

\revision{Building on the proposed method, promising future work includes developing more advanced reward metrics and integration mechanisms, as well as exploring alternative approaches, such as bridge diffusion, to enhance the model's controllability and reconstruction fidelity.}
\bibliographystyle{IEEEtran}
\bibliography{reference}
\vspace{-0.2cm}
\begin{IEEEbiography}[{\includegraphics[width=1in,height=1.25in,clip,keepaspectratio]{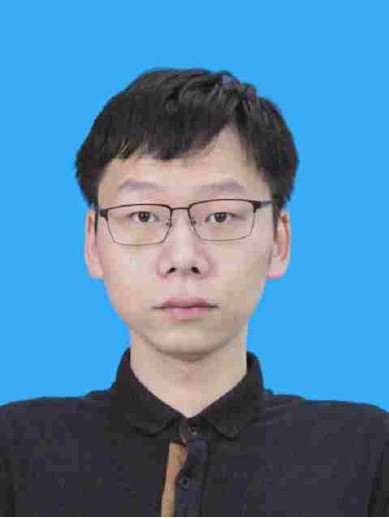}}]{Zhiyu Zhu}  received the B.E. and M.E. degrees in Mechatronic Engineering, both from Harbin Institute of Technology, in 2017 and 2019, respectively. He has also received a Ph.D. degree in Computer Science from the City University of Hong Kong in 2023, where he currently holds a postdoctoral position. His research interests include generative models and computer vision.
\end{IEEEbiography}
\vspace{-1.6cm}
\begin{IEEEbiography}[{\includegraphics[width=1in,height=1.25in,clip,keepaspectratio]{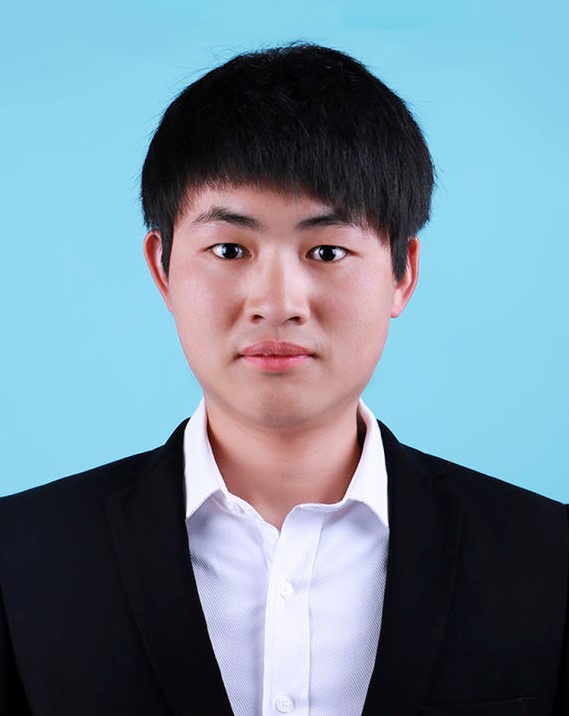}}]{Jinhui Hou} received the B.E. and M.E. degrees in Communication Engineering from Huaqiao University, Xiamen, China, in 2017 and 2020, respectively. He is pursuing a Ph.D. in Computer Science at the City University of Hong Kong. His research interests include image processing and computer vision.
\end{IEEEbiography}
\vspace{-1.6cm}
\begin{IEEEbiography}[{\includegraphics[width=1in,height=1.25in,clip,keepaspectratio]{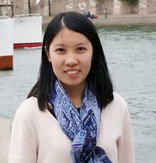}}]{Hui Liu} is currently an Assistant Professor in the School of Computing and Information Sciences at Saint Francis University, Hong Kong. She received the B.Sc. degree in Communication Engineering from Central South University, Changsha, China, the M.Eng. degree in Computer Science from Nanyang Technological University, Singapore, and the Ph.D. degree from the Department of Computer Science, City University of Hong Kong, Hong Kong. Her research interests include image processing and machine learning.
\end{IEEEbiography}
\vspace{-1.2cm}
\begin{IEEEbiography}[{\includegraphics[width=1in,height=1.25in,clip,keepaspectratio]{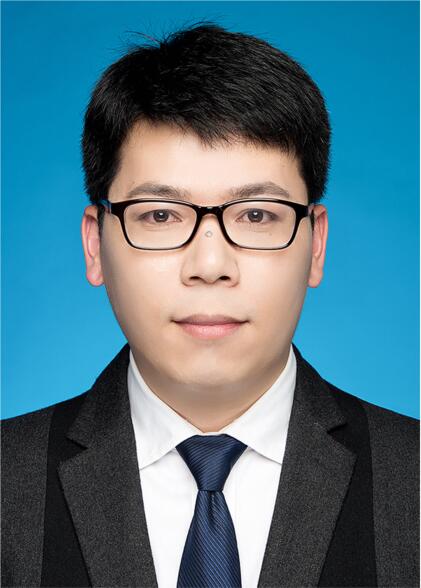}}]{Huanqiang Zeng} 
is currently a Full Professor at the School of Engineering and the School of Information Science and Engineering, Huaqiao University. 
He has been actively serving as an Associate Editor for IEEE Transactions on Image Processing, IEEE Transactions on Circuits and Systems for Video Technology, and Electronics Letters (IET). He has been actively serving as a Guest Editor for Journal of Visual Communication and Image Representation, Multimedia Tools and Applications, and Journal of Ambient Intelligence and Humanized Computing.
\end{IEEEbiography}
\vspace{-1.2cm}
\begin{IEEEbiography}[{\includegraphics[width=1in,height=1.25in,clip,keepaspectratio]{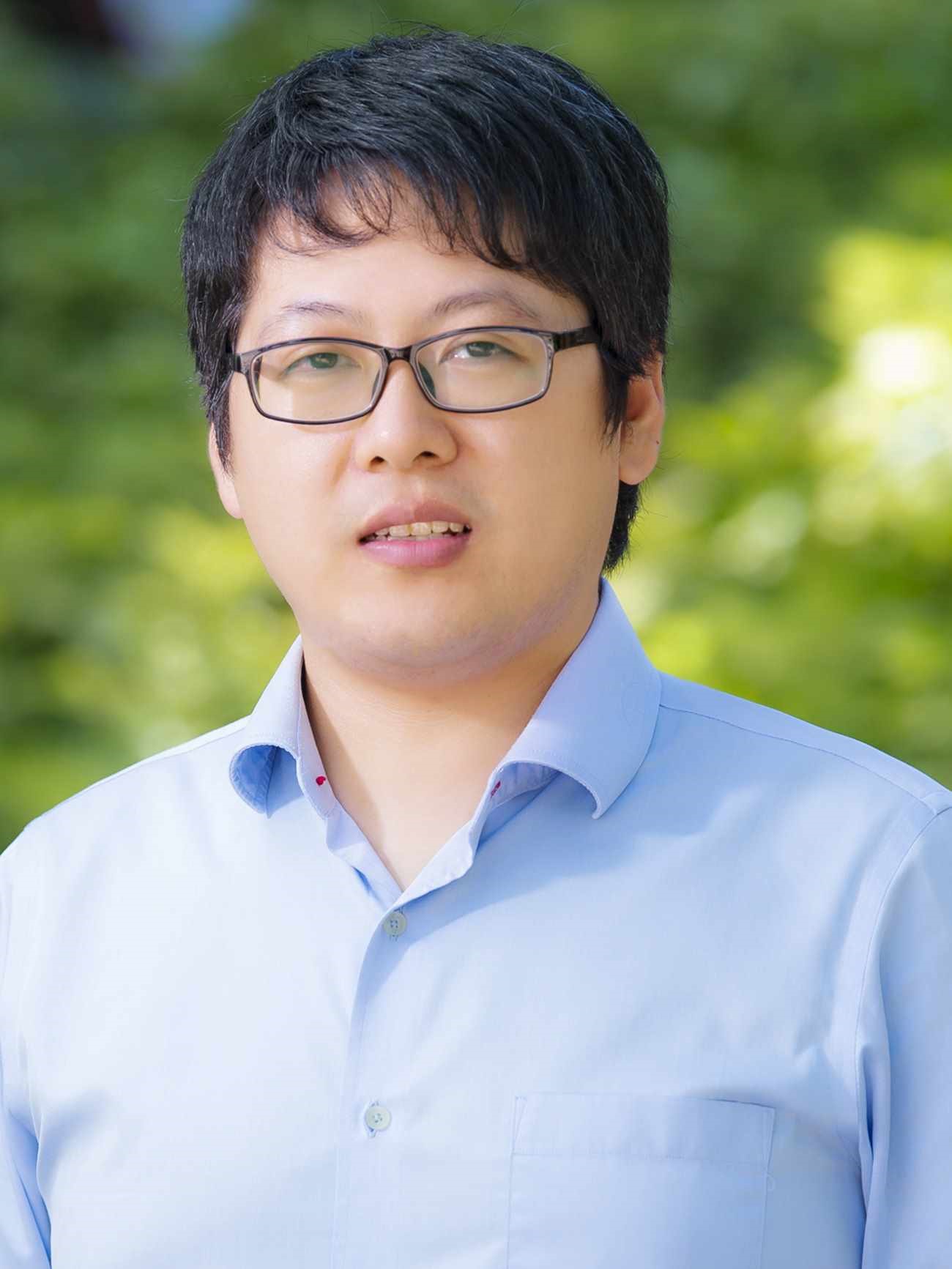}}]{Junhui Hou}
(Senior Member, IEEE) is an Associate Professor with the Department of Computer Science, City University of Hong Kong. 
His research interests are multi-dimensional visual computing.

Dr. Hou received the Early Career Award (3/381) from the Hong Kong Research Grants Council in 2018 and the NSFC Excellent Young Scientists Fund in 2024. He has served or is serving as an Associate Editor for \textit{IEEE Transactions on Visualization and Computer Graphics}, \textit{IEEE Transactions on Image Processing}, \textit{IEEE Transactions on Multimedia}, and \textit{IEEE Transactions on Circuits and Systems for Video Technology}.  
\end{IEEEbiography}

\clearpage

{\appendices
\section{Adaptively Adjusting Noise Intensity is Necessary for Alignment of Diffusion Trajectories}
\label{Appendex:T0}
In this section, we analyze that, for a reinforcement alignment process in diffusion models, it is necessary to adaptively adjust the intensity to compensate for the score estimation error.
Considering the time-stamp of the to-be-aligned feature map as $t$, according to the DPM-Solver, we can calculate $\mathbf{X}_0$ by $\mathbf{X}_t$ as
\begin{align}
    \mathbf{X}_0 = \frac{\alpha_0}{\alpha_t}\mathbf{X}_t + \left(\frac{\sigma_\tau}{\alpha_\tau} \Big|_{\tau=t}^{\tau=0}\right) \alpha_0 \bm{\epsilon}_{\bm{\theta}}\left(\mathbf{X}_t, t\right) .\nonumber
\end{align}
\noindent Then,  for the ground-truth $\mathbf{X}_{0}^{*}$, we have 
\begin{align}
\mathbf{X}_0^* = \frac{\alpha_0}{\alpha_t}\mathbf{X}_t^* + \left(\frac{\sigma_\tau}{\alpha_\tau} \Big|_{\tau=t}^{\tau=0}\right) \alpha_0 \bm{\epsilon}_{\bm{\theta}}\left(\mathbf{X}_t^*, t\right),\nonumber
\end{align}
where $\mathbf{X}_t^*$ is obtained via adding random perturbation to $\mathbf{X}_t$, i.e., $\mathbf{X}_t^* = \mathbf{X}_t + \gamma \bm{\epsilon}$, ($\bm{\epsilon} \sim \mathcal{N}(\mathbf{0},\mathbf{I})$). Thus, we have 
\begin{align}
    &\left\|\mathbf{X}_0 - \mathbf{X}_0^* \right\|_2 \nonumber \\
    &= \left\| \frac{\alpha_0}{\alpha_t} (\mathbf{X}_t - \mathbf{X}_t^*) + \left(\frac{\sigma_\tau}{\alpha_\tau}\Big|_{\tau = t}^{\tau = 0} \right) \alpha_0 \left( \bm{\epsilon}_{\bm{\theta}}(\mathbf{X}_t, t) - \bm{\epsilon}_{\bm{\theta}}(\mathbf{X}_t^*, t) \right) \right\|_2 \nonumber\\
    &\overset{\textcircled{1}}{\approx} \left\| \frac{\alpha_0}{\alpha_t} (\mathbf{X}_t - \mathbf{X}_t^*) + \left(\frac{\sigma_\tau}{\alpha_\tau}\Big|_{\tau = t}^{\tau = 0} \right) \alpha_0 k \left( \mathbf{X}_t - \mathbf{X}_t^* \right) \right\|_2 \nonumber\\
    &\overset{}{=} \left|\frac{\alpha_0}{\alpha_t} + \alpha_0 k\left(\frac{\sigma_\tau}{\alpha_\tau}\Big|_{\tau = t}^{\tau = 0} \right)\right| \left\| \mathbf{X}_t - \mathbf{X}_t^* \right\|_2  \nonumber,
\end{align}
where $\textcircled{1}$ holds under the assumption that the noise prediction network $\bm{\epsilon}_{\bm{\theta}}(\cdot)$ can accurately predict all the noise. Thus, we have $\bm{\epsilon}_{\bm{\theta}}(\mathbf{X}_t, t) - \bm{\epsilon}_{\bm{\theta}}(\mathbf{X}_t^*, t) \approx k(\mathbf{X}_t -\mathbf{X}_t^*)$ and 
\begin{align}
    \| \mathbf{X}_t - \mathbf{X}_t^* \|_2 \approx \frac{\|\mathbf{X}_0 - \mathbf{X}_0^* \|_2}{\left|\frac{\alpha_0}{\alpha_t} - k\alpha_0\left( \frac{\sigma_\tau}{\alpha_\tau}\Big|_{\tau = 0}^{\tau=t} \right)\right|}. \nonumber
\end{align}
Considering the parameterization $\mathbf{X}_t^* = \mathbf{X}_t + \gamma \bm{\epsilon}$, ($\bm{\epsilon} \sim \mathcal{N}(\mathbf{0},\mathbf{I})$), we have $\| \mathbf{X}_t - \mathbf{X}_t^* \|_2 = \gamma$.

Based on the above analyses, it can be concluded that $\gamma$ is correlated with the reconstruction error $\|\mathbf{X}_0 -\mathbf{X}_0^* \|_2$ and the timestamp $t$ to be adjusted. Thus, we parameterize a learnable function $\gamma_\phi(\|\mathbf{X}_0 - \mathbf{X}_0^* \|_2, t)$.

\section{Modulated-SDE and its Integral Solver}
\label{Appendex:T1}
\subsection{Modulated-SDE as a General Expression of Reverse Diffusion Derivative Equation}
The proof generally follows the diffusion ODE from ~\cite{song2020score}. Considering a general formulation of forward diffusion SDE as 
\begin{equation}
    d\mathbf{X} = f(\mathbf{X},t) dt + g(t)d \bm{\omega}, \nonumber
\end{equation}
the marginal probability $\mathbf{P}(\mathbf{X}_t)$ evolves with the following Kolmogorov’s forward equation~\cite{oksendal2003stochastic}:
\begin{align}
\begin{split}
    &\frac{d p(x)}{dt} =  -\frac{ d (f(x,t) p(x))}{ dx} + \frac{1}{2}\frac{d^2(g^2(x,t)p(x))}{dx^2}, \nonumber\\
    &=  -\frac{d (f(x,t) p(x))}{ dx} + \left[ \frac{1 + \gamma^2(t) }{2} - \frac{\gamma(t)^2}{2} \right] \frac{d^2(g^2(x,t)p(x))}{dx^2} \nonumber \\
    &=  -\frac{d}{dx}\left[f(x,t) p(x) - \frac{1 + \gamma^2(t)}{2} (\frac{p(x) d g^2(x,t)}{dx} \right.\\
   &\quad \quad \quad \quad\left. + \frac{g^2(x,t) dp(x)}{dx} ) \right]  - \frac{1}{2} \frac{ d^2\left[ \left[ \gamma(t) g(x,t) \right]^2 p(x)\right]}{dx^2}.
\end{split}
\end{align}
For the diffusion process, since $g(t)$ is independent of $\mathbf{X}$, we have
\begin{align}
\begin{split}
    \frac{d p(x)}{dt} & =  -\frac{d}{dx}\left[( f(x,t)  - \frac{1 + \gamma^2(t)}{2}  \frac{g^2(x,t) d\log p(x)}{dx} )p(x) \right] \\
    & - \frac{1}{2}\frac{d^2\left[ \left[ \gamma(t) g(x,t) \right]^2 p(x)\right]}{dx^2}. \nonumber
\end{split}
\end{align}
Considering for the reverse-time SDE with timestamp of $\hat{t}$, $d \hat{t} = - d t$, we then have
\begin{align}
\begin{split}
    \frac{d p(x)}{d\hat{t}} & =  - \frac{d}{dx} - \left[( f(x,t)  - \frac{1 + \gamma^2(t)}{2}  \frac{g^2(x,t) d\log p(x)}{dx} )p(x) \right] \\
    & + \frac{1}{2} \frac{d^2\left[ \left[ \gamma(t) g(x,t) \right]^2 p(x)\right]}{dx^2}. \nonumber
\end{split}
\end{align}

\noindent It actually corresponds to Kolmogorov’s forward equation with the following differential equations:
\begin{align}
\begin{split}
   d \mathbf{X} & = \hat{f}(\mathbf{X},t) d\hat{t} + \hat{g(t)} d\hat{\omega},  \\
    \hat{f}(\mathbf{X},t) & = - \left[ f(x,t)  - \frac{1 + \gamma^2(t)}{2}  \frac{g^2(x,t) d\log p(x)}{dx} \right],
    \\
    \hat{g(t)} & = \gamma(t) g(x,t).
    \nonumber
\end{split}
\end{align}

\noindent By substituting $d \hat{t} = - dt$ into above equation, we have 
\begin{align}
\begin{split}
   d \mathbf{X} & = \left[ f(x,t)  - \frac{1 +\gamma^2(t)}{2}  g^2(x,t) \nabla_x \log p(x) \right] d t \nonumber \\
   & \quad + \gamma(t) g(x,t) d\hat{\omega},
    \nonumber
\end{split}
\end{align}
which is exactly the Modulated-SDE as we mentioned. Note that the same SDE formulation has been introduced in~\cite{zhang2021diffusion,zhang2022fast}. However, our proof follows~\cite{song2020score} that is more straightforward and complete.

\subsection{Integral Solver of Modulated-SDE}
\label{Appendix:Solver}
Here we give the calculation of the integral solver for Modulated-SDE. To make such a semi-linear property as ~\cite{lu2022dpm}, we introduce the surrogate function $\mathcal{F}(\mathbf{X}_t, \alpha_t) = \frac{\mathbf{X}_t}{ \alpha_t}$. Furthermore, by substituting $f(t) = \frac{d \log \alpha_t}{dt}$ and $ g^2(t) = \frac{d \sigma_t^2}{dt} - 2 \frac{d \log \alpha_t}{d t} \sigma_t^2$ from~\cite{kingma2021variational}, we have
\begin{align}
    d\mathcal{F} & = \frac{d^{(\gamma)} \mathbf{X}_t}{\alpha_t} - \frac{\mathbf{X}_t d \alpha_t}{\alpha_t^2} \nonumber \\
    & =  \frac{1 + \gamma^2(t)}{2 \alpha_t \sigma_t} g^2(t) \bm{\epsilon}_{\bm{\theta}} dt + \gamma(t)g(t)d \hat{\omega} \nonumber
\end{align}
We take the first-order integral solver as an example. By making integral from both sides, we have
\begin{align}
\begin{split}
    \mathbf{X}_{t-\Delta t} =  \frac{\alpha_{t-\Delta t}}{\alpha_t} \mathbf{X}_{t} &- [1+\gamma^2(t)]\bm{\epsilon}_{\bm{\theta}} (\frac{\alpha_{t-\Delta t}}{\alpha_t} \sigma_t - \sigma_{t-\Delta t})
    \\
    &- \gamma(t)\bm{\epsilon} \sqrt{\int_{t-\Delta t}^t (\sigma_t d \sigma_t - \frac{\sigma_t^2}{\alpha_t} d \alpha_t)}. \nonumber
\end{split}
\end{align}
\noindent To derive the closed-form solution, we consider the specific VP and VE diffusion models. For the VP diffusion model ($\alpha_t^2 + \sigma_t^2 = 1$), we have
\begin{align}
    \mathbf{X}_{t-\Delta t} =  \frac{\alpha_{t-\Delta t}}{\alpha_t} \mathbf{X}_{t} &- [1+\gamma^2(t)]\bm{\epsilon}_{\bm{\theta}} (\frac{\alpha_{t-\Delta t}}{\alpha_t} \sigma_t - \sigma_{t-\Delta t})
    \nonumber\\
    &- \gamma(t)\bm{\epsilon} \alpha_{t-\Delta t}\sqrt{\log \frac{\alpha_{t-\Delta t}}{\alpha_t}}. \nonumber
\end{align}
While, for the VE diffusion model, we have
\begin{align}
    \mathbf{X}_{t-\Delta t} = & \frac{\alpha_{t-\Delta t}}{\alpha_t} \mathbf{X}_{t} - [1+\gamma^2(t)]\bm{\epsilon}_{\bm{\theta}} (\frac{\alpha_{t-\Delta t}}{\alpha_t} \sigma_t - \sigma_{t-\Delta t})
    \nonumber\\
    &- \gamma(t)\bm{\epsilon} \alpha_{t-\Delta t}\sqrt{\alpha_{t-\Delta t}^2 - \alpha_{t}^2}. \nonumber
\end{align}

\section{Probabilistic Flow of Rectified Flow}
\label{Appendix:rectified _Flow}
In this section, we analyze the probabilistic flow of FLUX~\cite{Flux} to prove that $\textcircled{1}$ the integral formulation of Eq.~(\textcolor{red}{13}) corresponds to the stochastic formulation of Eq.~(\textcolor{red}{12}); and $\textcircled{2}$ the ODE in Eq.~(\textcolor{red}{6}) and the SDE formulation in Eq.~(\textcolor{red}{12}) in our reinforcement alignment evolve on the same probabilistic flow. Specifically, rectified flow is designed to directly learn the following velocity as
\begin{align}
    \label{Eq:ODE}
     d x_t = \left(x_0 -x_T \right) dt.
\end{align}

\noindent The deterministic formulation of rectified flow is written as 
\begin{align}
    \label{Eq:Deter}
    x_{t-\Delta_t} = x_t - \Delta_t \frac{d x_t}{d t}.
\end{align}

\noindent The stochastic formulation of rectified flow is expressed as 
\begin{align}
    \label{Eq:Stochastic}
    x_{t-\Delta_t} =  \frac{\left[x_{t} - \alpha_t \Delta_t \frac{d x_t}{dt} - \beta_k\bm{\epsilon}\right]}{(1+\alpha_t\Delta_t-t)+\sqrt{(t-\alpha_t\Delta_t)^2+\beta_k^2}}, 
\end{align}
where $\alpha_t$ is a scalar ($\alpha_t > 1$), and $\beta_k$ is formulated as
\begin{equation}
\label{eq:beta}
    \beta_k = \sqrt{\frac{(t-\Delta_t)^2\left[1 - (t -\alpha\Delta_t)\right]^2}{[1-(t-\Delta_t)]^2} - \left(t - \alpha \Delta_t\right)^2}. 
\end{equation}

\noindent Then, we will illustrate that the aforementioned deterministic and stochastic formulations, i.e., Eq.~\eqref{Eq:Deter} and Eq.~\eqref{Eq:Stochastic}, correspond to the same probabilistic flow. For the probabilistic flow of Eq.~\eqref{Eq:Deter}, we can derive its formulation by substituting Eq.~\eqref{Eq:Stochastic} into Kolmogorov's forward equation as
\begin{align}
    &\frac{d p(x)}{dt} =  -\frac{ d (f(x,t) p(x))}{ dx} + \frac{1}{2}\frac{d^2(g^2(x,t)p(x))}{dx^2}, \nonumber\\
    &=-\frac{ d ((x_0 - x_T) p(x))}{ dx}.
\end{align}

\noindent To derive the probabilistic flow of the stochastic equation, we first substitute Eq.~\eqref{eq:beta} into the denominator expression of Eq.~\eqref{Eq:Stochastic}, obtaining 
\begin{align}
    &(1+\alpha_t\Delta_t-t)+\sqrt{(t-\alpha_t\Delta_t)^2+\beta_k^2} \nonumber \\
    &= (1-(t-\alpha_t\Delta_t)) +     \frac{(t-\Delta_t)\left[1 - (t -\alpha\Delta_t)\right]}{[1-(t-\Delta_t)]} \nonumber \\ 
    &= \frac{1 - (t -\alpha\Delta_t)}{1-(t-\Delta_t)}.\nonumber
\end{align}

\noindent Moreover, we also have 
\begin{align}
    \beta_k \overset{\Delta_t \rightarrow dt}{\approx} \sqrt{2(\alpha_t - 1)}\sqrt{dt} .\nonumber
\end{align}
Then, we can substitute the aforementioned two results accompanied with $x_t = (1-t) x_0 + t x_T$ together into Eq.~\eqref{Eq:Stochastic}, producing 
\begin{align}
    &x_{t-\Delta_t} = \frac{[1-(t-\Delta_t)]}{[1-(t-\alpha \Delta_t)]}[x_t - \alpha_t \Delta_t\frac{d x_t}{dt}-\sqrt{2t\alpha dt}\bm{\epsilon}], \nonumber \\
    &\frac{[1-(t-\alpha dt)]}{[1-(t-dt)]}(x_{t-\Delta_t} - x_t) \nonumber\\&\qquad\qquad= \frac{(1-\alpha_t)dt}{1-(t-dt)} x_t + \alpha_t(x_0 - x_T)dt+\sqrt{2t\alpha_t} d \bm{\omega}, \nonumber \\ 
    \nonumber
    & - d x = \frac{(1-\alpha_t)dt}{1-t}\left((1-t) x_0 + t x_T\right)  \nonumber \\&\qquad\qquad\qquad\qquad\qquad\qquad + \alpha_t(x_0 - x_T)dt +\sqrt{2t\alpha_t} d \bm{\omega}, \nonumber
\end{align}
\begin{align}
\label{Eq:SDE-RFlow}
    & \textcircled{1} \quad d x = x_0 dt + \frac{t-\alpha_t}{1-t} x_Tdt +\sqrt{2(\alpha_t-1)} d \bm{\omega}.
\end{align}
The corresponding Kolmogorov’s forward equation can be written as 
\begin{align}
    &\frac{d p(x)}{dt} =  -\frac{ d (f(x,t) p(x))}{ dx} + \frac{1}{2}\frac{d^2(g^2(x,t)p(x))}{dx^2}, \nonumber\\
    &=-\frac{ d \left(\left(x_0 dt + \frac{t-\alpha_t}{1-t} x_Tdt\right) p(x)\right)}{ dx} + \frac{(\alpha_t-1)\frac{d log p(x)}{dx}p(x)}{dx}.\nonumber
\end{align}
Since $\frac{d log p(x)}{dx} = \frac{x_T}{1-t}$, we can conclude the probabilistic flow of Eq.~\eqref{Eq:Stochastic} is 
\begin{align}
    &\textcircled{2} \quad \frac{d p(x)}{dt} =-\frac{ d \left(\left(x_0 - x_T \right) p(x)\right)}{ dx}, \nonumber
\end{align}
which is the same as Eq.~\eqref{Eq:Deter}. This indicates that the stochastic process described by Eq.~\eqref{Eq:SDE-RFlow} and the deterministic process in Eq.~\eqref{Eq:ODE} represent the same probabilistic flow. Consequently, the corresponding integral formulations in Eqs.~\eqref{Eq:Deter} and \eqref{Eq:Stochastic} also describe an identical probabilistic transition process.

\section{Trajectory Distillation Cost}
\label{Sec:TrajectoryDistill}

\subsection{Calculation of Distillation Cost}
In this section, we illustrate the detailed steps for the calculation of the distillation cost. We define the distillation cost as 
\begin{equation}
\label{Eq:Distillation}
\mathcal{C} = \sum_{i=k}^0 \left\| \check{\bm{\bm{\epsilon}}}\left(\frac{\mathbf{X}\big|^{i-1}_{i}}{t \big|^{i-1}_{i}} \Bigg| \frac{d \mathbf{X}_{\bm{\epsilon}}}{d t}\right) -  \epsilon_{\bm{\theta}}( \mathbf{X}_{t_i} ,t_i) \right\|_2.
\end{equation}

\noindent We then start by calculating $\frac{d \mathbf{X}}{d \alpha_t}$ via the DPM-Solver as,
\begin{equation}
    \frac{d \mathbf{X}}{d \alpha_t}  = \frac{\mathbf{X}_t}{\alpha_t} - \frac{(\sigma_{t-\Delta t}- \frac{\alpha_{t-\Delta t}}{\alpha_t}\sigma_t)\epsilon_{\bm{\theta}}}{\alpha_t -\alpha_{t-\Delta t}},\nonumber
\end{equation}
Subsequently, we can obtain
\begin{equation}
    \check{\epsilon} = \left[\frac{\mathbf{X}_t}{\alpha_t} - \frac{\mathbf{X}\big|^{i-1}_{i}}{\alpha \big|^{i-1}_{i}}\right]\frac{\alpha_t - \alpha_{t-\Delta t}}{\sigma_{t-\Delta t} - \frac{\alpha_{t-\Delta t}}{\alpha_t}\sigma_t}.\nonumber
\end{equation}
For the VE diffusion model, the gradient is given by 
\begin{equation}
    \frac{d \mathbf{X}}{d \sigma_t} = -\bm{\epsilon}_{\bm{\theta}}.\nonumber
\end{equation}
This leads to the inverted noise
\begin{equation}
    \check{\bm{\epsilon}} = - \frac{\mathbf{X}\big|^{i-1}_{i}}{\alpha \big|^{i-1}_{i}}.\nonumber
\end{equation}
Finally, the $\mathcal{C}$ can be derived by calculating the L2 Norm of $\check{\bm{\epsilon}}-\bm{\epsilon_\theta}$.

\subsection{Proof of Existence}
"In this section, we first demonstrate the existence of a low-cost distillation strategy. For an arbitrary continuous trajectory and $k\geq 2$, there always exists a distillation strategy with a lower cost than straight flow-based distillation, such as rectified flow or consistency models. However, for $k=1$, the cost remains the same as that of straight flow-based distillation.
Denote by $\mathbf{X}_{T}$ and $\mathbf{X}_{0}$  the clean image and initialized noise, where $T$ and $0$ are the corresponding timestamps, and $T_1 \in \left(0, T\right)$ a mid timestamp. Then, the distillation cost of the straight flow can be formulated as 
\begin{align}
    &\left\| \frac{d\mathbf{X}}{dt_0} - \frac{\mathbf{X}_{T} - \mathbf{X}_{0}}{T} \right\|_2^2  = \left\|  \frac{d\mathbf{X}}{dt_0} - 
 \frac{\mathbf{X}_T - \mathbf{X}_0 }{T} \right\|_2^2 \nonumber \\
    & = \left\| \mathbf{A} + \mathbf{B} \right\|_2^2,\nonumber\\
    &\mathbf{A} = \frac{T_1 \frac{d \mathbf{X}}{dt_0}- \int_0^{T_1}\frac{d\mathbf{X}}{dt} dt}{T} + \frac{(T -T_1)\frac{d\mathbf{X}}{d t_1} -\int_{T_1}^{T} \frac{d \mathbf{X}}{dt} dt}{T}, \nonumber\\
    & \mathbf{B} = \frac{(T-T_1)(\frac{d \mathbf{X}}{d t_0} -\frac{d \mathbf{X}}{d t_1})}{T}. \nonumber
\end{align}
Here, we utilize $T_1$ to divide $\left[0,T\right]$ into 2 segments. 
Considering a single variable $x$, we have
\begin{align}
    \mathbf{a} &= \frac{T_1 \frac{d \mathbf{x}}{dt_0}- \int_0^{T_1}\frac{d\mathbf{x}}{dt} dt}{T} + \frac{(T -T_1)\frac{d\mathbf{x}}{d t_1} -\int_{T_1}^{T} \frac{d \mathbf{x}}{dt} dt}{T}, \nonumber\\
    & \overset{\textcircled{1}}{=}   \frac{T_1}{T} \frac{dx}{dt_0} + \frac{T-T_1}{T}\frac{dx}{dt_1} - \frac{dx}{dt_i} \nonumber\\
     \mathbf{b} &= \frac{(T-T_1)(\frac{d \mathbf{x}}{d t_0} -\frac{d \mathbf{x}}{d t_1})}{T}, \nonumber
\end{align}
where $\textcircled{1}$ is based on the mean value theorem for integral. If $t_1 = t_i$, we then have 
\begin{align}
    \mathbf{a} =   \frac{T_1}{T} \left( \frac{dx}{dt_0} - \frac{dx}{dt_i}\right). \nonumber
\end{align}
Thus,  we then have $\left|\mathbf{a}+\mathbf{b}\right| = \left|\frac{dx}{dt_0} - \frac{dx}{dt_i}\right| \geq\frac{T_i}{T} \left|\frac{dx}{dt_0} - \frac{dx}{dt_i}\right|=\left|\mathbf{a}\right|$. There will always be a low-cost distillation point for $k=2$ than $k=1$ (rectified flow/ consistency model). We can then easily derive similar results for higher $k$ with iterative application of the aforementioned process in sub-intervals.

\section{Synthesizing Noise Latent Helps IR Diffusion}
\label{Sec:Synthesizing}

Assume the noise-prediction neural $\bm{\epsilon}_{\theta}(\cdot)$ has Lipschitz continuity, i.e., $\left\| \bm{\epsilon}_{\theta}(\mathbf{X}_0) - \bm{\epsilon}_{\theta}(\mathbf{X}_1)\right\|_2 \leq k\left\| \mathbf{X}_0 - \mathbf{X}_1 \right\|_2$ . Then, we take a one-step distillation model for example  
\begin{align}
    \mathbf{X}_{T \rightarrow 0} = &\frac{\alpha_{0}}{\alpha_{T}}\mathbf{X}_{T} + \alpha_{0} e^{-\lambda}\Big|_{\lambda_T}^{\lambda_{0}}  \hat{\bm{\epsilon}}_{\theta}(\hat{\mathbf{X}}_{T},T) ,\nonumber\\
    \mathbf{X}_{T-\delta \rightarrow 0} = &\frac{\alpha_{0}}{\alpha_{T-\delta}}\mathbf{X}_{T-\delta} + \alpha_{0} e^{-\lambda}\Big|_{\lambda_{T-\delta}}^{\lambda_{0}}  \hat{\bm{\epsilon}}_{\theta}(\hat{\mathbf{X}}_{T-\delta},T-\delta), \nonumber
\end{align}
We can than get the optimal noise estimation via making $\mathbf{X}_{T \rightarrow 0}$ $\mathbf{X}_{T-\delta \rightarrow 0}$ to be the ground-truth value $\mathbf{X}_0$. Then the optimal noise can be formulated as 
\begin{equation}
    \bm{\epsilon}_{T-\delta \rightarrow 0} = \frac{\frac{\mathbf{X}}{\sigma_0}- SNR_0 \mathbf{Y} - \frac{SNR_0}{SNR_{T-\delta}} \bm{\epsilon}}{1 - \frac{SNR_0}{SNR_{T-\delta}}},\nonumber
\end{equation}
\begin{equation}
    \bm{\epsilon}_{T \rightarrow 0} = \frac{\frac{\mathbf{X}}{\sigma_0}-  \frac{SNR_0}{SNR_{T}} \bm{\epsilon}}{1 - \frac{SNR_0}{SNR_{T}}},\nonumber
\end{equation}
where $SNR$ indicates the signal to noise ratio, i.e., $SNR_0 = \frac{\alpha_0}{\sigma_0}$. We assume that the potential error of a neural network $\mathcal{E}$ is positively correlated with the shift between the target and input, i.e., $\mathcal{E} = k \| \mathbf{X}_{in} - \mathbf{X}_{out}\|_2$, where $k>0$. Thus, we can easily measure the magnitude of error by calculating the following ratio:
\begin{align}
  \frac{\mathcal{E}_{T-\delta}}{\mathcal{E}_T} &= \mathbb{E}_{\bm{\epsilon} \sim \mathcal{N}(\mathbf{0}, \mathbf{I})} \frac{\|\bm{\epsilon}_{T-\delta \rightarrow 0} -\bm{\epsilon}\|_2}{\|\epsilon_{T\rightarrow 0} - \bm{\epsilon}\|_2} \nonumber\\
  &= \mathbb{E}_{\epsilon \sim \mathcal{N}(\mathbf{0}, \mathbf{I})} \frac{\left\|\frac{\frac{\mathbf{X}}{\sigma_0}-SNR_0\mathbf{Y}-\epsilon}{1 - \frac{SNR_0}{SNR_{T-\delta}}}\right\|_2}{\left\|\frac{\frac{\mathbf{X}}{\sigma_0}-\epsilon}{1 - \frac{SNR_0}{SNR_{T}}}\right\|_2} \nonumber\\
  &\overset{\textcircled{1}}{\approx}  \frac{SNR_{T-\delta}}{SNR_T}\frac{ \mathbb{E}_{\epsilon \sim \mathcal{N}(\mathbf{0}, \mathbf{I})} \left\| \frac{\mathbf{X}}{\sigma_0} - SNR_0\mathbf{Y}-\epsilon\right\|_2}{\mathbb{E}_{\epsilon \sim \mathcal{N}(\mathbf{0}, \mathbf{I})} \| \frac{\mathbf{X}}{\sigma_0} -\epsilon \|_2} \nonumber\\
  &=\frac{SNR_{T-\delta}}{SNR_T} \frac{\sqrt{\left\|\frac{\mathbf{X}}{\sigma_0} -SNR_0 Y \right\|_2^2 + dim(\epsilon)}}{\sqrt{\left\|\frac{\mathbf{X}}{\sigma_0} \right\|_2^2 + dim(\epsilon)}} \nonumber,\\
  &\overset{\textcircled{2}}{\approx}\frac{SNR_{T-\delta}}{SNR_T}\frac{\left\|\frac{\mathbf{X}}{\sigma_0} -SNR_0\mathbf{Y} \right\|_2}{\left\| \frac{\mathbf{X}}{\sigma_0} \right\|_2} \overset{\textcircled{3}}{\approx} \frac{SNR_{T-\delta}}{SNR_{T}}\frac{\|\mathbf{X} - \mathbf{Y} \|_2}{\|\mathbf{X}\|_2} \nonumber
\end{align}
where $dim(\cdot)$ indicates the number of elements in the input tensor. $\textcircled{1}$ for $\frac{SNR_0}{SNR_{T-\delta}}\gg 1$, we have $\frac{SNR_0}{SNR_{T}}\gg 1$. $\textcircled{2}$ for $\sigma_0\rightarrow 0$ thus $\|\frac{\mathbf{X}}{\sigma_0}\|_2^2 \gg dim(\epsilon)$ and $\textcircled{3}$ for $\alpha_0 \rightarrow 1$. Meanwhile, for image restoration tasks, we usually have $\| \mathbf{X} - \mathbf{Y}\|_2 \leq k_1 \|\mathbf{X} \|_2$, where $0 \leq k_1 \leq 1$. Thus, we have $\frac{\mathcal{E}_{T-\delta}}{\mathcal{E}_T}\leq k_1$. If we utilize a small network to adaptively learn an initialization value to replace $\mathbf{Y}$, we can further reduce $k_1$.

\begin{table*}[ht!]
\centering
\scriptsize
\caption{\secrevision{Results of the ablation study on the interactions between the proposed two training techniques.}}
\vspace{-0.2cm}
\label{table:Ablative_distill}
\setlength{\tabcolsep}{0.9mm}{
\renewcommand{\arraystretch}{1.2}
\scalebox{1.3}{
\begin{tabular}{l|c|ccc|ccc|ccc}
\toprule[1.0pt]
\multirow{2}{*}{Method} & \multirow{2}{*}{NFE} & \multicolumn{3}{c|}{UIEBD} & \multicolumn{3}{c|}{Raindrop} & \multicolumn{3}{c}{LOL-v2} \\
\cmidrule(lr){3-5} \cmidrule(lr){6-8} \cmidrule(lr){9-11}
& & PSNR$\uparrow$ & SSIM$\uparrow$ & LPIPS$\downarrow$ & PSNR$\uparrow$ & SSIM$\uparrow$ & LPIPS$\downarrow$ & PSNR$\uparrow$ & SSIM$\uparrow$ & LPIPS$\downarrow$ \\
\midrule
Baseline & 10 & 24.31 & 0.916 & 0.151 & 32.86 & 0.942 & 0.059 & 28.78 & 0.895 &0.094 \\
\midrule
\multirow{3}{*}{Distillation} 
& 4 & 25.54 & 0.931 & 0.133 & 33.24 & 0.943 & 0.048 & 29.31 & 0.899 & 0.099 \\
& 2 & 25.47 & {0.930} & {0.133} & {33.17} & {0.943} & 0.052 &{29.23}  &{0.898}  &{0.103}  \\
& 1 & 25.13 & {0.927} & {0.145} & {32.98} & {0.943} & 0.056 &{29.05}  &{0.888}  &{0.112}  \\
\midrule
\multirow{3}{*}{RL+Distillation}
& 4 & 26.37 & 0.940 &0.122 &33.74 &0.947 &0.047 &29.99 &0.905 &0.086 \\
& 2 & 26.30 & {0.939} & {0.124} & {33.67} & {0.946} & 0.050 &{29.95}  &{0.905}  &{0.089}  \\
& 1 & 26.25 & {0.938} & {0.128} & {33.63} & {0.946} & 0.052 &{29.91}  &{0.904}  &{0.101}  \\
\bottomrule[1.0pt]
\end{tabular}}}
\end{table*}
\section{More Visual Results}
\label{Sec:Synthesizing}

In this section, we show additional visual results. Specifically, Figs.~\ref{fig:underwater}, \ref{fig:low_light}, and \ref{fig:deraining} illustrate visual comparisons for underwater image enhancement, low-light enhancement, and deraining, respectively, under the task-specific setting. Fig.~\ref{fig:unified} showcases the experimental results for unified image restoration.  Fig.~\ref{fig:ablation_specific} shows the visual comparison of ablation studies. \secrevision{Finally, Fig.~\ref{fig:ablation_multidegrade} indicates the visual comparison of IR with multiple degradations.}

\begin{figure*}[t]
    \centering
    \vspace{-0.4cm}
    \includegraphics[width=1.0\linewidth]{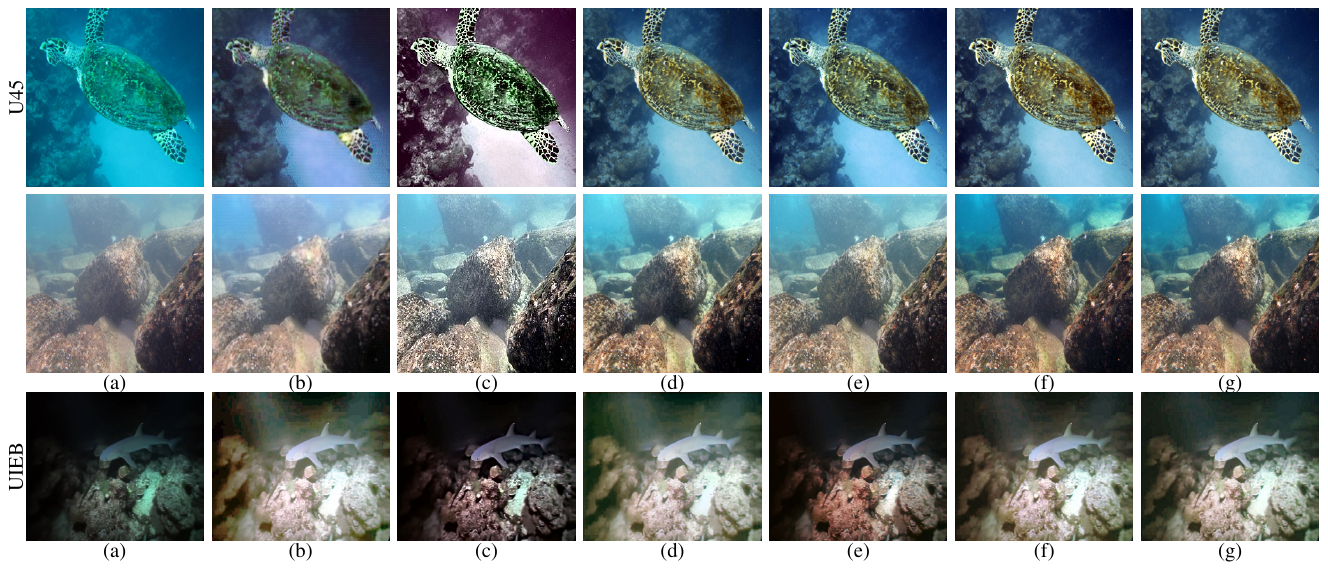}
    \vspace{-0.5cm}
    \caption{Visual comparison of underwater image enhancement on U45~\cite{ancuti2012enhancing} and UIEB~\cite{li2019underwater} datasets. U45 (\textbf{top}): (a) low-quality input, (b) CycleGAN~\cite{zhu2017unpaired}, (c) MLLE~\cite{zhang2022underwater}, (d) HCLR~\cite{zhou2024hclr}, (e) SemiUIR~\cite{huang2023contrastive}, (f) Ours($NFE=1$) and (g) Ours($NFE=10$). UIEB (\textbf{bottom}): except (b) reference image, the remaining columns are the same as those of U45. 
    }
    \label{fig:underwater}
\end{figure*}
\begin{figure*}[t]
    \centering
    \includegraphics[width=1.0\linewidth]{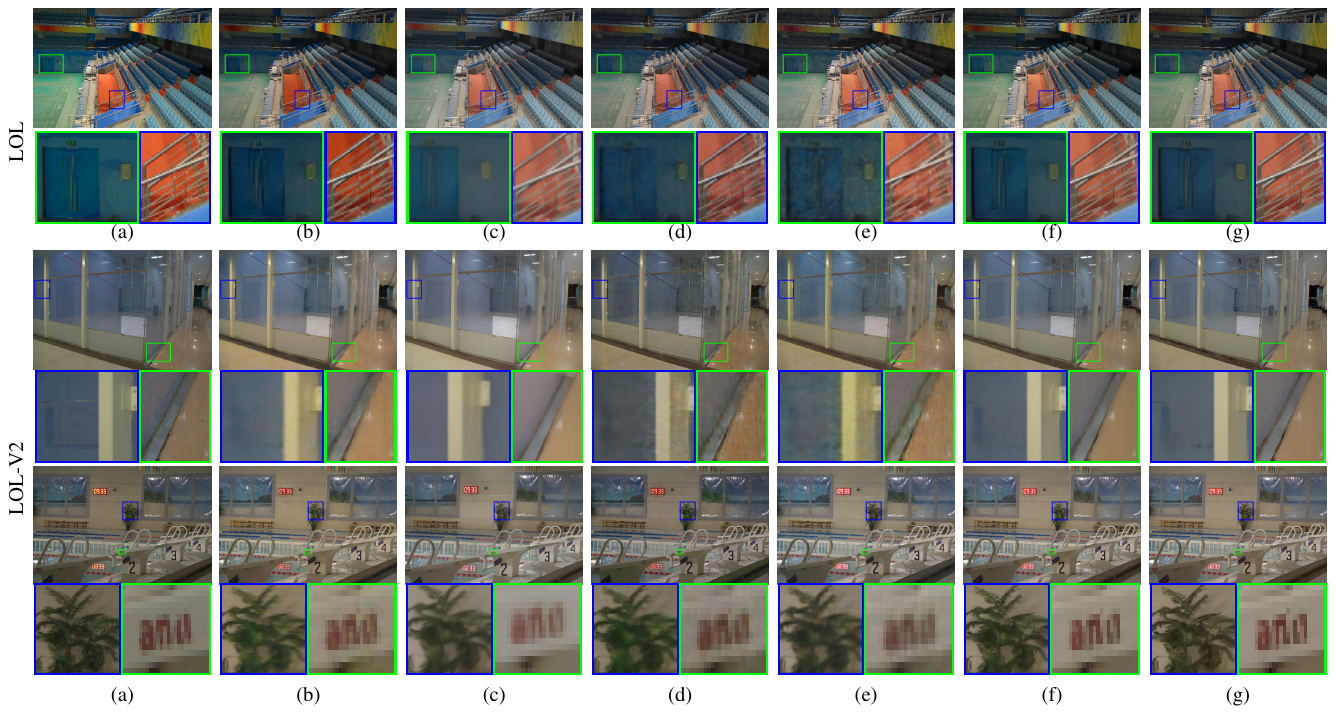}
    \vspace{-0.4cm}
    \caption{Visual comparison of low-light enhancement on LOL and LOLV2 datasets. LOL (\textbf{top}): (a) reference images, (b) CID~\cite{feng2024you}, (c) LLFlow~\cite{wang2022low}, (d) RetinexFormer~\cite{cai2023retinexformer}, (e) LLFormer~\cite{wang2023ultra}, (f) Ours ($NFE=1$), (g) Ours ($NFE=10$). LOLV2 (\textbf{bottom}): except (b) SNR-Aware~\cite{xu2022snr}, the remaining columns are the same as those of LOL. Below each figure, we also visualize zoom-in regions marked by the blue and green boxes.}
    \label{fig:low_light}
\end{figure*}

\begin{figure*}[t]
    \centering
    \includegraphics[width=1.0\linewidth]{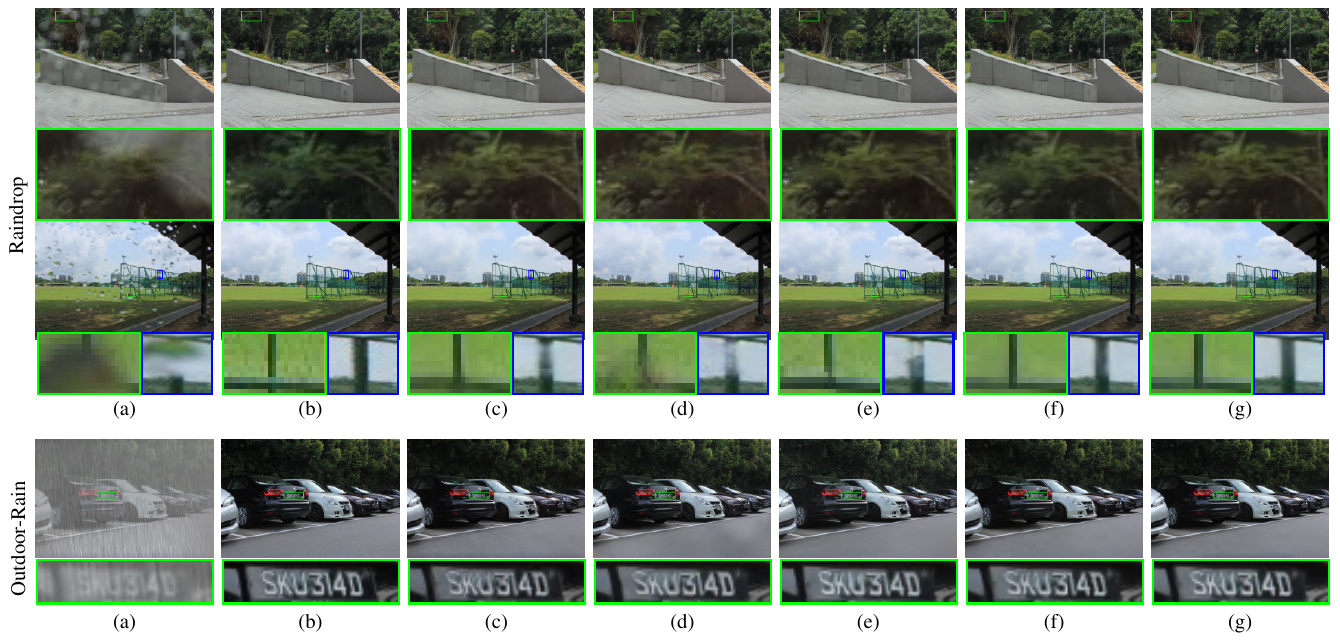} \vspace{-0.35cm}
    \caption{Visual comparison on the tasks of raindrop removal and image deraining. Raindrop removal (\textbf{top}): (a) low-quality input, (b) reference samples, (c) IDT~\cite{xiao2022image}, (d) GridFormer~\cite{wang2024gridformer}, (e) RainDropDiff~\cite{ozdenizci2023restoring}, (f) Ours ($NFE=1$), (g) Ours ($NFE=10$). Deraining (\textbf{bottom}): (a) low-quality input, (b) reference samples, (c) GridFormer~\cite{wang2024gridformer}, (d) WeatherDiff64~\cite{ozdenizci2023restoring}, (e) WeatherDiff128~\cite{ozdenizci2023restoring}, (f) Ours ($NFE=1$), (g) Ours ($NFE=10$).}
    \label{fig:deraining}
\end{figure*}

\begin{figure*}[t]
    \centering
    \includegraphics[width=1.0\linewidth]{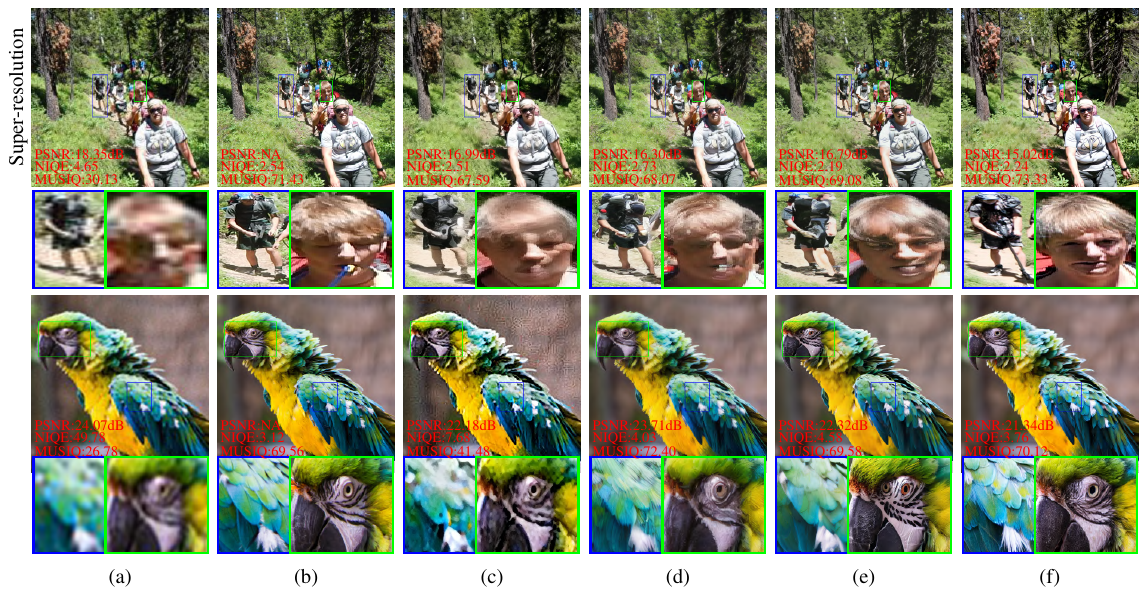}
    \vspace{-1.5em}
    \caption{Additional visual comparison of unified image restoration. In the super-resolution task, (a)-(f) indicates the input low-quality measurement, reference image, PASD~\cite{yang2023pasd}, SeeSR~\cite{wu2024seesr}, and FLUX-IR(Ours), respectively. We annotated evaluation metrics of corresponding images by PSNR~$\uparrow$, NIQE~$\downarrow$, and MUSIQ~$\uparrow$, respectively.}
    \label{fig:unified}
\end{figure*}

\begin{figure*}[t]
    \centering
\vspace{-0.2cm}
    \includegraphics[width=1\linewidth]{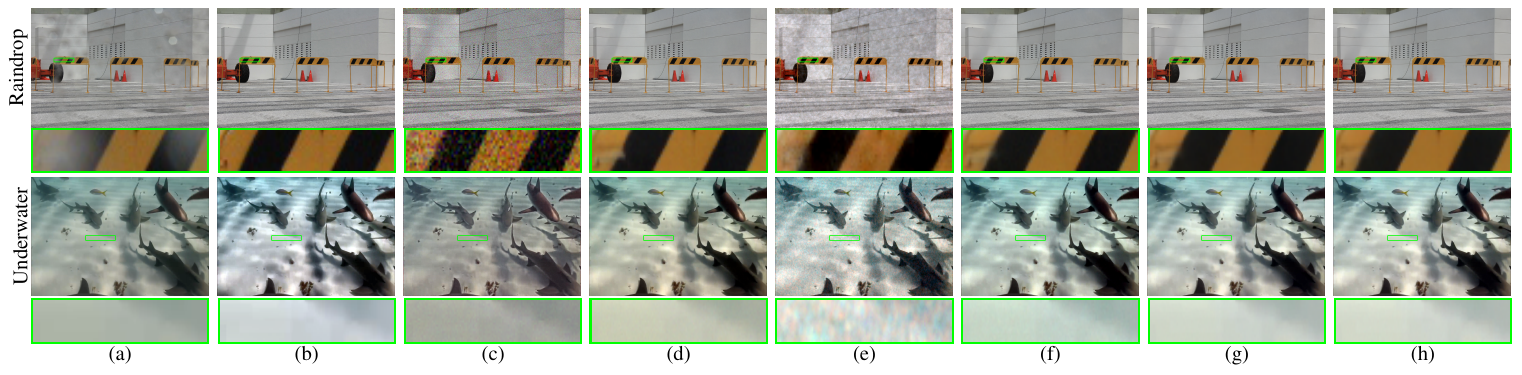}
    \caption{Visual demonstration of ablation studies. (a) low-quality image, (b) reference image, (c) pretrained model ($NFE=1$, Table~\textcolor{red}{XI}-\textcolor{red}{1}), (d) pretrained model ($NFE=10$, Table~\textcolor{red}{XI}-\textcolor{red}{1}), (e) RL ($NFE=1$, Table~\textcolor{red}{XI}-\textcolor{red}{2}), (f) RL w/ DISTILL ($NFE=1$, Table~\textcolor{red}{XI}-\textcolor{red}{3}), (g) (f) w/ latent INTER ($NFE=1$, Table~\textcolor{red}{XI}-\textcolor{red}{4}), (h) (g) w/ NGS ($NFE=1$, Table~\textcolor{red}{XI}-\textcolor{red}{5})}
    \label{fig:ablation_specific}
\vspace{-0.4cm}
\end{figure*}

\begin{figure*}[t]
    \centering
\vspace{-0.2cm}
    \includegraphics[width=0.9\linewidth]{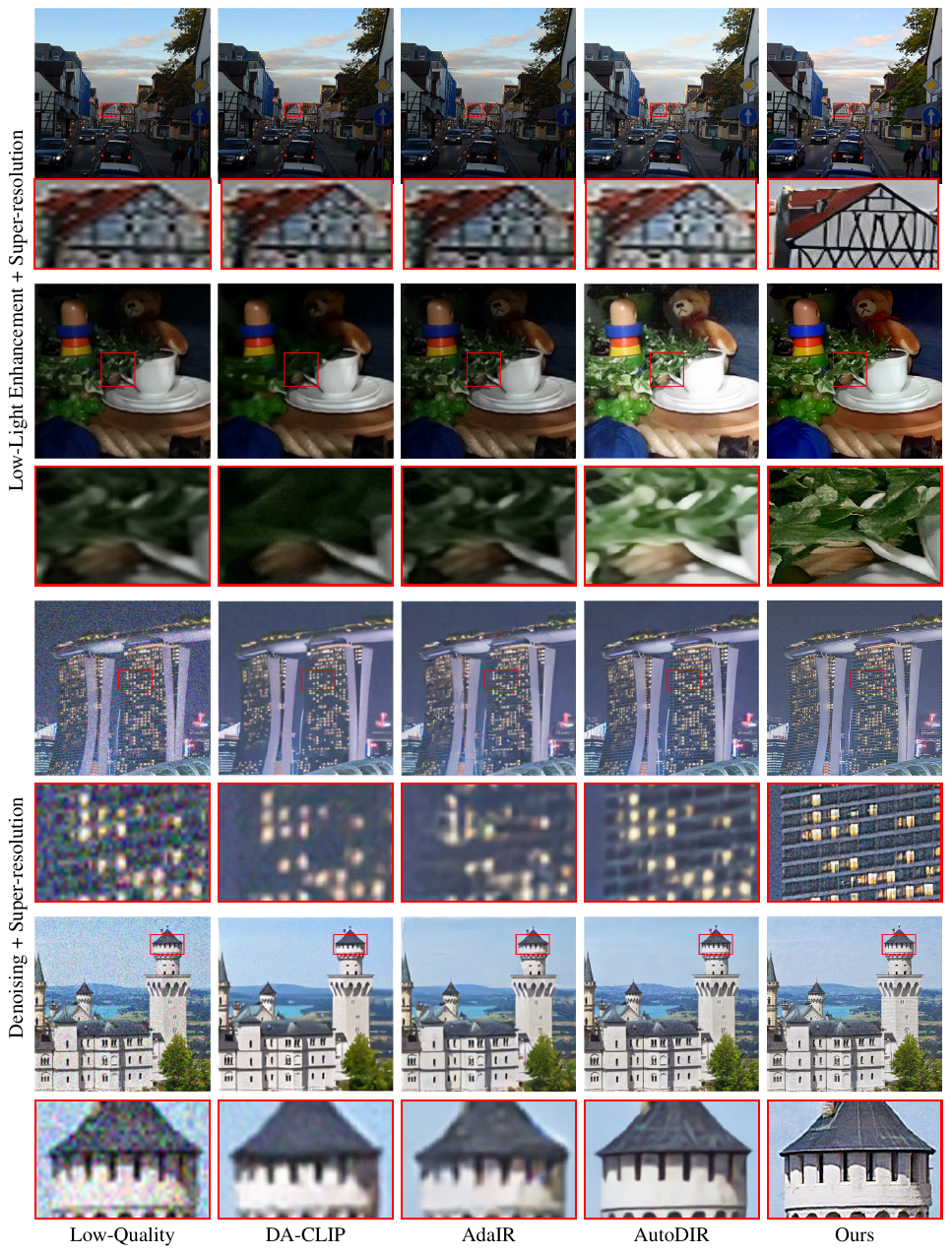}
    \caption{\secrevision{Visual comparisons of the proposed method against SOTA methods, e.g., DA-CLIP~\cite{luo2023controlling}, AdaIR~\cite{cui2024adair}, AutoDIR~\cite{jiang2023autodir}, in addressing various
combinations of image degradations.}
    \label{fig:ablation_multidegrade}}
\vspace{-0.4cm}
\end{figure*}

\section{\secrevision{Experimental Validation of Cross-Effect of Reinforcement Learning and Distillation Processes}}
\secrevision{We have conducted further experiments to validate the interactions between key components, i.e., the combination of RL and distillation versus distillation only. Experimental results are shown in Table~\ref{table:Ablative_distill}, which indicates that the complete training strategy (RL+Distillation) can consistently outperform the distillation baseline with different NFEs, which highlights the effectiveness of RL in the proposed method. Moreover, we can also observe the effectiveness of distillation, which can not only accelerate the process but also slightly improve performance in certain settings.}

\section{\secrevision{Experimental Comparisons on Different Reinforcement Learning Strategy}}
\label{Sec:Synthesizing}

\begin{table}[t!]
\centering
\scriptsize
\caption{\secrevision{Results of the ablation study on different reinforcement learning schemes.}}
\vspace{-0.2cm}
\label{table:Ablative_RL}
\setlength{\tabcolsep}{0.9mm}{
\renewcommand{\arraystretch}{1.2}
\scalebox{1.1}{
\begin{tabular}{l|cc|cc|cc}
\toprule[1.0pt]
\multirow{2}{*}{Method} & \multicolumn{2}{c|}{UIEBD} & \multicolumn{2}{c|}{Raindrop} & \multicolumn{2}{c}{LOL-v2} \\
\cmidrule(lr){2-3} \cmidrule(lr){4-5} \cmidrule(lr){6-7}
& PSNR$\uparrow$ & LPIPS$\downarrow$ & PSNR$\uparrow$ & LPIPS$\downarrow$ & PSNR$\uparrow$ & LPIPS$\downarrow$ \\
\midrule
Baseline & 24.31   &0.151 &32.86 &0.059 &28.78 &0.094 \\
ImageReward~\cite{xu2023imagereward} &24.65 &0.273 &33.04 &0.048 &28.98 &0.094  \\
DPOK~\cite{fan2023dpok}        &24.45 &0.193 &33.06 &0.050 &29.11 &0.096  \\
D3PO~\cite{yang2024using}        &24.51 &0.242 &32.96 &0.049 &28.89 &0.098  \\
\textbf{Proposed Method} & \textbf{25.08} & \textbf{0.162} & \textbf{33.32}  & \textbf{0.044} & \textbf{29.54} &\textbf{0.086}  \\
\bottomrule[1.0pt]
\end{tabular}}
}
\end{table}

\secrevision{In this section, we conduct further experimental comparisons with three different reinforcement learning algorithms, e.g., ImageReward~\cite{xu2023imagereward}, DPOK~\cite{fan2023dpok}, and D3PO~\cite{yang2024using}. Specifically, we utilized those different reinforcement learning algorithms to fine-tune the same pretraining diffusion models on three different tasks. All methods were trained with 30,000 iterations. 
\begin{table}[t!]
\centering
\scriptsize
\caption{ \secrevision{Experimental validation of the performance with different neural network architectures.}}
\vspace{0.1cm}
\label{table:Ablative_structure}
\setlength{\tabcolsep}{0.9mm}{
\renewcommand{\arraystretch}{1.2}
\scalebox{1.2}{
\begin{tabular}{l|c|cc|cc}
\toprule[1.0pt]
\multirow{2}{*}{Architecture} & \multirow{2}{*}{Training method} & \multicolumn{2}{c|}{Rain 200 L} & \multicolumn{2}{c}{Rain 200 H}  \\
\cmidrule(lr){3-4} \cmidrule(lr){5-6}
& & PSNR$\uparrow$  & SSIM$\uparrow$ & PSNR$\uparrow$  & SSIM$\uparrow$ \\
\midrule
\multirow{2}{*}{Conv-Unet~\cite{hou2024global}}  & Original~\cite{hou2024global}  &39.45  &0.9851  &28.60  &0.9029   \\
                            & Ours      & 40.51 & 0.9871 & 30.33 &  0.9181   \\
\midrule
\multirow{2}{*}{Transformer~\cite{chen2023learning}}& Original~\cite{chen2023learning}  & 41.23 & 0.9894 & 32.17 & 0.9326    \\
                            & Ours      &41.87  &0.9906  &32.62 &0.9366     \\
\bottomrule[1.0pt]
\end{tabular}}
}
\end{table}
Experimental results in Table~\ref{table:Ablative_RL} indicate that our method outperforms the ImageReward~\cite{xu2023imagereward}, DPOK~\cite{fan2023dpok}, and D3PO~\cite{yang2024using} on all task-specific IR. Moreover, it demonstrates that all methods outperform the baseline model on the metric of PSNR, which is also the utilized reward. However, for the perceptual metrics of LPIPS, the compared methods' performance even degrades, which indicates an overfitting-like result. However, our methods achieve consistently good performance, showing the robustness of the proposed RL methods.  Moreover, we want to note that our RL method is unique for the consideration of the probabilistic nature of the diffusion method, i.e., aligning ODE to M-SDE, which may benefit the robustness.}

\section{\secrevision{Experimental Validation of Applying the Proposed Method on Different Architectures.}}
 \secrevision{The proposed strategy is orthogonal to the neural network structure, i.e., we can also utilize transformer-based network to build a diffusion model and apply our algorithm to boost their performance. To further validate the claim, in this section, we modified a SOTA transformer-based method, DRSformer~\cite{chen2023learning} to a diffusion model and then applied the proposed method to boost the plain transformer-based diffusion model. Experimental results in Table~\ref{table:Ablative_structure} show that the proposed method can improve both Conv Unet-based and Transformer-based baselines. Note that through such integration, the proposed method actually achieves the SOTA performance on Rain 200L/H dataset. The SOTA performance of the proposed method on 13 datasets across 3 task-specific IR tracks illustrates the effectiveness of the proposed method.}
}

\vfill
\end{document}